\documentclass{article}

\usepackage[final]{neurips_data_2024}

\usepackage[utf8]{inputenc} 
\usepackage[T1]{fontenc}    
\usepackage{hyperref}       
\usepackage{url}            
\usepackage{booktabs}       
\usepackage{amsfonts}       
\usepackage{nicefrac}       
\usepackage{microtype}      
\usepackage{colortbl}
\usepackage{MnSymbol,bbding,pifont}
\usepackage{natbib}
\setcitestyle{numbers,square}

\usepackage{caption}
\usepackage{
tikz,
relsize,
amsmath,
booktabs,
tikz
}

\usepackage{xcolor}         
\definecolor{Slide}{HTML}{F27970}
\definecolor{Research Report}{HTML}{BB9727}
\definecolor{Financial Report}{HTML}{54B345}
\definecolor{Brochure}{HTML}{32B897}
\definecolor{Academic Paper}{HTML}{FFA500}
\definecolor{Guideline}{HTML}{05B9E2}
\definecolor{Webpage Screenshot}{HTML}{8983BF}
\definecolor{Poster}{HTML}{C76DA2}
\definecolor{Industry File}{HTML}{000000}
\definecolor{darkgreen}{RGB}{50,100,0}
\definecolor{darkred}{RGB}{200, 0, 0}
\definecolor{firstBest}{rgb}{0.86, 1, 0.86}
\definecolor{secondBest}{rgb}{1, 0.91, 0.93}

\newcommand{\cmark}{\textcolor{darkgreen}{\ding{51}}} 
\newcommand{\xmark}{\textcolor{darkred}{\ding{55}}} 
\newcommand{\cxmark}{\ding{52}\rotatebox[origin=c]{-9.2}{\kern-0.7em\ding{55}}}

\newcommand{\ie}{\emph{i.e.,}\xspace}
\newcommand{\eg}{\emph{e.g.,}\xspace}
\newcommand{\etc}{\emph{etc.}\xspace}
\newcommand{\header}[1]{\text{#1}}

\usepackage{xspace}
\usepackage{multirow}
\usepackage{makecell}
\usepackage{tablefootnote}
\usepackage{graphicx}
\usepackage{wrapfig} 
\usepackage{lipsum}   
\usepackage{boxedminipage}
\usepackage{listings}

\let\svthefootnote\thefootnote
\newcommand\blankfootnote[1]{%
  \let\thefootnote\relax\footnotetext{#1}%
  \let\thefootnote\svthefootnote%
}

\lstdefinestyle{prompt}{
  basicstyle=\footnotesize\ttfamily,
  columns=fullflexible,
  breaklines=true,
  frame=none,
  extendedchars=true,
  escapechar=@,
  literate={á}{{\'a}}1 {ã}{{\~a}}1 {é}{{\'e}}1 {£}{{\pounds}}1 {–}{{-}}1 {’}{{'}}1,
  frame=lines
}

\newcommand{\benchmarkname}{\textsc{MMLongBench-Doc}\xspace}

\title{\benchmarkname: Benchmarking Long-context Document Understanding with Visualizations}

\author{%
  Yubo Ma$^{1}$, Yuhang Zang$^{2\ast}$, \textbf{Liangyu Chen}$^{1}$,  \textbf{Meiqi Chen}$^{3}$, \textbf{Yizhu Jiao}$^{4}$ \\ \textbf{Xinze Li}$^{1}$, \textbf{Xinyuan Lu}$^{5}$, \textbf{Ziyu Liu}$^{6}$, \textbf{Yan Ma}$^{7}$, \textbf{Xiaoyi Dong}$^{2}$, \textbf{Pan Zhang}$^{2}$ \\ \textbf{Liangming Pan}$^{8}$, \textbf{Yu-Gang Jiang}$^{9}$, \textbf{Jiaqi Wang}$^{2}$, \textbf{Yixin Cao}$^{9\ast}$, \textbf{Aixin Sun}$^{1}$ \\
  $^{1}$ S-Lab, Nanyang Technological University, $^{2}$ Shanghai AI Laboratory,
  $^{3}$ Peking University \\
  $^{4}$ University of Illinois Urbana-Champaign, $^{5}$ National University of Singapore, $^{6}$ Wuhan University \\ $^{7}$ Singapore Management University, $^{8}$ University of Arizona, $^{9}$ Fudan University \\
}

\begin{document}

\maketitle

\renewcommand{\thefootnote}{\fnsymbol{footnote}}
\footnotetext[1]{Corresponding Authors.}
\renewcommand{\thefootnote}{\arabic{footnote}}

\begin{abstract}
Understanding documents with rich layouts and multi-modal components is a long-standing and practical task. Recent Large Vision-Language Models (LVLMs) have made remarkable strides in various tasks, particularly in single-page document understanding (DU). However, their abilities on long-context DU remain an open problem.
This work presents \textbf{\benchmarkname}\blankfootnote{Project Page: \href{https://mayubo2333.github.io/MMLongBench-Doc}{\texttt{https://mayubo2333.github.io/MMLongBench-Doc}}}, a long-context, multi-modal benchmark comprising 1,082 expert-annotated questions. Distinct from previous datasets, it is constructed upon 135 lengthy PDF-formatted documents with an average of 47.5 pages and 21,214 textual tokens. Towards comprehensive evaluation, answers to these questions rely on pieces of evidence from (1) different sources (text, image, chart, table, and layout structure) and (2) various locations (\ie page number). Moreover, 33.7\% of the questions are \textit{cross-page questions} requiring evidence across multiple pages. 20.6\% of the questions are designed to be \textit{unanswerable} for detecting potential hallucinations. Experiments on 14 LVLMs demonstrate that long-context DU greatly challenges current models. Notably, the best-performing model, GPT-4o, achieves an F1 score of only 44.9\%, while the second-best, GPT-4V, scores 30.5\%. Furthermore, 12 LVLMs (all except GPT-4o and GPT-4V) even present worse performance than their LLM counterparts which are fed with lossy-parsed OCR documents. These results validate the necessity of future research toward more capable long-context LVLMs.
\end{abstract}

\begin{figure}[!h]
\centering
    \includegraphics[width=0.96\linewidth]{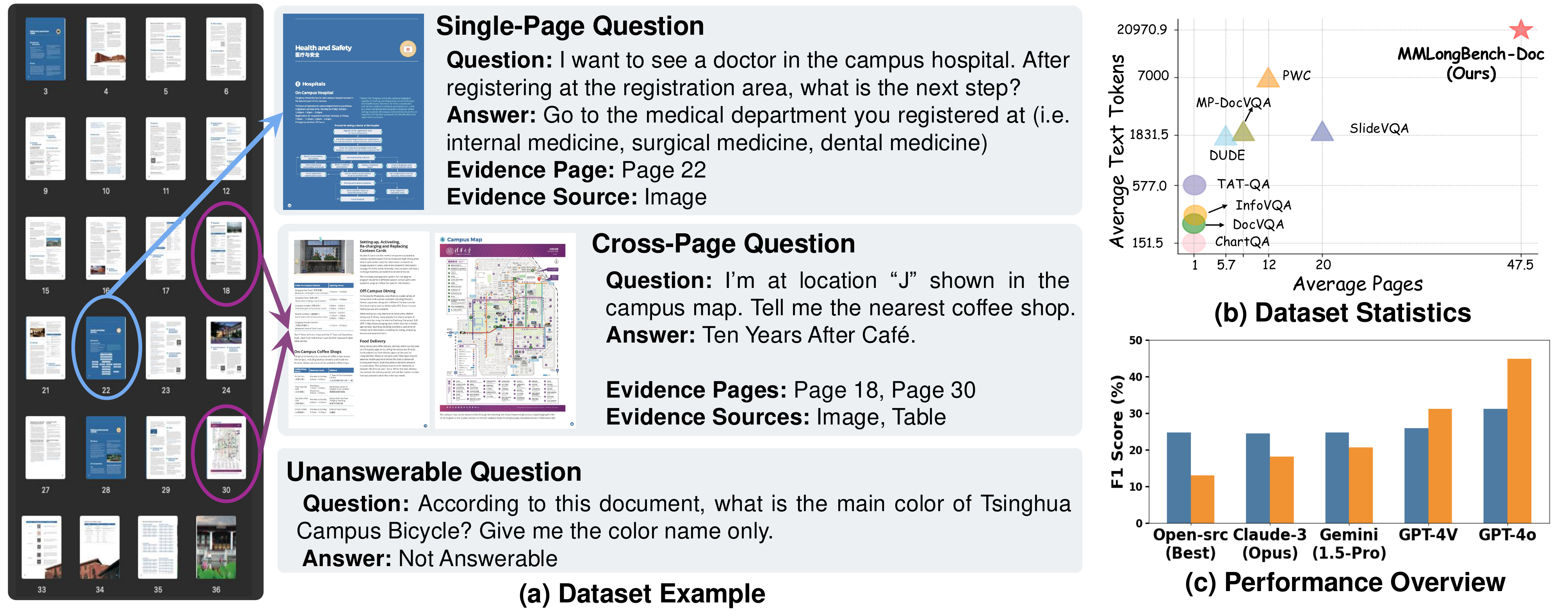}
\caption{\benchmarkname evaluates understanding abilities of LVLMs on lengthy documents that span tens of pages and incorporate multi-modal elements. Experiments (bottom-right) indicate that most LVLMs struggle, even falling behind LLMs that are fed with only OCR-parsed documents.}
\label{fig: examples}
\end{figure}
\section{Introduction}

Documents are one of the fundamental forms of information preservation and exchange. In each year, tens of millions of documents are created, read, saved, and dispatched~\cite{Bornmann2014GrowthRO}. Beyond unstructured pure-text, documents feature both complicated layout structures and information across distinct modalities such as text, table, chart, image, \etc Accordingly, the automatic understanding of documents (Document Understanding; DU) stands as a long-standing task in urgent and practical needs.

Recently, a number of LVLMs, both closed-source ones (GPT-4o~\cite{gpt4o}, Gemini-1.5~\cite{geminiteam2024gemini}, Claude-3~\cite{claude3}, \etc) and open-source ones (InternLM-XC2-4KHD~\cite{dong2024internlm}, InternVL-Chat~\cite{chen2024far}, Otter~\cite{li2023otter}, LLaVA-NeXT~\cite{li2024llavanext-strong}, CogVLM~\cite{wang2023cogvlm}, mPLUG-DocOwl 1.5~\cite{hu2024mplugdocowl}, TextMonkey~\cite{liu2024textmonkey}, \etc) have been developed and presented the great potential to handle documents. Most of them have achieved promising performance on single-page DU datasets like DocVQA~\cite{Mathew2020DocVQAAD}, ChartQA~\cite{masry-etal-2022-chartqa}, InfoVQA~\cite{Mathew2021InfographicVQA}, TAT-DQA~\cite{zhu2022towards}, \etc However, considerable amounts of documents in the real world are long-context documents with tens or even hundreds of pages. The understanding of these lengthy documents brings new challenges for LVLMs from at least two aspects: (1) \textbf{Localization}: identify and retrieve information from massive, heterogeneous information (similar to the \textit{needle in a haystack} task); (2) \textbf{Cross-page comprehension}: collect and reason over multi-source information across different pages. These two kinds of abilities are beyond the evaluation scopes of the aforementioned single-page DU datasets. Some recent DU datasets~\cite{tito2023hierarchical, VanLandeghem2023DocumentUD, SlideVQA2023} feature multiple-page DU, but almost all their documents are either as short of only several pages or of low information density, making the localization-related questions over-simple. Additionally, few (if any) questions in these datasets necessitate cross-page comprehension. See more detailed related work in Section~\ref{sec: related_work}. In summary, there lacks a unified and high-quality benchmark on lengthy documents, leaving the evaluation of long-context DU largely unexplored.

In this paper, we present \benchmarkname, a benchmark designed to evaluate the \textbf{M}ulti-\textbf{M}odality \textbf{Long}-context \textbf{Doc}ument understanding abilities of LVLMs. Towards a comprehensive benchmark, it incorporates lengthy documents from both four existing datasets~\cite{masry-etal-2022-chartqa, VanLandeghem2023DocumentUD, SlideVQA2023, islam2023financebench} and other various papers, brochures, \etc Consequently, our benchmark includes 135 PDF-formatted documents spanning across 7 diverse domains, with each document averaging 47.5 pages and 21,214.1 textual tokens. Regarding the questions, we employ ten expert-level annotators to (1) edit questions associated with documents from existing datasets to meet our benchmark's standard and (2) create new questions for all collected documents to expand the scale of the benchmark. Then a three-round, semi-automatic reviewing process ensures the benchmark's annotation quality. As a result, \benchmarkname comprises 1,082 human-annotated questions, with 184 sourced from four existing datasets and 898 newly annotated. Being a multi-modal benchmark, the answer to each question requires evidence from one or more of these five in-document sources: \textit{text}, \textit{layout}, \textit{chart}, \textit{table}, and \textit{image}. Questions are categorized into three types based on the number of evidence pages~\footnote{Given a document $D$ and a question $q$ upon $D$, We call page $P$ (in document $D$) an \textit{evidence page} of $q$ if the answer of $q$ necessitates one or more pieces of evidence in page $P$.}, with examples illustrated in Figure~\ref{fig: examples}(a): (1) 494 \textit{single-page} questions (with one evidence page) mainly to evaluate localization abilities, (2) 365 \textit{cross-page} questions (with multiple evidence pages) to assess cross-page comprehension, and (3) 223 \textit{unanswerable} questions (no evidence for answering it, \ie no evidence pages) to reduce shortcuts and measure LVLMs' potential hallucinations. Meta-information including evidence pages, sources, and answer formats, is preserved for fine-grained evaluation and analysis. Detailed descriptions of the annotation pipeline and statistics can be found in Section~\ref{sec: dataset}.

We conduct extensive experiments on \benchmarkname to evaluate the long-context DU abilities of 14 LVLMs, including 4 proprietary and 10 open-source ones. Given a document, we screenshot each page and feed all of these PNG-formatted images to LVLMs in an end-to-end approach. For comparison, we also convert the documents to textual format by optical character recognition (OCR) and evaluate another 6 proprietary and 4 open-source 10 LLMs (6 proprietary and 4 open-source ones). The results in Figure~\ref{fig: examples}(c) highlight the challenges that current LVLMs face with long-context DU. The best-performing LVLM, GPT-4o, achieves an overall F1 score of only 44.9\%, while the second-best LVLM, GPT-4V, scores 30.5\%. Moreover, all the remaining LVLMs tested with multi-modal documents performed worse than single-modal LLMs handling lossy, OCR-parsed texts. Specifically, the Gemini-1.5-Pro and Claude-3-Opus present 4.2\% and 6.4\% absolute decrease when the inputs change from document screenshots to OCR-parsed texts. Regarding open-source models, the best-performing LVLM lags behind the best-performing LLM by 11.7\%. These results reveal that long-context DU is a far-from-resolved task for current LVLMs.

\begin{table}[t]
\centering
\small
\caption{Comparison between our benchmark and previous DU datasets. \textbf{Unans.}: unanswerable question. \textbf{TXT/L/C/TAB/I}: pure text/generalized layout/chart/table/image. \textbf{Doc. Rel.}: document relevance. Whether document information is indispensable for the answer. \textbf{Avg. Position}: the average page index on which the answer evidence is located. *:Statistics from \cite{Borchmann2021DUEED}.}
\resizebox{0.95\linewidth}{!}{
\begin{tabular}{l|cc|ccccc}
\toprule
\multirow{2}{*}{\textbf{Benchmarks}}  & \multicolumn{2}{c}{\textbf{Document}} & \multicolumn{2}{|c}{\textbf{Question type}} & \multicolumn{3}{c}{\textbf{Answer Evidence}} \\
& \# Pages & \# Tokens & Cross-page (\%) & Unans. (\%) & Doc. Rel. & Source & Avg. Position  \\
\midrule
DocVQA~\cite{Mathew2020DocVQAAD} & 1.0 & 151.5 &  \xmark & \xmark & \cxmark &  TXT/L/C/TAB/I & - \\
ChartQA~\cite{masry-etal-2022-chartqa}\tablefootnote{We view website screenshots and posters as generalized documents and define \textit{equivalent page number} (\texttt{EPN}) to measure their context lengths: $\texttt{EPN(D) = ceil}(\frac{\texttt{Pixel(D)}}{P})$. Here \texttt{Pixel(D)} is the pixel number of generalized document \texttt{D}, and \texttt{P} is the average pixel numbers of each page (converting from .pdf to .png format with resolution 240) in \benchmarkname. \label{equivalent pages}} & 1.0 & 236.9 & \xmark & \xmark & \cmark & C & -  \\
InfoVQA~\cite{Mathew2021InfographicVQA}\textsuperscript{\ref{equivalent pages}} & 1.2 & 288.0  & \xmark & \xmark & \cxmark &  L/C/TAB/I & - \\
TAT-DQA~\cite{zhu2022towards} & 1.1 & 577.0 & \xmark & \xmark & \cxmark & TXT/TAB & -   \\
VisualWebBench~\cite{liu2024visualwebbench}\textsuperscript{\ref{equivalent pages}} & 1.0 & 452.4  & \xmark & \xmark & \cmark & LAY/I & - \\
PWC~\cite{kardas-etal-2020-axcell} & \textasciitilde 12* & \textasciitilde 7000* & \xmark & \xmark & \cxmark & TAB & - \\
MP-DocVQA~\cite{tito2023hierarchical}  & 8.3 & 2026.6 & \xmark & \xmark & \cxmark & TXT/L/C/TAB/I & 6.0 \\
DUDE~\cite{VanLandeghem2023DocumentUD} & 5.7 & 1831.5 & \cmark (2.1\%) & \cmark(12.7\%) & \cxmark & TXT/L/C/TAB/I & 2.5 \\
SlideVQA~\cite{SlideVQA2023} & 20.0 & 2030.5 & \cmark (13.9\%) & \xmark & \cxmark & TXT/L/C/TAB/I & 9.1 \\
\midrule
\benchmarkname & 47.5 & 21214.1 & \cmark (33.0\%) & \cmark (22.5\%) & \cmark & TXT/L/C/TAB/I & 23.6 \\
\bottomrule
\end{tabular}}
\label{tab:dataset_comparison}
\end{table}
\section{Related Work}
\label{sec: related_work}

\textbf{Benchmarks for Document Understanding.} A great amount of datasets have emerged to evaluate the DU capabilities of LVLMs. Many datasets focus exclusively on either a single component (\eg table, chart)~\cite{masry-etal-2022-chartqa, zhu2022towards, liu2024visualwebbench, kardas-etal-2020-axcell} or a single page~\cite{Mathew2020DocVQAAD, Mathew2021InfographicVQA} from the full documents. Some recent DU datasets~\cite{tito2023hierarchical, VanLandeghem2023DocumentUD, SlideVQA2023, saadfalcon2023pdftriage, islam2023financebench} attempt to assess multi-page documents, but still exhibit shortcomings in terms of document length (page number), information density (token number) and the construction approaches. Specifically, MP-DocVQA~\cite{tito2023hierarchical} is an extension of DocVQA~\cite{Mathew2020DocVQAAD} and inherently absent of both cross-page and unanswerable questions. Annotating from scratch, DUDE~\cite{VanLandeghem2023DocumentUD} includes a small percentage of cross-page questions (2.1\%) and unanswerable questions (12.7\%). However, due to the relatively short context length (5.3 pages on average) and the use of crowd-sourced annotations, questions in DUDE tend to be less challenging and somewhat less rigorous. SlideVQA features 20-page documents and cross-page questions (12.9\%). Nevertheless, the documents in SlideVQA are in slide-deck format and of relatively low information density. Moreover, these cross-page questions are HotpotQA-style~\cite{yang-etal-2018-hotpotqa} created by instantiating entity graphs and co-referencing in-graph entities across multiple pages. The entity graph from a closed document tends to be sparse and has significant shortcuts (see examples in Appendix~\ref{subsec: Existing Question Editing}). These shortcuts sometimes lead to false cross-page questions that actually do not require answer evidence across different pages. The recent FinanceBench~\cite{islam2023financebench} features both extremely long-context documents and practical, scalable cross-page questions. However, its documents are exclusively financial reports. Additionally, the reference answers are in open-ended formats, making the expert-level manual evaluation indispensable. The above reasons limit the broader applicability of FinanceBench. To our best knowledge, \benchmarkname is the first comprehensive, qualified, and easy-to-use benchmark on the long-context DU task. More detailed descriptions and comparisons are presented in Table~\ref{tab:dataset_comparison}.

\paragraph{Models for Document Understanding.} 
There are two main branches of models for automatic DU tasks. The first approach employs two-stream, OCR-dependent architectures to separately encode textual information (parsed via OCR) and visual information (images and/or layout structures)~\cite{LayoutLMv1, xu-etal-2021-layoutlmv2, LayoutLMv3}. In contrast, the second approach develops OCR-free models that understand documents in an end-to-end manner~\cite{kim2022donut, Pix2Struct}. With the rapid advancement of LVLMs, the latter approach has dominated the current DU solutions. As mentioned above, a range of LVLMs demonstrate promising performance on single-page DU datasets. However, as shown in Section~\ref{sec:evaluation}, even the most advanced LVLMs fall significantly short of achieving satisfactory performance on our benchmark. It reveals that understanding lengthy documents still poses great challenges to current LVLMs.

\paragraph{Long-context LVLMs and LLMs.} 
Lengthy documents necessitate the use of LVLMs or LLMs with extended context sizes. Several benchmarks~\cite{shaham-etal-2022-scrolls, an2023leval, bai2023longbench, zhang2024inftybench} and solutions~\cite{Tworkowski2023FocusedTC, Chen2023LongLoRAEF, Peng2023YaRNEC, bai2024longalign} have been proposed to evaluate and develop long-context LLMs. However, there exists limited related work for long-context LVLMs, leaving this area largely unexplored. Until very recently, contemporary studies~\cite{song2024milebench, jiang2024mantis, lu2024text} assess and/or improve LVLMs' multi-image understanding capabilities. Evaluations on both \benchmarkname and these works indicate that current LVLMs are still not fully equipped to handle long-context DU and many other practical tasks that require extensive contextual comprehension.
\section{\benchmarkname}
\label{sec: dataset}
We design a three-stage annotation pipeline for the construction of our benchmark. The three stages will be introduced in Section~\ref{subsec: document collection}, Section~\ref{subsec: qa collection}, and Section~\ref{subsec: quality control}, respectively. We also provide key statistics of our benchmark in Section~\ref{subsec: dataset_overview}.


\subsection{Document Collection}
\label{subsec: document collection}
As a long-context DU benchmark, the documents shall be of diverse topics and lengthy enough. To this end, we crawl a great amount of documents from various sources. Then we select the lengthy ones from these documents. Specifically, we encompass a diverse array of documents from two approaches. (1) \textbf{Existing documents} from four previous datasets: DUDE~\cite{VanLandeghem2023DocumentUD}, SlideVQA~\cite{SlideVQA2023}, ChartQA~\cite{masry-etal-2022-chartqa}, and FinanceBench~\cite{islam2023financebench}. (2) \textbf{Newly-collected documents} from Arxiv~\footnote{\texttt{https://arxiv.org}}, ManualsLib~\footnote{\texttt{https://www.manualslib.com}} and Google Search~\footnote{\texttt{https://www.google.com.sg}}. Then we (1) filter out the documents with fewer than 15 pages or license restrictions and (2) down-sample documents from DUDE, SlideVQA, and FinanceBench for a more balanced distribution. Detailed descriptions of our selection and processing procedure can be found in Appendix~\ref{appendix: Existing Document Collection} and Appendix~\ref{appendix: Newly-annotated Document Collection}.

In summary, we collect a total of 135 documents. Among them, 76 documents are from existing datasets 
and incorporate previously annotated questions (represented as triangles). The remaining 59 documents are newly collected 
and incorporate no existing questions. We manually categorize them into 7 types: \textit{Research Report}, \textit{Financial Report}, \textit{Academic Paper}, \textit{Brochure}, \textit{Guideline}, \textit{Administration \& Industry File}, \textit{Tutorial / Workshop}. We showcase some instances of these documents in Appendix~\ref{appendix: Document Examples}.

\subsection{Question and Answer Collection}
\label{subsec: qa collection}
To serve as a high-quality and comprehensive benchmark, the question annotation of our benchmark adheres to the following standards: (1) All questions shall be neither over-easy nor over-difficult. (2) Questions are not repetitively derived from the same page or the same pattern. (3) The distribution of evidence numbers, evidence sources, and evidence locations for the questions shall be balanced. (4) No questions shall be answered correctly without accessing the relevant documents. 

Ten authors serve as expert-level annotators for the question-and-answer collection. All of them are doctors or Ph.D. students proficient in English reading and writing. Before formal annotation, they undergo a training session and pre-annotate three documents for practice. We iteratively review their annotation results and provide personalized feedback until their annotations meet the standards mentioned above. Regarding the formal annotation, we divide 135 documents into 54 batches (each having 2-4 documents) and dispatch these batches to annotators. We then ask the annotators to submit their results in units of batches and set reasonable time intervals for each batch's submission. We timely evaluate their annotations after each submission and remind the annotators if their questions in this turn diverge from the standards. It avoids the annotators rushing all assignments in a short time and benefits the annotation quality. We recommend the annotators take 60-90 minutes on each document. Specifically, the annotators shall rapidly read through the whole document in the first 15-30 minutes. For the remaining time, they shall dive deep into specific components to modify existing annotations and/or add new annotations as detailed below.

\noindent \textbf{Modify Existing Questions.} 
Documents collected from existing datasets had been annotated with some questions and answers from previous work. However, their crowd-sourcing annotations inevitably make some questions, answers, and other meta information unqualified. Therefore, we edit their annotations before including them as a component of our benchmark.

Specifically, we classify six potential problems in original annotations: \textit{Wrong Answers or Evidence Pages}, \textit{Repetitive Question}, \textit{Ambiguous Question}, \textit{Decontextualization-required Question}, \textit{Low Document-relevant Question} and \textit{Potential Shortcut}. See detailed explanations and examples about these problems in Appendix~\ref{subsec: Existing Question Editing}. Given an existing document, the annotators are tasked to evaluate each existing question's quality according to whether they have one or more above problems and assign a label from \{\texttt{Retain}, \texttt{Revise}, \texttt{Remove}\} for each question. Then the annotators would revise the \texttt{Revise} questions to meet our quality criteria and remove the \texttt{Remove} questions. 
Among all 425 original questions from 76 existing documents, 32.2\% of them are revised and 46.1\% are removed. We finally collect 211 questions in this procedure. The corresponding GUI is shown in Appendix~\ref{appendix: GUI Screenshots}.

\noindent \textbf{Add New Questions.} 
We newly annotate questions on both existing and newly collected documents to expand the questions in our benchmark. 
Specifically, we ask annotators to add about 3 questions on existing documents, and 6 questions on newly-collected documents. Given most existing questions (even after editing) are single-page ones and sourced from texts, we put more focus on (1) cross-page and unanswerable questions and (2) questions sourced from tables, charts, and images for newly added questions to balance the distribution. We detail the quantitative requirements in Appendix~\ref{appendix: New Question Annotation}. Associated with questions, annotators also provide reference answers and meta-information (\ie evidence sources, answer format, evidence locations) for all samples. We finalized a collection of 965 samples in this procedure. The corresponding GUI is shown in Appendix~\ref{appendix: GUI Screenshots}.

\subsection{Quality Control}
\label{subsec: quality control}

Combining the merits of humans and LVLMs, we adopt a three-round, semi-automatic quality control procedure to improve the annotation quality of our benchmark. We detail each round in the following components and leave the discussion of potential bias in Appendix~\ref{appendix: potential_bias}.

\noindent \textbf{Document-relevant Detection.} Our benchmark is designed to evaluate LVLMs' long-context document understanding abilities. All questions are expected to be unanswerable without access to corresponding documents. To remove low document-relevant questions (\ie questions not relying on documents), we feed each annotated question \textbf{WITHOUT} documents to GPT-4o. A question will be identified as \textit{low document-relevant} question if GPT-4o correctly predicts under this case. Ultimately, 94 samples are identified as low document-relevant questions and removed in this round.

\noindent \textbf{Self-reflection.} We draw inspirations from MMBench~\cite{mmbench2023} and leverage LVLMs to reduce the wrongly-annotated samples. Specifically, we feed the remaining questions from the last round \textbf{WITH} their documents to GPT-4o. Samples whose model predictions are inconsistent with the reference answers are sent back to corresponding annotators. The annotators are asked to check each question and identify whether the inconsistency is caused by \textit{problematic annotation} or not. As a result, 13.8\% of the samples are identified as problematic annotations. The annotators revise them accordingly.

\noindent \textbf{Cross-checking.}
In parallel, annotators cross-check the annotated samples from other annotators and determine the inconsistency reasons the same as described above. We calculate Cohen’s kappa value of their identifications as 0.42 (17.5\% inconsistent samples), showing a moderate agreement. Regarding the 17.5\% inconsistent samples, two primary authors serve as meta-annotators and make final decisions on them (and if necessary, revise accordingly).

\begin{figure}[htbp!]
 \begin{minipage}{0.5\textwidth} 
 \centering
 \fontsize{8.2pt}{\baselineskip}\selectfont 
 \renewcommand\tabcolsep{2.2pt} 
 \renewcommand\arraystretch{0.8} 
 \begin{tabular}{lc}
 \toprule
 \textbf{Statistic} & \textbf{Number} \\
 \midrule
  \textbf{Documents} & 135 \\
  ~- Type & 7 \\
  ~- Average/Medium pages &  47.5 / 28 \\
  ~- Average/Medium length & 21,214.1 / 12,179 \\
 \midrule
  \textbf{Total questions} & 1,082 \\
  ~- Single-page question & 494 (45.7\%) \\
  ~- Cross-page questions & 365 (33.7\%)  \\
  ~- Unanswerable questions & 223 (20.6\%) \\
 \midrule
  ~- Derived questions & 184 (17.0\%) \\
  ~- Newly-annotated questions & 898 (83.0\%) \\
\midrule
  (Evidence source) \\ 
  ~- Pure-text & 305 (35.5\%) \\
  ~- Layout & 119 (13.9\%) \\
  ~- Table & 218 (25.4\%) \\
  ~- Chart & 178 (20.7\%) \\
  ~- Image & 304 (35.4\%) \\
 \midrule
(Answer Format) \\ 
  ~- String   & 250 (29.1\%) \\
  ~- Integer  & 299 (34.8\%) \\
  ~- Float & 159 (18.5\%) \\
  ~- List & 151 (17.6\%) \\
 \midrule
 Avg./Max. question length & 16.4 / 60 \\
 Avg./Max. answer length & 2.8 / 54 \\
 \bottomrule
 \end{tabular}
\captionof{table}{Dataset Statistics}
 \label{tab:statistics}
 \end{minipage} 
 \hfill
 \begin{minipage}{0.5\textwidth}
 \centering
\includegraphics[width=0.8\linewidth, bb=0 0 284 411]{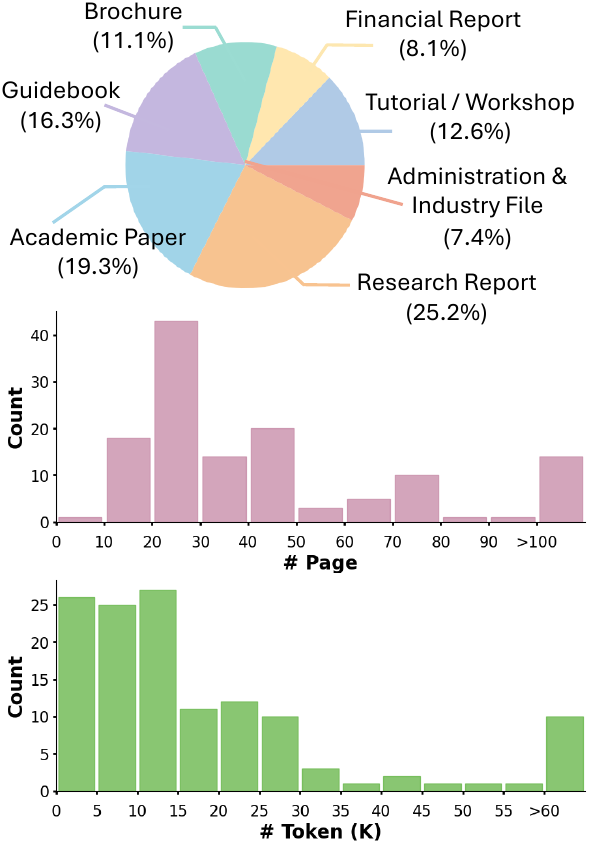}
 \caption{Detailed distribution of documents. \textbf{Top}: Document type. \textbf{Middle}: Page Number. \textbf{Bottom}: Token Number.}
 \label{fig: document_distribution}
 \end{minipage}
\end{figure}

\begin{figure}[!htbp]
    \centering
    \includegraphics[width=\linewidth]{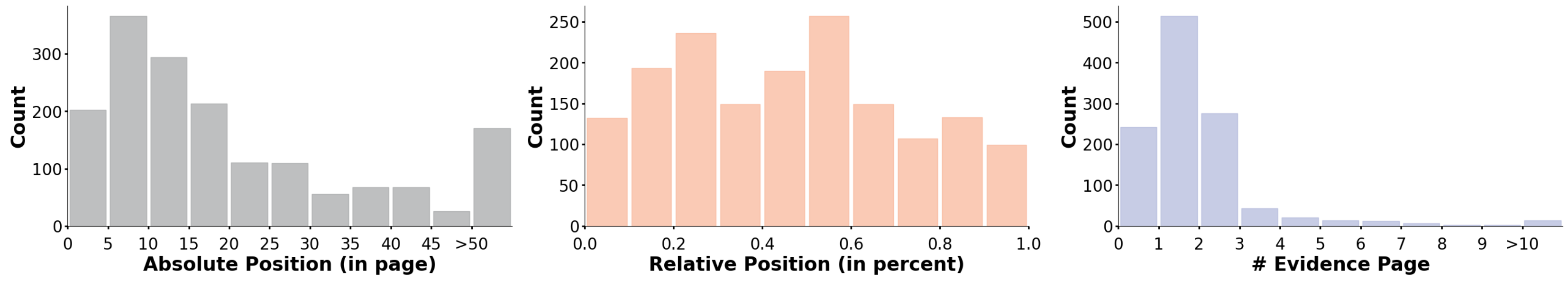}
    \caption{Detailed distribution of questions \& answers. \textbf{Left}: Absolute position of answer evidences (the page index). \textbf{Middle}: Relative position (the page index/document page number). \textbf{Right}: Evidence page number of each question. (0: unanswerable question; >2: cross-page question).}
    \label{fig: question_distribution}
\end{figure}

\subsection{Dataset Overview and Analysis}
\label{subsec: dataset_overview}

The main statistics of \benchmarkname are presented in Table~\ref{tab:statistics}. Overall, our benchmark consists of 1,082 questions. These questions are constructed upon 135 lengthy documents across 7 document types, with an average of 47.5 pages and 21,214.1 tokens. Please see detailed distributions of these documents in Figure~\ref{fig: document_distribution}. Regarding the questions, there are 494 single-page questions (1 evidence page), 365 cross-page questions (2+ evidence pages), and 223 unanswerable questions (no evidence page). These three types of questions evaluate the LVLMs's long-context DU capabilities from complementary aspects: the localization ability, the cross-page comprehension ability, and the hallucination severity, respectively. For single-page and cross-page questions, their answer evidence is scattered among different context sources (\ie text, layout, table, chart, image) and evenly distributed across different locations of the documents (see Table~\ref{tab:statistics}, Figure~\ref{fig: question_distribution} Left and Middle). Also notably, 28.6\% of cross-page questions have more than two evidence pages, which further enhances the challenge of our benchmark.

\begin{table*}[!htbp]
 \centering
 \footnotesize
 \caption{\textbf{Evaluation of various models on \benchmarkname.} We report the generalized accuracy of five types of evidence sources including pure text (TXT), layout (LAY), chart (CHA), table (TAB), and image (IMG). We also present the generalized accuracy of questions categorized by the number of evidence pages: single-page (SIN), cross-page (MUL), and unanswerable (UNA) questions. The \colorbox{firstBest}{\bf best} and \colorbox{secondBest}{\textbf{second-best}} performance in each section are highlighted.}
 \label{tab:main_results}
 \renewcommand\tabcolsep{2.5pt} 
 \renewcommand\arraystretch{1.00} 
 \resizebox{0.95\linewidth}{!}{
    \begin{tabular}{l cc|ccccc|ccc|cc}
    \toprule
     \multirow{2}{*}{\textbf{Model}} & \multirow{2}{*}{\textbf{\#Param}} & \textbf{Context} & \multicolumn{5}{c|}{\textbf{Evidence Source}} & \multicolumn{3}{c|}{\textbf{Evidence Page}} &  \multirow{2}{*}{\textbf{ACC}} & \multirow{2}{*}{\textbf{F1}} \\
     &  & \textbf{Window} & \header{TXT} & \header{LAY} & \header{CHA} & \header{TAB} & \header{FIG} & SIN & MUL & UNA \\ 
    \midrule
    \multicolumn{13}{l}{\hfill \textit{OCR (Tesseract~\cite{smith2007overview}) + Large Language Models (LLMs)} } \\
    \midrule
    \textcolor{gray}{\textit{Open-source Models}} \\
    ChatGLM-128k~\cite{bai2024longalign} & 6B & 128k & 23.4 & 12.7 & 9.7 & 10.2 & 12.2 & 18.8 & 11.5 & 18.1 & 16.3 & 14.9 \\
    Mistral-Instruct-v0.2~\cite{jiang2023mistral} & 7B & 32k & 19.9 & 13.4 & 10.2 & 10.1 & 11.0 & 16.9 & 11.3 & 24.1 & 16.4 & 13.8 \\
    Mixtral-Instruct-v0.1~\cite{jiang2024mixtral} & 8x7B & 32k & 24.2 & 14.8 & 12.5 & 15.0 & 13.7 & 21.3 & 14.1 & 13.1 & 17.0 & 16.9 \\
    Mixtral-Instruct-v0.1~\cite{jiang2024mixtral} & 8x22B & 64k & 34.2 & 21.3 & 19.5 & 21.3 & 19.2 & 27.7 & 21.9 & 32.4 & 26.9 & 24.7 
    \\
    \textcolor{gray}{\textit{Proprietary Models}} \\
    QWen-Plus~\cite{qwen1.5} & - & 32k & 17.4 & 15.6 & 7.4 & 7.9 & 8.8 & 14.2 & 10.6 & 42.2 & 18.9 & 13.4 \\
    DeepSeek-V2~\cite{deepseekv2} & - & 32k & 27.8 & 19.6 & 8.8 & 17.0 & 9.4 & 20.2 & 15.4 & 48.1 & 24.9 & 19.6 \\
    Claude-3 Opus~\cite{claude3} & - & 32k & 30.8 & \colorbox{secondBest}{30.1} & 16.4 & 24.4 & 16.3 & 32.0 & 18.6 & 30.9 & 26.9 & 24.5 \\
    Gemini-1.5-Pro~\cite{geminiteam2024gemini} & - & 32k & 29.3 & 15.9 & 12.5 & 17.7 & 11.5 & 21.2 & 16.4 & \colorbox{firstBest}{\textbf{73.4}} & 31.2 & 24.8 \\
    GPT-4-turbo~\cite{openai2024gpt4} & - & 128k & 36.5 & 21.0 & 20.7 & 24.3 & 17.3 & 28.7 & 23.8 &  31.2 & 27.6 & 25.9 \\
    GPT-4o~\cite{gpt4o} & - & 128k & \colorbox{secondBest}{41.1} & 23.4 & \colorbox{secondBest}{28.5} & \colorbox{secondBest}{38.1} & 22.4 & 35.4 & \colorbox{secondBest}{29.3} & 18.6 & 30.1 & 30.5 \\
    \midrule
    \multicolumn{13}{l}{\hfill \textit{Large Visual Language Models (LVLMs)}} \\
    \midrule
    \textcolor{gray}{\textit{Open-source, 7-14B Models}} \\
    DeepSeek-VL-Chat~\cite{lu2024deepseek} & 7.3B & 4k & 7.2 & 6.5 & 1.6 & 5.2 & 7.6 & 5.2 & 7.0 & 12.8 & 7.4 & 5.4 \\
    Idefics2~\cite{laurençon2024matters} & 8B & 8k & 9.0 & 10.6 & 4.8 & 4.1 & 8.7 & 7.7 & 7.2 & 5.0 & 7.0 & 6.8 \\
    MiniCPM-Llama3-V2.5~\cite{yu2024rlaifv,xu2024llava-uhd} & 8B & 2k & 11.9 & 10.8 & 5.1 & 5.9 & 12.2 & 9.5 & 9.5 & 4.5 & 8.5 & 8.6 \\
    InternLM-XC2-4KHD~\cite{dong2024internlm} & 8B & 16k & 9.9 & 14.3 & 7.7 & 6.3 & 13.0 & 12.6 & 7.6 & 9.6 & 10.3 & 9.8 \\
    mPLUG-DocOwl 1.5~\cite{hu2024mplug} & 8.1B & 4k & 8.2 & 8.4 & 2.0 & 3.4 & 9.9 & 7.4 & 6.4 & 6.2 & 6.9 & 6.3 \\
    Qwen-VL-Chat~\cite{bai2023qwenvl} & 9.6B & 6k & 5.5 & 9.0 & 5.4 & 2.2 & 6.9 & 5.2 & 7.1 & 6.2 & 6.1 & 5.4 \\
    Monkey-Chat~\cite{li2023monkey} & 9.8B & 2k & 6.8 & 7.2 & 3.6 & 6.7 & 9.4 & 6.6 & 6.2 & 6.2 & 6.2 & 5.6 \\
    \textcolor{gray}{\textit{Open-source, >14B Models}} \\
    CogVLM2-LLaMA3-Chat~\cite{wang2023cogvlm} & 19B & 8k & 3.7 & 2.7 & 6.0 & 3.2 & 6.9 & 3.9 & 5.3 & 3.7 & 4.4 & 4.0 \\
    InternVL-Chat-v1.5~\cite{chen2024far} & 26B & 4k & 14.0 & 16.2 & 7.1 & 10.1 & 16.6 & 14.9 & 12.2 & 17.5 & 14.6 & 13.0 \\
    EMU2-Chat~\cite{sun2023generative} & 37B & 2k & 6.1 & 9.7 & 2.6 & 3.8 & 7.7 & 5.7 & 6.1 & 16.5 & 8.3 & 5.5 \\
    \textcolor{gray}{\textit{Proprietary Models}} \\
    Claude-3 Opus~\cite{claude3} & - & 200k & 24.9 & 24.7 & 14.8 & 13.0 & 17.1 & 25.6 & 13.8  & 7.6 & 17.4 & 18.1 \\
    Gemini-1.5-Pro~\cite{geminiteam2024gemini} & - & 128k & 21.0 & 17.6 & 6.9 & 14.5 & 15.2 & 21.1 & 11.1 &  \colorbox{secondBest}{69.2} & 28.2 & 20.6 \\
    GPT-4V(ision)~\cite{openai2024gpt4} & - & 128k & 34.4 & 28.3 & 28.2 & 32.4 & \colorbox{secondBest}{26.8} & \colorbox{secondBest}{36.4} & 27.0 & 31.2 & \colorbox{secondBest}{32.4} & \colorbox{secondBest}{31.2} \\
    GPT-4o~\cite{gpt4o} & - & 128k & \colorbox{firstBest}{\textbf{46.3}} & \colorbox{firstBest}{\textbf{46.0}} & \colorbox{firstBest}{\textbf{45.3}} & \colorbox{firstBest}{\textbf{50.0}} & \colorbox{firstBest}{\textbf{44.1}} & \colorbox{firstBest}{\textbf{54.5}} & \colorbox{firstBest}{\textbf{41.5}} & 20.2 & \colorbox{firstBest}{\textbf{42.8}} & \colorbox{firstBest}{\textbf{44.9}} \\
    \bottomrule
    \multicolumn{13}{l}{\hfill \textit{Human Baseline}} \\
    \midrule
    Human Experts & - & -  & - & -  & -  & -  & -  &-  & -  & - & 65.8 & 66.0 \\
    \bottomrule
    \end{tabular}
    }
\end{table*}

\section{Evaluation}\label{sec:evaluation}

\subsection{Evaluation Protocol}
We follow MATHVISTA~\cite{lu2024mathvista} to conduct a three-step evaluation protocol: \textit{response generation}, \textit{answer extraction}, and \textit{score calculation}. We adopt such a protocol out of three considerations: (1) Current LVLMs are instructed to generate long responses, rather than short-form answers, in conventional settings. (2) The evaluation of long responses, however, remains an open and challenging problem. (3) We focus on the document understanding (not instruction following) abilities of LVLMs.

Specifically, we impose no limitations on \textit{response generation} stage to encourage LVLMs to answer the questions in a freestyle. Then we propose a unified LLM-based \textit{answer extractor} (GPT-4o under our setting) to convert their long responses to short-form answers. Finally, we use a rule-based \textit{score calculator} to evaluate the converted short answers. We report both generalized accuracy and generalized F1 score to balance the answerable (positive) and unanswerable (negative) questions. The used prompt, the high correlation between our automatic \textit{answer extractor} and human evaluation, and the detailed rules of our \textit{score calculation} are described in Appendix~\ref{appendix: experimental_details}.

\subsection{Experimental Setup}
We evaluate 14 LVLMs on \benchmarkname, including 4 proprietary LVLMs and 10 open-source LVLMs. To purely evaluate LVLMs' long-context DU abilities, we screenshot each page of the PDF-formatted document with 144 DPI and feed all these PNG-formatted images to LVLMs in an end-to-end approach. Notably, all evaluated open-source LVLMs do not support multi-image inputs or present significant performance drops when fed with excessive images (\eg more than 10 or 20 images). Therefore, we employ a concatenation strategy that combines all screenshot pages into 1 or 5 images and feeds these concatenated images to open-source LVLMs. Regarding proprietary LVLMs, we adopt the same concatenation strategy and reduce the image number to 20 for Claude-3-Opus to fit its maximum image threshold. For GPT-4o, GPT-4V, and Gemini-1.5-Pro, we directly send all original screenshots to them (\ie the image number equals the page number).

For comparison, we also use the Tesseract~\cite{smith2007overview} OCR model to recognize and extract texts from the documents and feed the parsed documents to 10 LLMs, including 6 proprietary and 4 open-source ones. Texts exceeding their context lengths are truncated. Notably, as a key component of the classical solution for the DU task, the OCR model can handle most flattened texts and some structured tables in the document. However, it cannot perceive the information from the charts or images. Thus the TXT-formatted, OCR-parsed documents are lossy documents in which the information is not fully preserved. More detailed hyperparameters are introduced in Appendix~\ref{appendix: Hyperparameters}. Additionally, we also conduct manual evaluation on a subset of our datasets (238 questions from 29 documents) to indicate the difficulty of this task for humans.

\subsection{Main Results}
We compare the performance of different LVLMs and LLMs in Table~\ref{tab:main_results}, reporting their generalized accuracy and F1 scores (shown in the last two columns). Regarding LVLMs, we draw several conclusions as below: (1) The performance demonstrates that long-context DU is still a challenging and unsolved task for current LVLMs. The best-performing LVLM, GPT-4o, merely achieves a 44.9\% F1 score. The second best-performing LVLM, GPT-4V, lags behind by over 10\% percent and presents a 31.4\% F1 score. All other LVLMs only achieve about 20\% or even lower F1 scores. (2) Though far from satisfactory, GPT-4o performs much better than all other models (including GPT-4V). Thus we speculate that the multi-modal pre-training paradigm significantly benefits LVLMs' cross-modality understanding capabilities. (3) Proprietary LVLMs perform better than open-source LVLMs by a large margin. We attribute it to the difference of acceptable image numbers: open-source LVLMs only support single-image or several-image
inputs, while proprietary LVLMs can be fed with at least 20 images or even more. Given that lengthy documents have tens of even hundreds of pages, it is impractical for open-source LVLMs to accurately perceive the information in the documents from the excessively concatenated images. (4) The performances of different models are highly correlated with their acceptable image numbers and maximum image resolutions. Notably, open-source LVLMs that support high-resolution images (\ie InternLM-XC2-4KHD and InternVL-Chat-v1.5) exhibit superior performance compared to those with lower resolution limits.

Surprisingly, LVLMs even demonstrate overall worse performance than LLMs, even LLMs are fed with lossy OCR-parsed documents. Specifically, Gemini-1.5-Pro and Claude-3 Opus have 4.2\% and 6.4\% absolute F1-score degradations on vision versions. And the best-performing LLM (Mixtral) also surpasses the best-performing LVLM (InternVL-v1.5) by 11.7\%. The above results clearly reveal that most current LVLMs are still not proficient in cross-modality, long-context document understandings. It is promising that GPT-4o and GPT-4-turbo achieve better performance when seeing multi-modality PDF documents than parsed text by 14.4\% and 5.3\% F1-score, respectively. Their performances validate the feasibility, benefit, and necessity of understanding documents in an end-to-end, cross-modality approach. We speculate that the scarce related pre-training corpus (\ie extremely multi-image or lengthy documents) hinders the long-context DU capabilities of other LVLMs. We will leave related explorations for future work.

Regarding the human evaluation, we observe 66.0\% F1-score from our annotators and a significant performance gap (exceeding 20\% in absolute) between the current LVLMs and humans. This gap highlights the challenges of document understanding for LVLMs and the necessity of our benchmark.

\subsection{Fine-grained Results.} 

\begin{figure}
    \centering
    \includegraphics[width=0.8\textwidth]{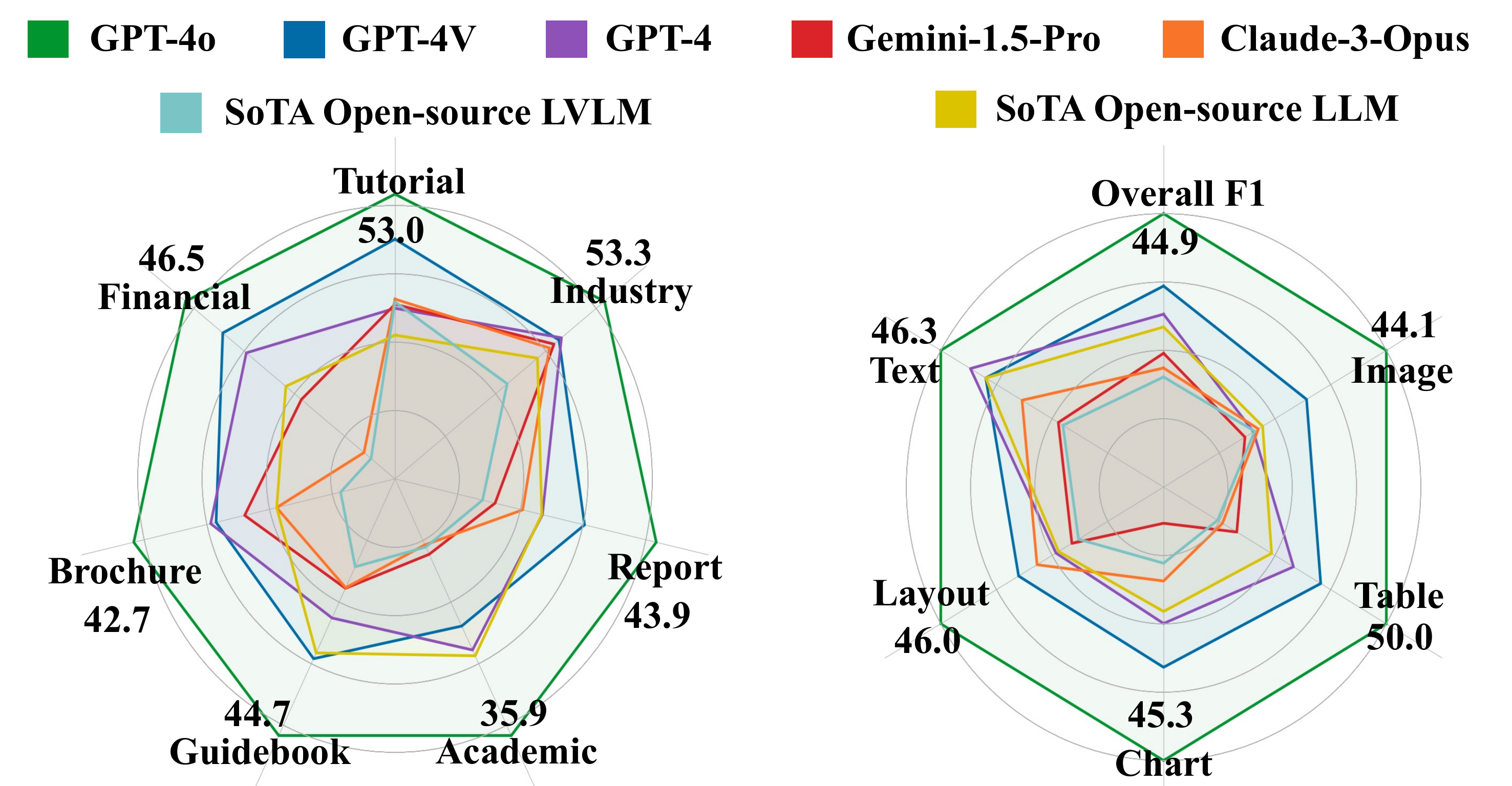}
    \caption{Fine-grained results on various document types and evidence sources.}
    \label{fig: radar_plot}
\end{figure}

\noindent{\textbf{Document Type.}} As illustrated in Figure~\ref{fig: radar_plot}, LVLMs and LLMs exhibit distinct performance patterns across various document types. Our findings include: (1) All evaluated models demonstrate decent performance on industrial documents, which tend to have more standardized formats and less non-textual information. (2) The GPT series and Mixtral (\ie the SoTA open-source LLM) show relatively balanced performance across different document types. In contrast, other models perform significantly worse in specialized domains such as academic papers and financial reports. (3) When equipped with OCR, LLM-based models like GPT-4 and Mixtral achieve comparable or even superior performance on industrial documents, academic papers, and brochures. Conversely, end-to-end LVLMs outperform OCR+LLMs in areas such as tutorials, research reports, and guidelines. We speculate that comprehending these latter document types requires more extensive multi-modal information, from which LVLMs significantly benefit.

\noindent{\textbf{Evidence Source.}} 
We categorize questions based on their evidence sources and present fine-grained results in Figure~\ref{fig: radar_plot} and Table~\ref{tab:main_results}. Our observations reveal that only GPT-4o exhibits relatively balanced performance across the different sources. Other LVLMs, however, show inferior performance on questions related to charts and/or images compared to those related to text and/or layout. Additionally, LLMs generally demonstrate better or comparable performance to LVLMs on text- and table-related questions but show worse performance on questions involving other elements. This highlights the limitations of OCR (and other PDF parsers) when dealing with charts and images, as well as the gap in OCR capabilities between LVLMs and pure-text LLMs.

\begin{wrapfigure}{r}{0.42\textwidth}
    \centering
    \includegraphics[width=\linewidth, bb=0 0 680 496]{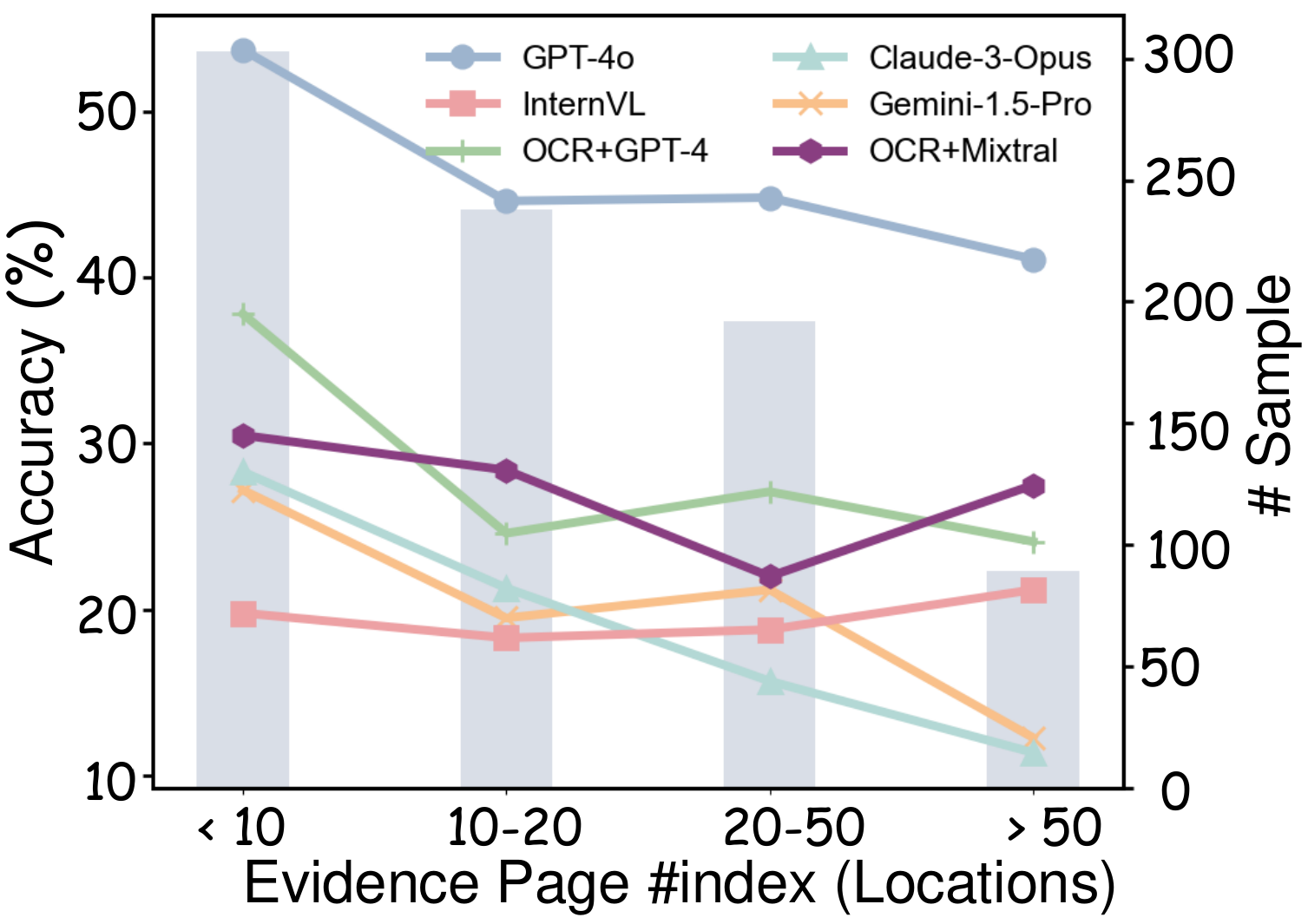}
    \caption{Relationships between evidence positions and model performances.}
    \label{fig:evidence_page}
\end{wrapfigure}

\noindent{\textbf{Evidence Position.}}
We also examine how the evidence locations (\ie the page indexes where the answer evidence is found) affect model performance. The results shown in Figure~\ref{fig:evidence_page} reinforce that \benchmarkname poses significant challenges for current models, at least partially due to the extended length of the documents. Almost all models (except InternVL-v1.5) exhibit their best performance on questions derived from the initial pages, while their performance declines progressively as the page index increases. Interestingly, two proprietary models, Gemini-Pro-1.5 and Claude-3-Opus, experience particularly sharp declines in performance.

\noindent{\textbf{Number of Evidence Page.}} 
We observe a consistent trend that all models achieve higher scores on single-page questions than cross-page questions. It reveals that gathering and reasoning over all necessary information across different pages is not trivial for current LVLMs and LLMs. More interestingly, evaluated LVLMs behave differently on unanswerable questions. GPT-4o and Claude-3 Opus adopt more aggressive strategies and usually tend to provide some answers. It makes their answers more likely helpful, but also increases the risk of hallucination and unfaithfulness (see their scores on unanswerable questions are much lower than answerable questions). On the contrary, Gemini-1.5-Pro, DeepSeek-VL-Chat, and EMU2-Chat are much more cautious and tend to refuse to answer questions about which they are uncertain. It makes their answers safer but less helpful (with large amounts of responses like \textit{I don't know}).
\section{Analysis \& Discussion}
\subsection{Oracle Setting}
\begin{figure}[!htbp]
    \centering
    \includegraphics[width=\linewidth, bb=0 0 1219 312]{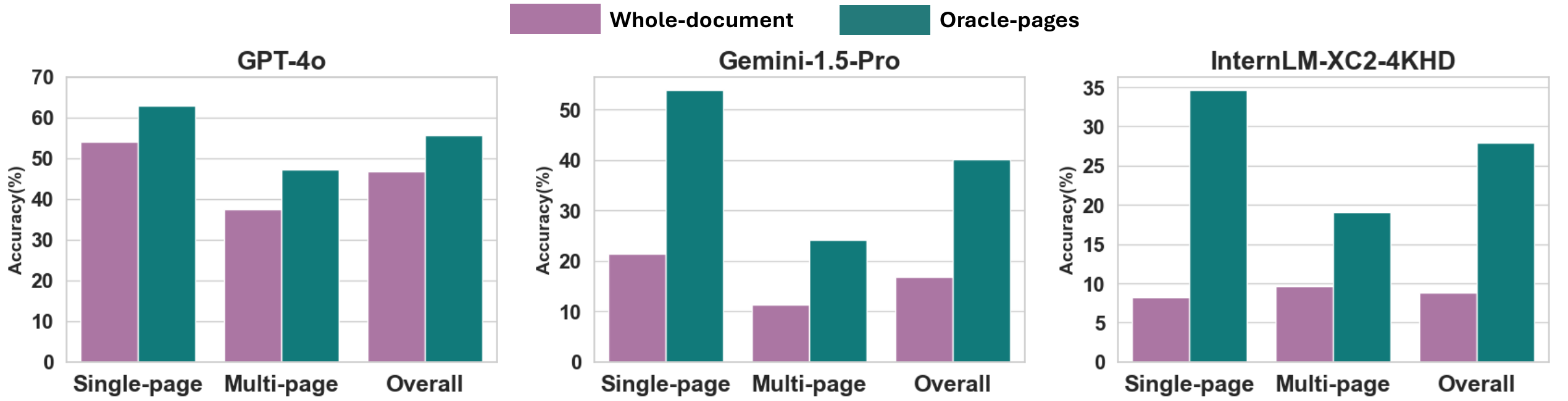}
    \caption{Performance comparisons between normal setting (feeding models with the whole documents) and oracle setting (feeding models only with the evidence pages) among three LVLMs.}
    \label{fig: oracle_setting}
\end{figure}

We conduct additional experiments to explore to what extent the challenges of \benchmarkname are caused by the long-context lengths of documents. Specifically, we feed 820 answerable questions along with their oracle evidence pages (instead of the whole documents) to three representative LVLMs and show results in Figure~\ref{fig: oracle_setting}. On one hand, it indicates that long-context length is a significantly challenging factor for document understanding. Compared with the oracle-page setting, lengthy documents lead to more than 20\% absolute performance degradation on Gemini-1.5-Pro and InternLM-XC2-4KHD. Regarding the single-page questions, the performance difference even achieves up to 30\%. On the other hand, the overall performance achieves only about 40\% and 30\% for Gemini-1.5-Pro and InternLM-XC2-4KHD even under oracle-page setting. And the improvement for GPT-4o is much less (about 10\%). It demonstrates that the development of long-context LVLMs can largely facilitate, though still can not fully solve, the long-context DU task.

\begin{wrapfigure}{l}{0.33\textwidth}
    \centering
    \includegraphics[width=\linewidth, bb=0 0 215 198]{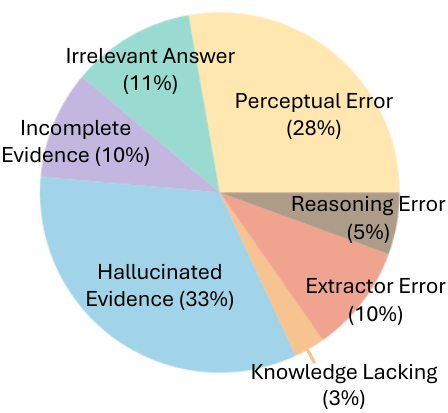}
    \caption{Error distribution}
    \label{fig:error_analysis}
\end{wrapfigure}

\subsection{Error Analysis}
We further conduct error analysis to understand the bottleneck of current LVLMs in a qualitative approach. Specifically, we randomly select 72 error predictions from GPT-4o's responses and manually check their error reasons. These errors are categorized into seven types: \textit{Perceptual Error}, \textit{Irrelevant Answer}, \textit{Incomplete Evidence}, \textit{Hallucinated Evidence}, \textit{Extractor Error}, \textit{Reasoning Error} and \textit{Knowledge Lacking}.
The distribution of these errors is illustrated in Figure~\ref{fig:error_analysis}. It indicates that most errors come from the model's hallucination (\ie wrong explanations and answers to unanswerable questions) and perceptual errors (mainly in visual contexts). Additionally, GPT-4o sometimes misunderstands the intent of questions and provides irrelevant responses. The errors caused by collecting incomplete evidence (for cross-page questions) are also unignorable. The descriptions and examples of these error types are detailed in Appendix~\ref{appendix: error_analysis}.



\section{Conclusion}
\label{sec:conclusion}
In this work, we present \benchmarkname to evaluate the long-context DU capabilities of LVLMs. Extensive experiments on 14 LVLMs (and 10 LLMs for comparison) reveal that the understanding of lengthy documents poses great challenges to current LVLMs. Even though the performance of GPT-4o proves the benefit of end-to-end, multi-modality perception for DU tasks, most LVLMs struggle on long visual contexts (\ie extremely multiple images) and show inferior performance compared to OCR+LLM pipelines. We hope that the construction of our benchmark could push forward the development of more powerful LVLMs on lengthy document understanding.

\section*{Acknowledgements}
This study is supported under the RIE2020 Industry Alignment Fund – Industry Collaboration Projects (IAF-ICP) Funding Initiative, as well as cash and in-kind contribution from the industry partner(s). This work is also supported by Shanghai Artificial Intelligence Laboratory, the National Key R\&D Program of China (2022ZD0160201).

\newpage
{\small
\bibliographystyle{unsrt}
\bibliography{reference}

\begin{thebibliography}{10}

\bibitem{Bornmann2014GrowthRO}
Lutz Bornmann and R{\"u}diger Mutz.
\newblock Growth rates of modern science: A bibliometric analysis based on the number of publications and cited references.
\newblock {\em Journal of the Association for Information Science and Technology}, 66, 2014.

\bibitem{gpt4o}
Open AI.
\newblock Hello gpt-4o, 2024.

\bibitem{geminiteam2024gemini}
Gemini Team.
\newblock {Gemini 1.5}: Unlocking multimodal understanding across millions of tokens of context, 2024.

\bibitem{claude3}
Anthropic.
\newblock Introducing the next generation of claude, 2024.

\bibitem{dong2024internlm}
Xiaoyi Dong, Pan Zhang, Yuhang Zang, Yuhang Cao, Bin Wang, Linke Ouyang, Songyang Zhang, Haodong Duan, Wenwei Zhang, Yining Li, et~al.
\newblock {Internlm-Xcomposer2-4KHD}: A pioneering large vision-language model handling resolutions from 336 pixels to 4k hd.
\newblock {\em ArXiv preprint}, abs/2404.06512, 2024.

\bibitem{chen2024far}
Zhe Chen, Weiyun Wang, Hao Tian, Shenglong Ye, Zhangwei Gao, Erfei Cui, Wenwen Tong, Kongzhi Hu, Jiapeng Luo, Zheng Ma, et~al.
\newblock How far are we to gpt-4v? closing the gap to commercial multimodal models with open-source suites.
\newblock {\em ArXiv preprint}, abs/2404.16821, 2024.

\bibitem{li2023otter}
Bo~Li, Yuanhan Zhang, Liangyu Chen, Jinghao Wang, Jingkang Yang, and Ziwei Liu.
\newblock Otter: A multi-modal model with in-context instruction tuning, 2023.

\bibitem{li2024llavanext-strong}
Bo~Li, Kaichen Zhang, Hao Zhang, Dong Guo, Renrui Zhang, Feng Li, Yuanhan Zhang, Ziwei Liu, and Chunyuan Li.
\newblock {LLaVA-NeXT}: Stronger llms supercharge multimodal capabilities in the wild, 2024.

\bibitem{wang2023cogvlm}
Weihan Wang, Qingsong Lv, Wenmeng Yu, Wenyi Hong, Ji~Qi, Yan Wang, Junhui Ji, Zhuoyi Yang, Lei Zhao, Xixuan Song, Jiazheng Xu, Bin Xu, Juanzi Li, Yuxiao Dong, Ming Ding, and Jie Tang.
\newblock {CogVLM}: Visual expert for pretrained language models, 2023.

\bibitem{hu2024mplugdocowl}
Anwen Hu, Haiyang Xu, Jiabo Ye, Ming Yan, Liang Zhang, Bo~Zhang, Chen Li, Ji~Zhang, Qin Jin, Fei Huang, and Jingren Zhou.
\newblock mplug-docowl 1.5: Unified structure learning for ocr-free document understanding, 2024.

\bibitem{liu2024textmonkey}
Yuliang Liu, Biao Yang, Qiang Liu, Zhang Li, Zhiyin Ma, Shuo Zhang, and Xiang Bai.
\newblock Textmonkey: An ocr-free large multimodal model for understanding document, 2024.

\bibitem{Mathew2020DocVQAAD}
Minesh Mathew, Dimosthenis Karatzas, R.~Manmatha, and C.~V. Jawahar.
\newblock Docvqa: A dataset for vqa on document images.
\newblock {\em 2021 IEEE Winter Conference on Applications of Computer Vision (WACV)}, pages 2199--2208, 2020.

\bibitem{masry-etal-2022-chartqa}
Ahmed Masry, Xuan~Long Do, Jia~Qing Tan, Shafiq Joty, and Enamul Hoque.
\newblock {C}hart{QA}: A benchmark for question answering about charts with visual and logical reasoning.
\newblock In {\em Findings of the Association for Computational Linguistics: ACL 2022}, pages 2263--2279, Dublin, Ireland, 2022. Association for Computational Linguistics.

\bibitem{Mathew2021InfographicVQA}
Minesh Mathew, Viraj Bagal, Rub{\`e}n~P{\'e}rez Tito, Dimosthenis Karatzas, Ernest Valveny, and C.V. Jawahar.
\newblock Infographicvqa.
\newblock {\em 2022 IEEE/CVF Winter Conference on Applications of Computer Vision (WACV)}, pages 2582--2591, 2021.

\bibitem{zhu2022towards}
Fengbin Zhu, Wenqiang Lei, Fuli Feng, Chao Wang, Haozhou Zhang, and Tat-Seng Chua.
\newblock Towards complex document understanding by discrete reasoning.
\newblock In {\em Proceedings of the 30th ACM International Conference on Multimedia}, pages 4857--4866, 2022.

\bibitem{tito2023hierarchical}
Rubèn Tito, Dimosthenis Karatzas, and Ernest Valveny.
\newblock Hierarchical multimodal transformers for multi-page docvqa, 2023.

\bibitem{VanLandeghem2023DocumentUD}
Jordy~Van Landeghem, Rub{\`e}n~P{\'e}rez Tito, Łukasz Borchmann, Michal Pietruszka, Pawel J'oziak, Rafal Powalski, Dawid Jurkiewicz, Micka{\"e}l Coustaty, Bertrand Ackaert, Ernest Valveny, Matthew~B. Blaschko, Sien Moens, and Tomasz Stanislawek.
\newblock Document understanding dataset and evaluation ({DUDE}).
\newblock In {\em ICCV}, 2023.

\bibitem{SlideVQA2023}
Ryota Tanaka, Kyosuke Nishida, Kosuke Nishida, Taku Hasegawa, Itsumi Saito, and Kuniko Saito.
\newblock {SlideVQA}: A dataset for document visual question answering on multiple images.
\newblock In {\em AAAI}, 2023.

\bibitem{islam2023financebench}
Pranab Islam, Anand Kannappan, Douwe Kiela, Rebecca Qian, Nino Scherrer, and Bertie Vidgen.
\newblock {FinanceBench}: A new benchmark for financial question answering, 2023.

\bibitem{Borchmann2021DUEED}
Łukasz Borchmann, Michal Pietruszka, Tomasz Stanislawek, Dawid Jurkiewicz, Michał Turski, Karolina Szyndler, and Filip Gralinski.
\newblock Due: End-to-end document understanding benchmark.
\newblock In {\em NeurIPS Datasets and Benchmarks}, 2021.

\bibitem{liu2024visualwebbench}
Junpeng Liu, Yifan Song, Bill~Yuchen Lin, Wai Lam, Graham Neubig, Yuanzhi Li, and Xiang Yue.
\newblock {VisualWebBench}: How far have multimodal llms evolved in web page understanding and grounding?, 2024.

\bibitem{kardas-etal-2020-axcell}
Marcin Kardas, Piotr Czapla, Pontus Stenetorp, Sebastian Ruder, Sebastian Riedel, Ross Taylor, and Robert Stojnic.
\newblock {AxCell}: Automatic extraction of results from machine learning papers.
\newblock In {\em Proceedings of the 2020 Conference on Empirical Methods in Natural Language Processing (EMNLP)}, pages 8580--8594, Online, 2020. Association for Computational Linguistics.

\bibitem{saadfalcon2023pdftriage}
Jon Saad-Falcon, Joe Barrow, Alexa Siu, Ani Nenkova, David~Seunghyun Yoon, Ryan~A. Rossi, and Franck Dernoncourt.
\newblock Pdftriage: Question answering over long, structured documents, 2023.

\bibitem{yang-etal-2018-hotpotqa}
Zhilin Yang, Peng Qi, Saizheng Zhang, Yoshua Bengio, William Cohen, Ruslan Salakhutdinov, and Christopher~D. Manning.
\newblock {H}otpot{QA}: A dataset for diverse, explainable multi-hop question answering.
\newblock In Ellen Riloff, David Chiang, Julia Hockenmaier, and Jun{'}ichi Tsujii, editors, {\em Proceedings of the 2018 Conference on Empirical Methods in Natural Language Processing}, pages 2369--2380, Brussels, Belgium, October-November 2018. Association for Computational Linguistics.

\bibitem{LayoutLMv1}
Yiheng Xu, Minghao Li, Lei Cui, Shaohan Huang, Furu Wei, and Ming Zhou.
\newblock Layoutlm: Pre-training of text and layout for document image understanding.
\newblock In Rajesh Gupta, Yan Liu, Jiliang Tang, and B.~Aditya Prakash, editors, {\em {KDD} '20: The 26th {ACM} {SIGKDD} Conference on Knowledge Discovery and Data Mining, Virtual Event, CA, USA, August 23-27, 2020}, pages 1192--1200. {ACM}, 2020.

\bibitem{xu-etal-2021-layoutlmv2}
Yang Xu, Yiheng Xu, Tengchao Lv, Lei Cui, Furu Wei, Guoxin Wang, Yijuan Lu, Dinei Florencio, Cha Zhang, Wanxiang Che, Min Zhang, and Lidong Zhou.
\newblock {L}ayout{LM}v2: Multi-modal pre-training for visually-rich document understanding.
\newblock In {\em Proceedings of the 59th Annual Meeting of the Association for Computational Linguistics and the 11th International Joint Conference on Natural Language Processing (Volume 1: Long Papers)}, pages 2579--2591, Online, 2021. Association for Computational Linguistics.

\bibitem{LayoutLMv3}
Yupan Huang, Tengchao Lv, Lei Cui, Yutong Lu, and Furu Wei.
\newblock Layoutlmv3: Pre-training for document ai with unified text and image masking.
\newblock In {\em Proceedings of the 30th ACM International Conference on Multimedia}, MM '22, page 4083–4091, New York, NY, USA, 2022. Association for Computing Machinery.

\bibitem{kim2022donut}
Geewook Kim, Teakgyu Hong, Moonbin Yim, JeongYeon Nam, Jinyoung Park, Jinyeong Yim, Wonseok Hwang, Sangdoo Yun, Dongyoon Han, and Seunghyun Park.
\newblock Ocr-free document understanding transformer.
\newblock In {\em European Conference on Computer Vision (ECCV)}, 2022.

\bibitem{Pix2Struct}
Kenton Lee, Mandar Joshi, Iulia Turc, Hexiang Hu, Fangyu Liu, Julian Eisenschlos, Urvashi Khandelwal, Peter Shaw, Ming-Wei Chang, and Kristina Toutanova.
\newblock Pix2struct: screenshot parsing as pretraining for visual language understanding.
\newblock In {\em Proceedings of the 40th International Conference on Machine Learning}, ICML'23. JMLR.org, 2023.

\bibitem{shaham-etal-2022-scrolls}
Uri Shaham, Elad Segal, Maor Ivgi, Avia Efrat, Ori Yoran, Adi Haviv, Ankit Gupta, Wenhan Xiong, Mor Geva, Jonathan Berant, and Omer Levy.
\newblock {SCROLLS}: Standardized {C}ompa{R}ison over long language sequences.
\newblock In {\em Proceedings of the 2022 Conference on Empirical Methods in Natural Language Processing}, pages 12007--12021, Abu Dhabi, United Arab Emirates, 2022. Association for Computational Linguistics.

\bibitem{an2023leval}
Chenxin An, Shansan Gong, Ming Zhong, Xingjian Zhao, Mukai Li, Jun Zhang, Lingpeng Kong, and Xipeng Qiu.
\newblock L-eval: Instituting standardized evaluation for long context language models, 2023.

\bibitem{bai2023longbench}
Yushi Bai, Xin Lv, Jiajie Zhang, Hongchang Lyu, Jiankai Tang, Zhidian Huang, Zhengxiao Du, Xiao Liu, Aohan Zeng, Lei Hou, Yuxiao Dong, Jie Tang, and Juanzi Li.
\newblock {LongBench}: A bilingual, multitask benchmark for long context understanding.
\newblock {\em ArXiv preprint}, abs/2308.14508, 2023.

\bibitem{zhang2024inftybench}
Xinrong Zhang, Yingfa Chen, Shengding Hu, Zihang Xu, Junhao Chen, Moo~Khai Hao, Xu~Han, Zhen~Leng Thai, Shuo Wang, Zhiyuan Liu, and Maosong Sun.
\newblock $\infty$bench: Extending long context evaluation beyond 100k tokens, 2024.

\bibitem{Tworkowski2023FocusedTC}
Szymon Tworkowski, Konrad Staniszewski, Mikolaj Pacek, Yuhuai Wu, Henryk Michalewski, and Piotr Milo's.
\newblock Focused transformer: Contrastive training for context scaling.
\newblock {\em ArXiv preprint}, abs/2307.03170, 2023.

\bibitem{Chen2023LongLoRAEF}
Yukang Chen, Shengju Qian, Haotian Tang, Xin Lai, Zhijian Liu, Song Han, and Jiaya Jia.
\newblock Longlora: Efficient fine-tuning of long-context large language models.
\newblock {\em ArXiv preprint}, abs/2309.12307, 2023.

\bibitem{Peng2023YaRNEC}
Bowen Peng, Jeffrey Quesnelle, Honglu Fan, and Enrico Shippole.
\newblock Yarn: Efficient context window extension of large language models.
\newblock {\em ArXiv preprint}, abs/2309.00071, 2023.

\bibitem{bai2024longalign}
Yushi Bai, Xin Lv, Jiajie Zhang, Yuze He, Ji~Qi, Lei Hou, Jie Tang, Yuxiao Dong, and Juanzi Li.
\newblock {LongAlign}: A recipe for long context alignment of large language models.
\newblock {\em ArXiv preprint}, abs/2401.18058, 2024.

\bibitem{song2024milebench}
Dingjie Song, Shunian Chen, Guiming~Hardy Chen, Fei Yu, Xiang Wan, and Benyou Wang.
\newblock Milebench: Benchmarking mllms in long context.
\newblock {\em ArXiv preprint}, abs/2404.18532, 2024.

\bibitem{jiang2024mantis}
Dongfu Jiang, Xuan He, Huaye Zeng, Cong Wei, Max Ku, Qian Liu, and Wenhu Chen.
\newblock Mantis: Interleaved multi-image instruction tuning, 2024.

\bibitem{lu2024text}
Yujie Lu, Xiujun Li, Tsu-Jui Fu, Miguel Eckstein, and William~Yang Wang.
\newblock From text to pixel: Advancing long-context understanding in mllms, 2024.

\bibitem{mmbench2023}
Yuan Liu, Haodong Duan, Yuanhan Zhang, Bo~Li, Songyang Zhang, Wangbo Zhao, Yike Yuan, Jiaqi Wang, Conghui He, Ziwei Liu, et~al.
\newblock {MMbench}: Is your multi-modal model an all-around player?
\newblock {\em ArXiv preprint}, abs/2307.06281, 2023.

\bibitem{smith2007overview}
Ray Smith.
\newblock An overview of the tesseract ocr engine.
\newblock In {\em ICDAR}, 2007.

\bibitem{jiang2023mistral}
Albert~Q. Jiang, Alexandre Sablayrolles, Arthur Mensch, Chris Bamford, Devendra~Singh Chaplot, Diego de~las Casas, Florian Bressand, Gianna Lengyel, Guillaume Lample, Lucile Saulnier, Lélio~Renard Lavaud, Marie-Anne Lachaux, Pierre Stock, Teven~Le Scao, Thibaut Lavril, Thomas Wang, Timothée Lacroix, and William~El Sayed.
\newblock Mistral 7b, 2023.

\bibitem{jiang2024mixtral}
Albert~Q. Jiang, Alexandre Sablayrolles, Antoine Roux, Arthur Mensch, Blanche Savary, Chris Bamford, Devendra~Singh Chaplot, Diego de~las Casas, Emma~Bou Hanna, Florian Bressand, Gianna Lengyel, Guillaume Bour, Guillaume Lample, Lélio~Renard Lavaud, Lucile Saulnier, Marie-Anne Lachaux, Pierre Stock, Sandeep Subramanian, Sophia Yang, Szymon Antoniak, Teven~Le Scao, Théophile Gervet, Thibaut Lavril, Thomas Wang, Timothée Lacroix, and William~El Sayed.
\newblock Mixtral of experts, 2024.

\bibitem{qwen1.5}
Qwen Team.
\newblock Introducing qwen1.5, 2024.

\bibitem{deepseekv2}
DeepSeek-AI.
\newblock {DeepSeek-V2}: A strong, economical, and efficient mixture-of-experts language model, 2024.

\bibitem{openai2024gpt4}
OpenAI.
\newblock {GPT-4} technical report, 2024.

\bibitem{lu2024deepseek}
Haoyu Lu, Wen Liu, Bo~Zhang, Bingxuan Wang, Kai Dong, Bo~Liu, Jingxiang Sun, Tongzheng Ren, Zhuoshu Li, Yaofeng Sun, et~al.
\newblock {DeepSeek-VL}: towards real-world vision-language understanding.
\newblock {\em ArXiv preprint}, abs/2403.05525, 2024.

\bibitem{laurençon2024matters}
Hugo Laurençon, Léo Tronchon, Matthieu Cord, and Victor Sanh.
\newblock What matters when building vision-language models?, 2024.

\bibitem{yu2024rlaifv}
Tianyu Yu, Haoye Zhang, Yuan Yao, Yunkai Dang, Da~Chen, Xiaoman Lu, Ganqu Cui, Taiwen He, Zhiyuan Liu, Tat-Seng Chua, and Maosong Sun.
\newblock {RLAIF-V}: Aligning mllms through open-source ai feedback for super gpt-4v trustworthiness.
\newblock {\em ArXiv preprint}, abs/2405.17220, 2024.

\bibitem{xu2024llava-uhd}
Ruyi Xu, Yuan Yao, Zonghao Guo, Junbo Cui, Zanlin Ni, Chunjiang Ge, Tat-Seng Chua, Zhiyuan Liu, and Gao Huang.
\newblock {LLaVA-UHD}: an lmm perceiving any aspect ratio and high-resolution images.
\newblock {\em ArXiv preprint}, abs/2403.11703, 2024.

\bibitem{hu2024mplug}
Anwen Hu, Haiyang Xu, Jiabo Ye, Ming Yan, Liang Zhang, Bo~Zhang, Chen Li, Ji~Zhang, Qin Jin, Fei Huang, et~al.
\newblock {mPLUG-DocOwl} 1.5: Unified structure learning for ocr-free document understanding.
\newblock {\em ArXiv preprint}, abs/2403.12895, 2024.

\bibitem{bai2023qwenvl}
Jinze Bai, Shuai Bai, Shusheng Yang, Shijie Wang, Sinan Tan, Peng Wang, Junyang Lin, Chang Zhou, and Jingren Zhou.
\newblock {Qwen-VL}: A frontier large vision-language model with versatile abilities.
\newblock {\em ArXiv preprint}, abs/2308.12966, 2023.

\bibitem{li2023monkey}
Zhang Li, Biao Yang, Qiang Liu, Zhiyin Ma, Shuo Zhang, Jingxu Yang, Yabo Sun, Yuliang Liu, and Xiang Bai.
\newblock {Monkey}: Image resolution and text label are important things for large multi-modal models.
\newblock {\em ArXiv preprint}, abs/2311.06607, 2023.

\bibitem{sun2023generative}
Quan Sun, Yufeng Cui, Xiaosong Zhang, Fan Zhang, Qiying Yu, Zhengxiong Luo, Yueze Wang, Yongming Rao, Jingjing Liu, Tiejun Huang, et~al.
\newblock Generative multimodal models are in-context learners.
\newblock {\em ArXiv preprint}, abs/2312.13286, 2023.

\bibitem{lu2024mathvista}
Pan Lu, Hritik Bansal, Tony Xia, Jiacheng Liu, Chunyuan Li, Hannaneh Hajishirzi, Hao Cheng, Kai-Wei Chang, Michel Galley, and Jianfeng Gao.
\newblock Mathvista: Evaluating mathematical reasoning of foundation models in visual contexts.
\newblock In {\em International Conference on Learning Representations (ICLR)}, 2024.

\bibitem{Stanislawek2021KleisterKI}
Tomasz Stanislawek, Filip Grali'nski, Anna Wr'oblewska, Dawid Lipi'nski, Agnieszka Kaliska, Paulina Rosalska, Bartosz Topolski, and P.~Biecek.
\newblock Kleister: Key information extraction datasets involving long documents with complex layouts.
\newblock In {\em IEEE International Conference on Document Analysis and Recognition}, 2021.

\bibitem{deepform}
S.~Svetlichnaya.
\newblock {DeepForm}: Understand structured documents at scale., 2020.

\bibitem{Jaume2019FUNSDAD}
Guillaume Jaume, Hazim~Kemal Ekenel, and Jean-Philippe Thiran.
\newblock Funsd: A dataset for form understanding in noisy scanned documents.
\newblock {\em 2019 International Conference on Document Analysis and Recognition Workshops (ICDARW)}, 2:1--6, 2019.

\bibitem{Huang2019ICDAR2019CO}
Zheng Huang, Kai Chen, Jianhua He, Xiang Bai, Dimosthenis Karatzas, Shijian Lu, and C.~V. Jawahar.
\newblock Icdar2019 competition on scanned receipt ocr and information extraction.
\newblock {\em 2019 International Conference on Document Analysis and Recognition (ICDAR)}, pages 1516--1520, 2019.

\bibitem{textbookqa}
Aniruddha Kembhavi, Minjoon Seo, Dustin Schwenk, Jonghyun Choi, Ali Farhadi, and Hannaneh Hajishirzi.
\newblock Are you smarter than a sixth grader? textbook question answering for multimodal machine comprehension.
\newblock In {\em 2017 IEEE Conference on Computer Vision and Pattern Recognition (CVPR)}, pages 5376--5384, 2017.

\bibitem{Methani_2020_WACV}
Nitesh Methani, Pritha Ganguly, Mitesh~M. Khapra, and Pratyush Kumar.
\newblock Plotqa: Reasoning over scientific plots.
\newblock In {\em The IEEE Winter Conference on Applications of Computer Vision (WACV)}, March 2020.

\bibitem{Tanaka2021VisualMRCMR}
Ryota Tanaka, Kyosuke Nishida, and Sen Yoshida.
\newblock Visualmrc: Machine reading comprehension on document images.
\newblock {\em ArXiv}, abs/2101.11272, 2021.

\bibitem{chen-etal-2021-websrc}
Xingyu Chen, Zihan Zhao, Lu~Chen, JiaBao Ji, Danyang Zhang, Ao~Luo, Yuxuan Xiong, and Kai Yu.
\newblock {W}eb{SRC}: A dataset for web-based structural reading comprehension.
\newblock In Marie-Francine Moens, Xuanjing Huang, Lucia Specia, and Scott Wen-tau Yih, editors, {\em Proceedings of the 2021 Conference on Empirical Methods in Natural Language Processing}, pages 4173--4185, Online and Punta Cana, Dominican Republic, November 2021. Association for Computational Linguistics.

\end{thebibliography}
}

\newpage
\appendix
\section{Benchmark Construction Details}

\subsection{Existing Document Collection}
\label{appendix: Existing Document Collection}
Although previous datasets contain a relatively small proportion of lengthy documents, their absolute quantity should not be disregarded. Therefore, we compile lengthy documents from various datasets to include them as part of the documents in this benchmark. Specifically, we review and consider 21 previous document understanding (DU) datasets, and ultimately select 4 of them for further document selection. The selection reasons are shown in Table~\ref{tab:selection_reason}. All of these four datasets are licensed under the Creative Commons license (CC-BY) or other open-source licenses. Regarding the 4 selected datasets: DUDE~\cite{VanLandeghem2023DocumentUD}, SlideVQA~\cite{SlideVQA2023}, ChartQA~\cite{masry-etal-2022-chartqa} and FinanceBench~\cite{islam2023financebench}, we collect a total of 76 documents and detail our collection procedures as below.

\begin{table*}[!htbp]
 \centering
 \footnotesize
 \caption{Comparison of selected and considered datasets for our benchmark.}
 \label{tab:selection_reason}
 \renewcommand\tabcolsep{2.5pt} 
 \renewcommand\arraystretch{1.00} 
  \resizebox{1.\linewidth}{!}{
    \begin{tabular}{l|c|c}
    \textbf{Dataset} & \textbf{Selected} & \textbf{Comment} \\
    \toprule
    DUDE~\cite{VanLandeghem2023DocumentUD} & \cmark & - \\
    SlideVQA~\cite{SlideVQA2023} & \cmark & - \\
    ChartQA~\cite{masry-etal-2022-chartqa} & \cmark & - \\
    FinanceBench~\cite{islam2023financebench} & \cmark & - \\
    \midrule
    DocVQA~\cite{Mathew2020DocVQAAD} & \xmark & Repetitive with some documents/questions in DUDE; Single-page documents only \\
    MP-DocVQA~\cite{tito2023hierarchical} & \xmark & Repetitive with some documents/questions in DUDE;  Single-page questions only \\    
    Kleister Charity~\cite{Stanislawek2021KleisterKI} & \xmark & Repetitive with some documents/questions in DUDE; Over-simple \\ 
    Kleister NDA~\cite{Stanislawek2021KleisterKI} & \xmark & Repetitive with some documents/questions in DUDE; Over-simple \\ 
    DeepForm~\cite{deepform} & \xmark & Repetitive with some documents/questions in DUDE; Over-simple \\ 
    FUNSD~\cite{Jaume2019FUNSDAD} & \xmark & Repetitive with some documents/questions in DUDE; Over-simple \\
    SROIE~\cite{Huang2019ICDAR2019CO} & \xmark & Repetitive with some documents/questions in DUDE; Over-simple \\
    Infograohics VQA~\cite{Mathew2021InfographicVQA} & \xmark & Infographs are not long-context documents \\ 
    TAT-QA~\cite{zhu2022towards} & \xmark & Repetitive with some documents/questions in FinanceBench \\
    PWC~\cite{kardas-etal-2020-axcell} & \xmark & Repetitive with our self-annotated questions from academic papers \\  
    PaperQA~\cite{lu2024mathvista} & \xmark & Repetitive with our self-annotated questions from academic papers \\ 
    TextbookQA~\cite{textbookqa} & \xmark & Low document-relevance; Over-simple \\
    PlotQA~\cite{Methani_2020_WACV} & \xmark & Repetitive with our self-annotated questions from academic papers and research reports \\
    VisualMRC~\cite{Tanaka2021VisualMRCMR} & \xmark & Human performance reached; Website screenshots are not long-context documents \\
    WebSRC~\cite{chen-etal-2021-websrc} & \xmark & Human performance reached; Website screenshots are not long-context documents \\
    VisualWebBench~\cite{liu2024visualwebbench} & \xmark & Human performance reached; Website screenshots are not long-context documents \\
    PDFTriage~\cite{saadfalcon2023pdftriage} & \xmark & Not publicly available \\
    \bottomrule
    \end{tabular}
    }
\end{table*}

\noindent{\textbf{DUDE:}} We first filter all documents over 15 pages in the validation set of the original dataset, resulting in 87 documents. From these, we randomly sample 23 to include as a component of our benchmark documents.

\noindent{\textbf{SlideVQA}:} We download slide decks in the test set by following the instructions in the original repository~\footnote{https://github.com/nttmdlab-nlp/SlideVQA}. Pursuing lengthy documents, we slightly modified the code to remove the 20-page truncation procedure. Then we randomly select 27 slide decks for our benchmark documents.

\noindent{\textbf{FinanceBench}: We randomly sample 5 financial reports from the test set.}

\noindent{\textbf{ChartQA}: Different from the above three datasets, ChartQA only contains chart screenshots cropped from documents. We take the following steps to recover these original documents: (1) We use the Tesseract OCR model~\cite{smith2007overview} to recognize the text within the charts. (2) We use these texts as keywords to search for related documents on Google Search. (3) We manually identify these documents and remove all those that are less than 15 pages. From the ChartQA test set, we finalize a collection of 53 research reports from the Pew Research Center. We randomly sample 18 of these documents to include as a component of our benchmark documents.}

\newpage
\subsection{Newly-annotated Document Collection}
\label{appendix: Newly-annotated Document Collection}
Most documents collected from previous datasets are \textit{Industrial Files}, \textit{Tutorial \& Workshop}, \textit{Finance Report} and \textit{Research Report}. To diversify our benchmark, we additionally collect 59 documents including \textit{Academic Paper}, \textit{Brochure}, and \textit{Guideline}. We detail the collection procedures as below.

\noindent\textbf{Academic Paper} We collect 24 academic papers from Arxiv. All selected papers are over 15 pages (including references and appendix). To ensure annotation quality, each paper is either written or thoroughly read by at least one of the annotators.

\noindent\textbf{Guideline and Brochure} We collect 21 guidelines and 14 brochures from either ManualsLib or Google Search, covering diverse topics such as school, company, institution, products, service \etc. Each document is manually reviewed by one corresponding annotator and other primary authors to ensure its availability for academic use~\footnote{Should any authors request the removal of their documents, we will promptly comply.}.

\subsection{Document Examples}
\label{appendix: Document Examples}

As stated in Section 2.1, the documents in \benchmarkname can be categorized into seven types. We show the examples of each type as below.

\begin{figure}[!htbp]
    \centering
    \includegraphics[width=0.95\linewidth]{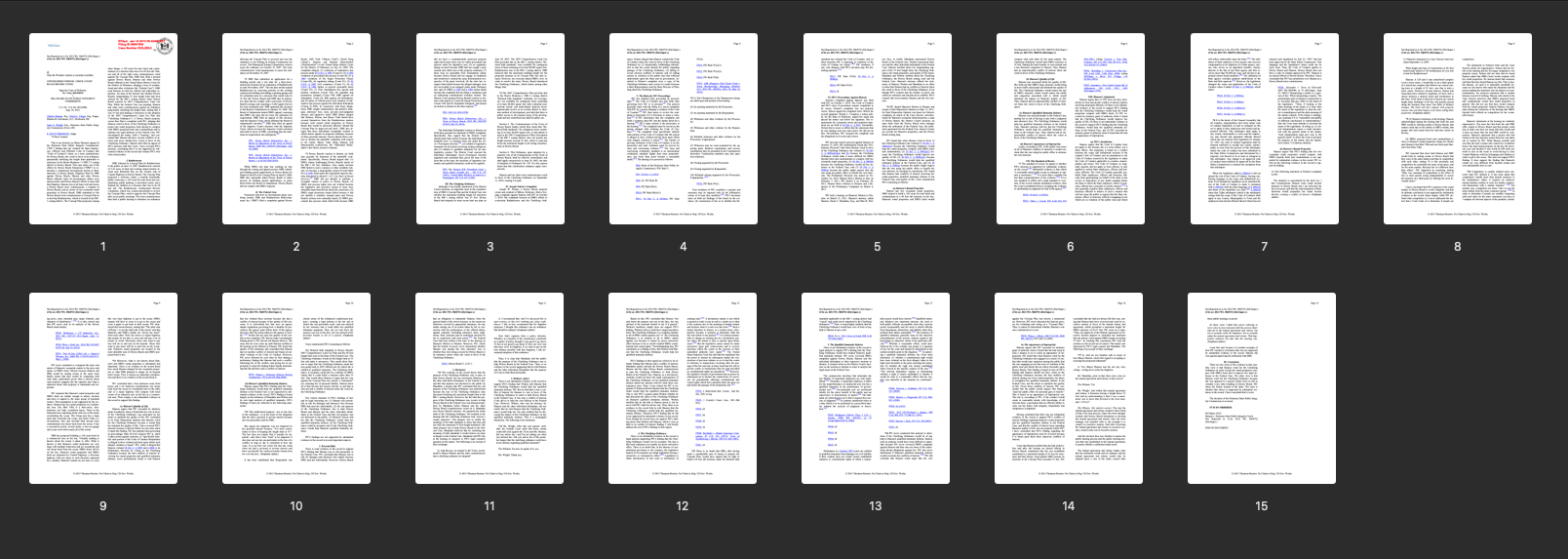}
    \caption{Document example about \textbf{Administration \& Industrial File}}
    \label{fig:enter-label}
\end{figure}

\begin{figure}[!htbp]
    \centering
    \includegraphics[width=0.95\linewidth]{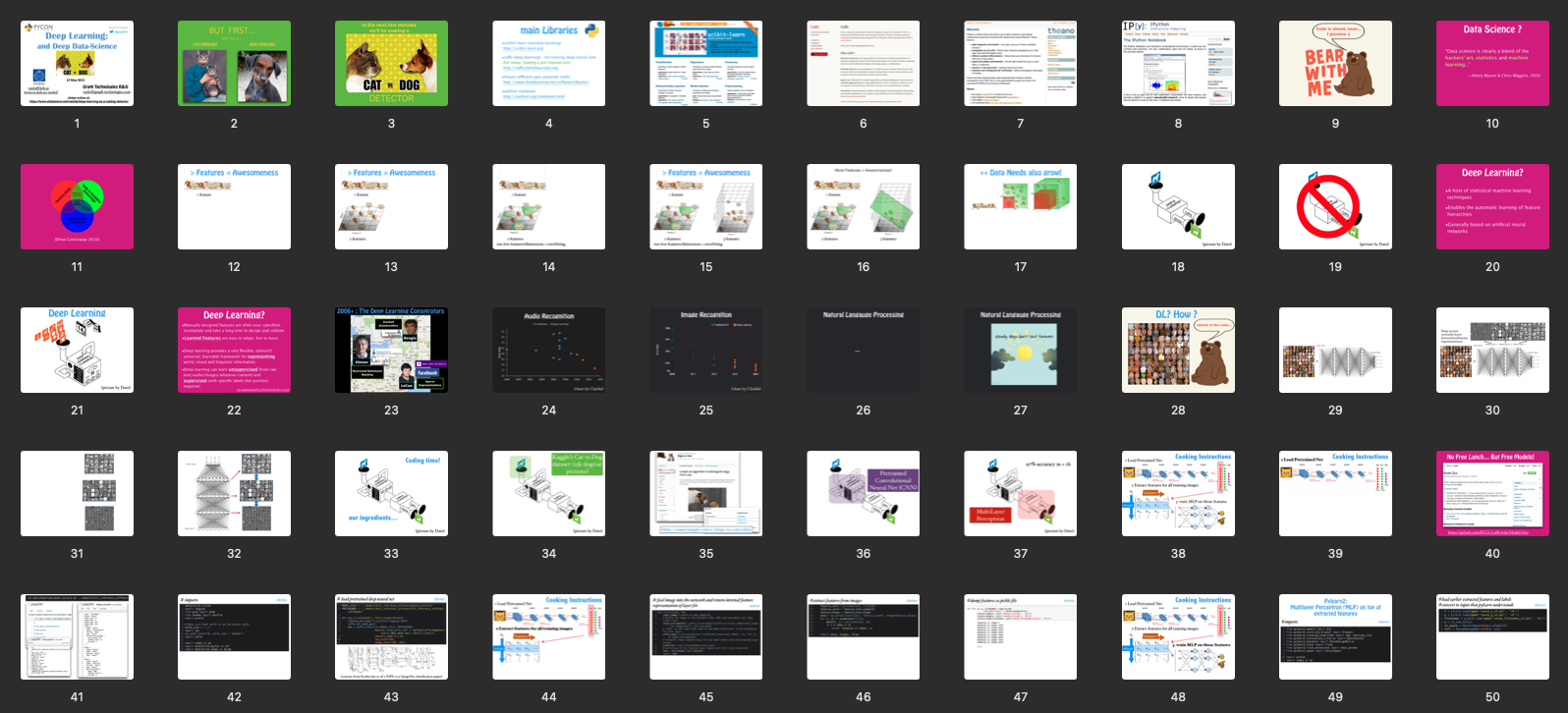}
    \caption{Document example about \textbf{Tutorial \& Workshop} (only show first 50 pages)}
    \label{fig:enter-label_t}
\end{figure}

\begin{figure}[!htbp]
    \centering
    \includegraphics[width=0.95\linewidth]{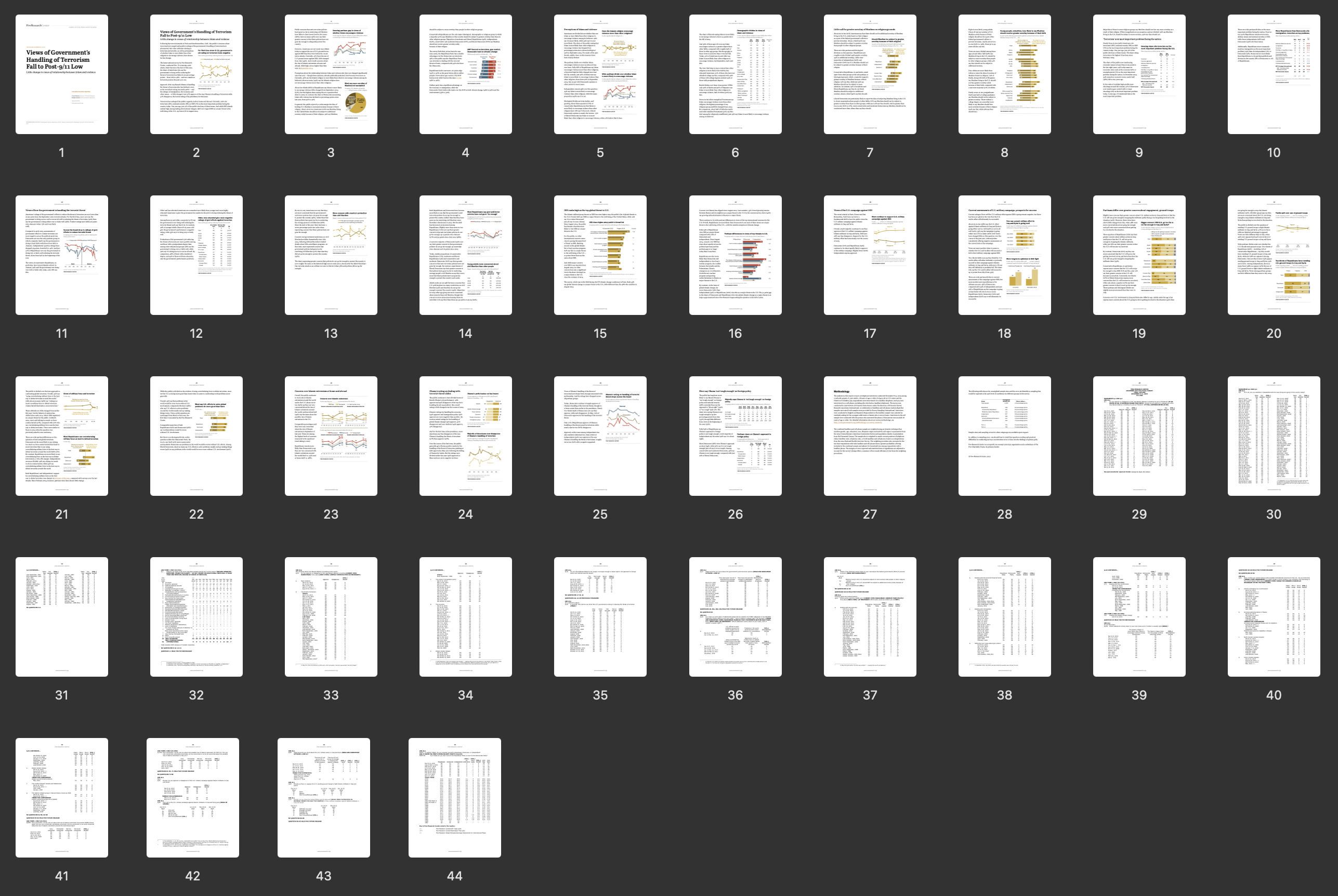}
    \caption{Document example about \textbf{Research Report}}
    \label{fig:enter-label_r}
\end{figure}

\begin{figure}[!htbp]
    \centering
    \includegraphics[width=0.95\linewidth]{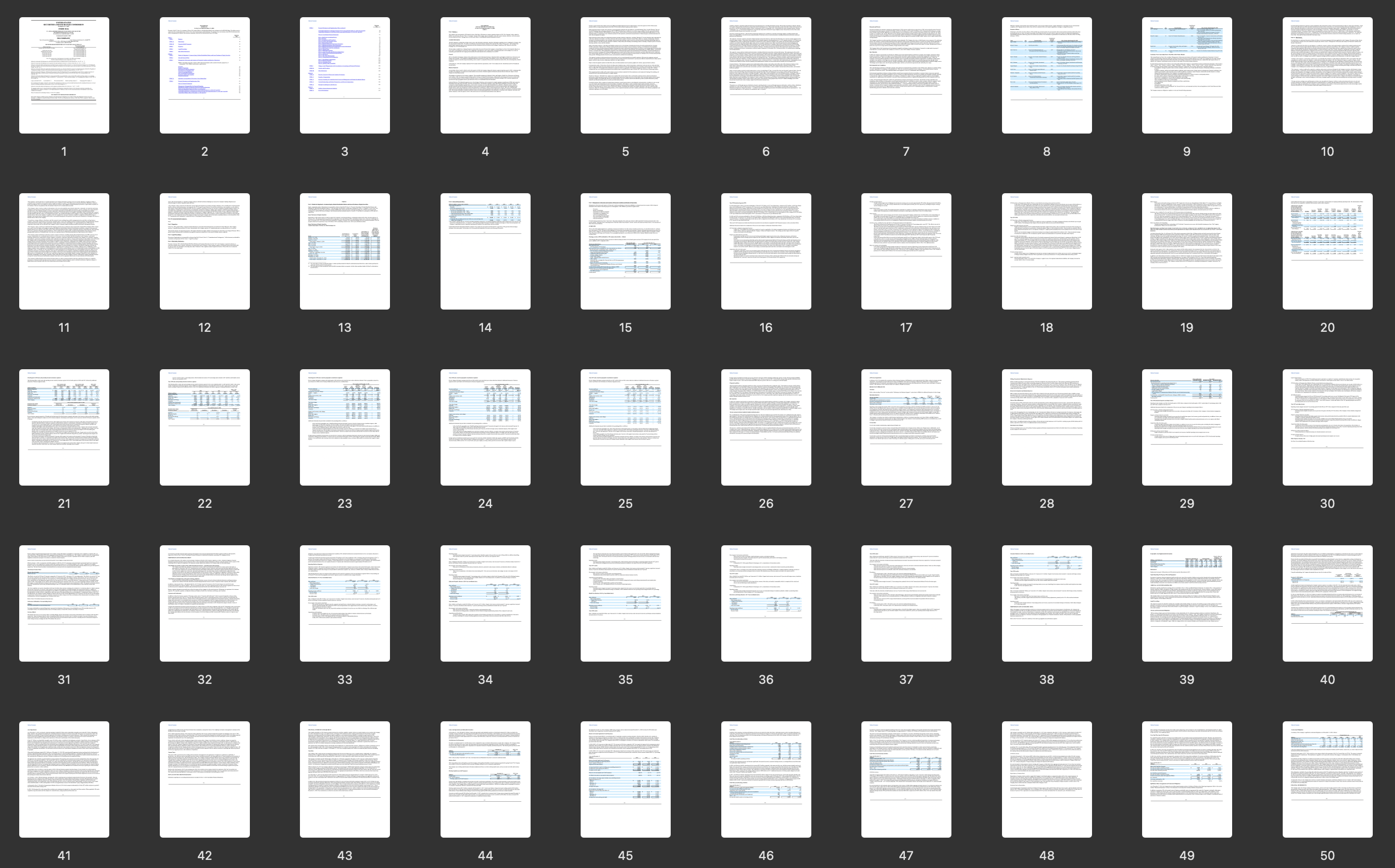}
    \caption{Document example about \textbf{Financial Report} (only show first 50 pages)}
    \label{fig:enter-label_f}
\end{figure}

\begin{figure}[!htbp]
    \centering
    \includegraphics[width=0.95\linewidth]{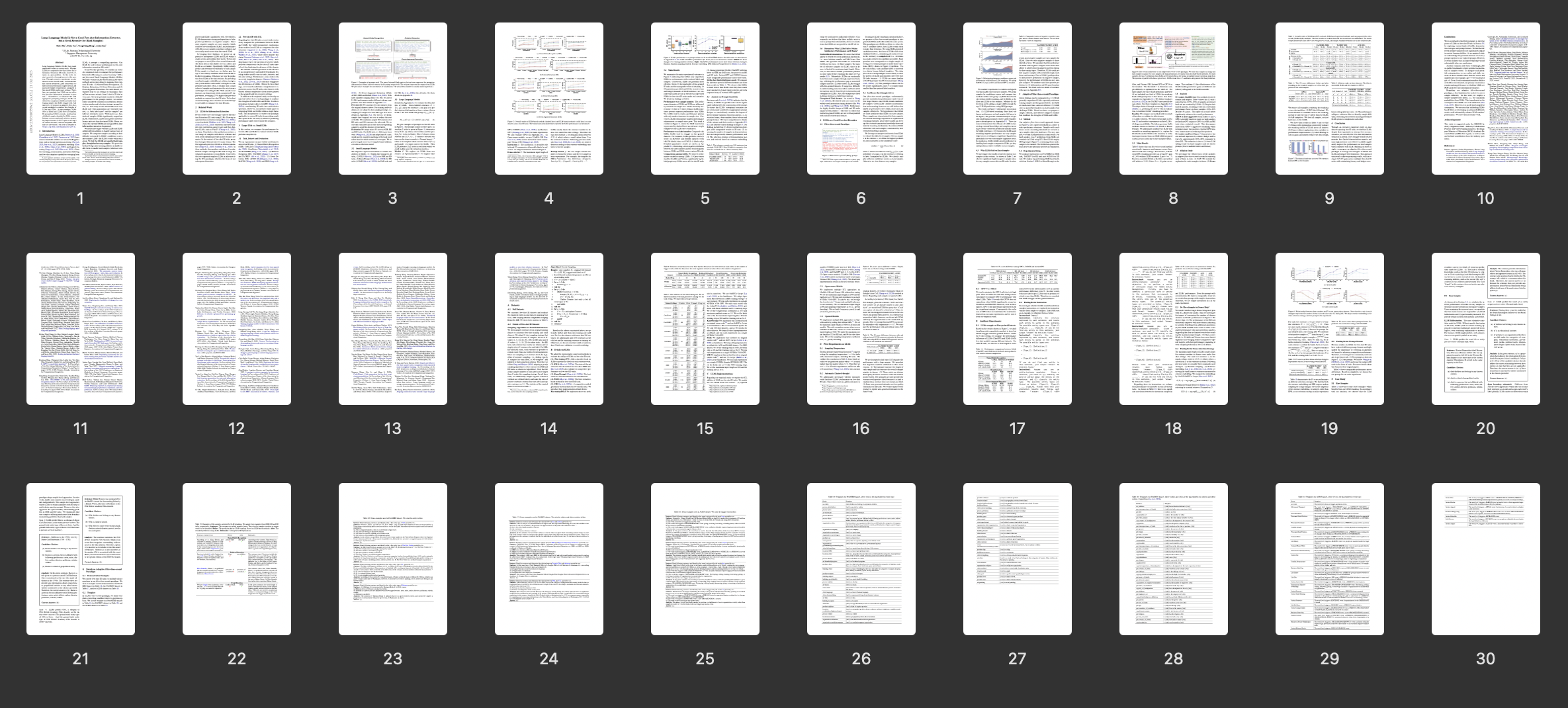}
    \caption{Document example about \textbf{Academic Paper}}
    \label{fig:enter-label_a}
\end{figure}

\begin{figure}[!htbp]
    \centering
    \includegraphics[width=0.95\linewidth]{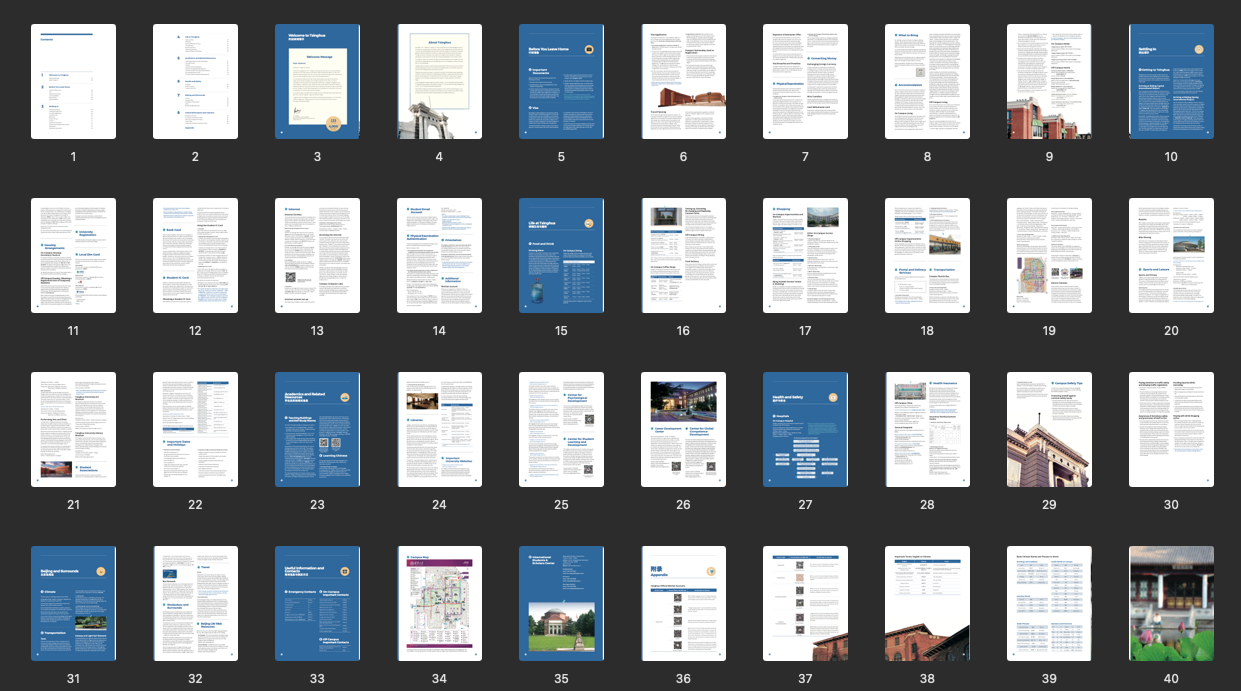}
    \caption{Document example about \textbf{Guidebook}}
    \label{fig:enter-label_g}
\end{figure}

\begin{figure}[!htbp]
    \centering
    \includegraphics[width=0.95\linewidth]{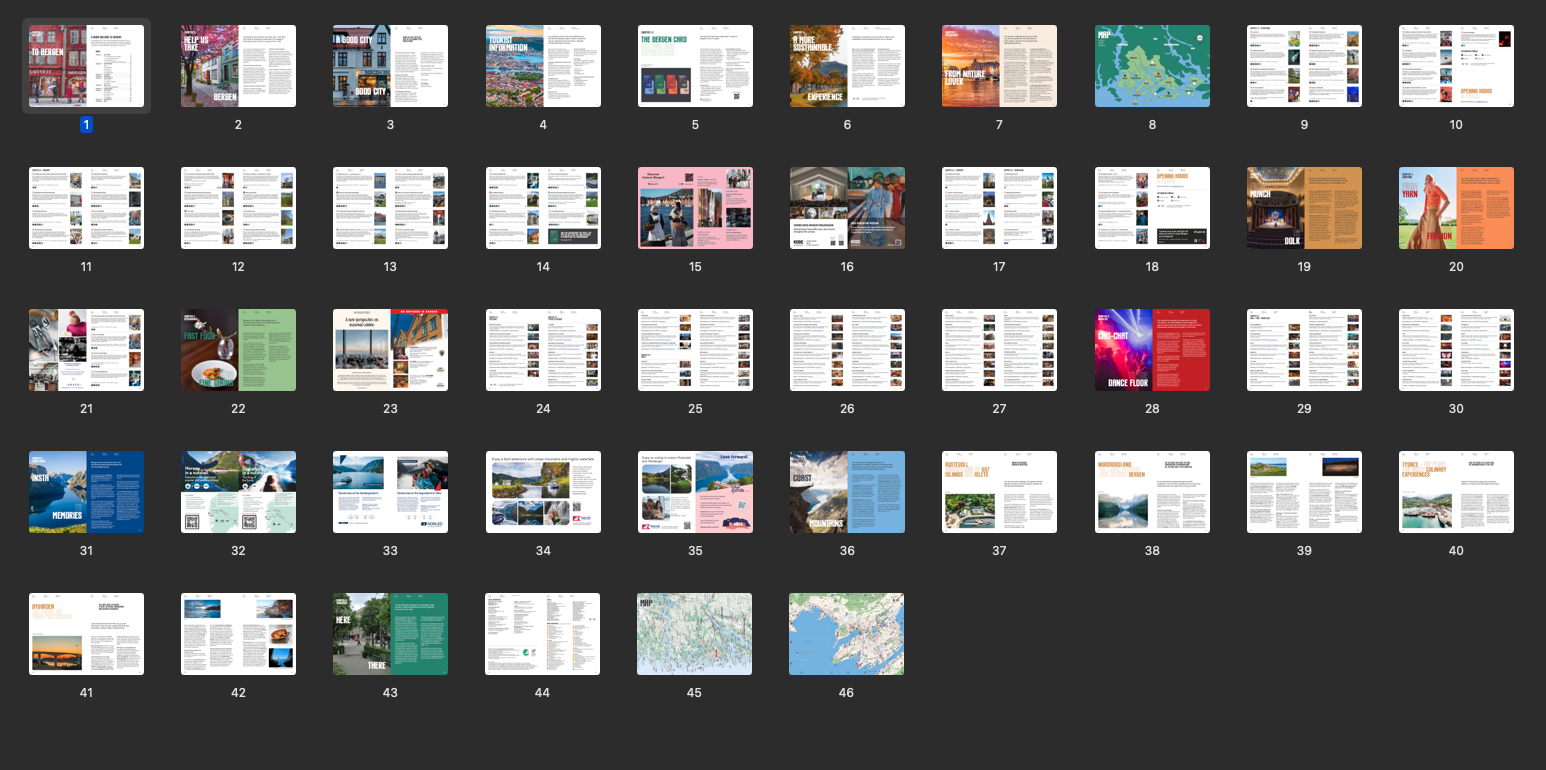}
    \caption{Document example about \textbf{Brochure}}
    \label{fig:enter-label_b}
\end{figure}

\newpage
\subsection{Existing Question Editing}
\label{subsec: Existing Question Editing}
Documents collected from existing datasets had been annotated with some questions and answers. However, their crowd-sourcing annotations inevitably make some questions, answers, and other meta information unqualified. So we conduct a systematic and manual pipeline to edit their annotations. Specifically, we classify six potential problems in original annotations. The definitions and examples of these problems are shown below.

\noindent\textbf{1. Wrong Answer or Evidence Pages:} The reference answers and/or evidence pages in original datasets are wrongly annotated.

\begin{figure}[!htbp]
    \centering
    \includegraphics[width=0.95\linewidth]{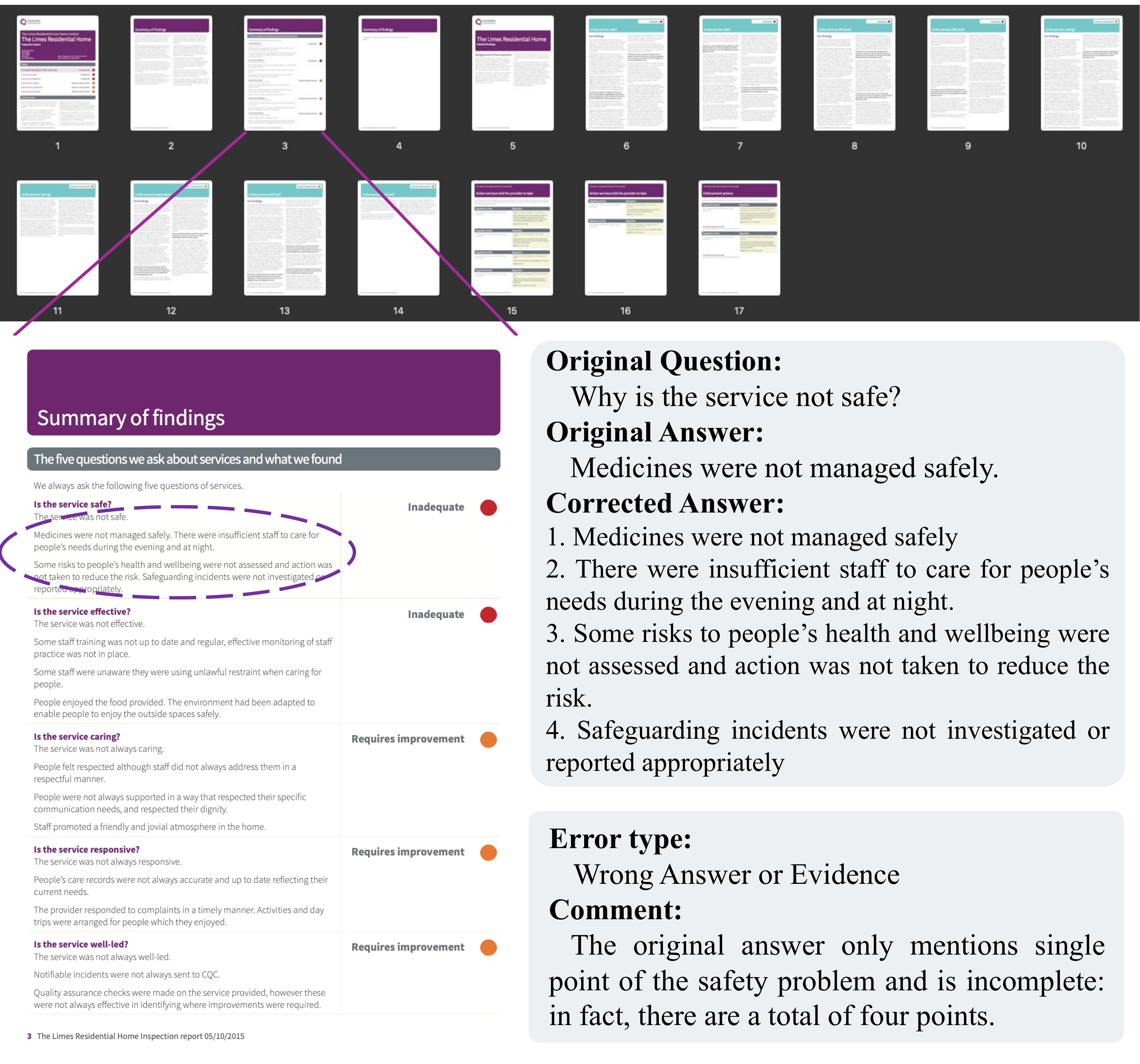}
    \caption{Example of the original annotation with \textit{Wrong Answer or Evidence Pages}.}
    \label{fig:enter-label_e}
\end{figure}

\clearpage
\noindent\textbf{2. Repetitive Question:} Too many questions with the same types (\eg key information extraction) occur in a single document (or even on the same page or point).
 
\begin{figure}[!htbp]
    \centering
    \includegraphics[width=0.95\linewidth]{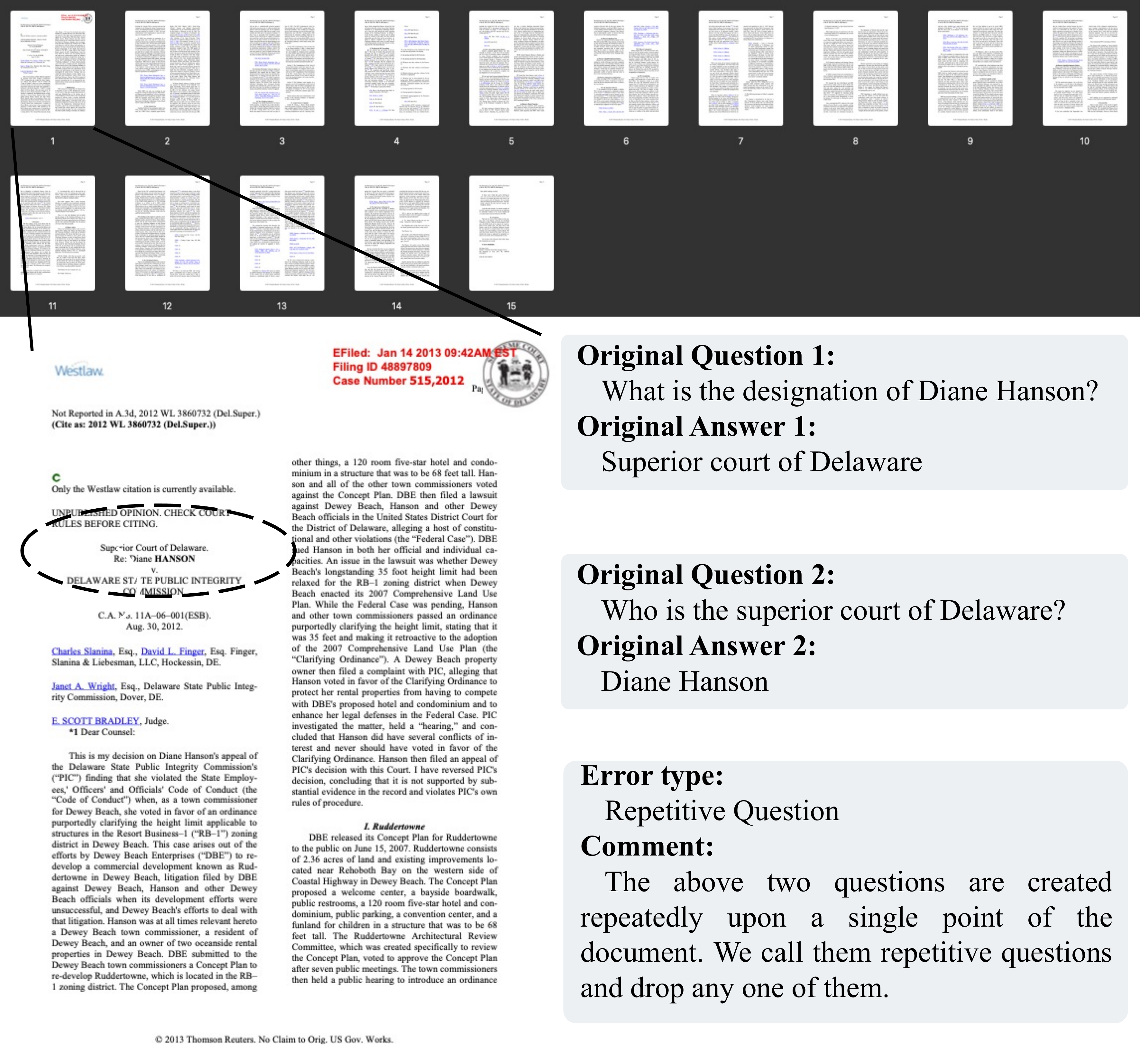}
    \caption{Example of the original annotation with \textit{Repetitive Question}.}
    \label{fig:enter-label_q}
\end{figure}

\clearpage
\noindent\textbf{3. Ambiguous Question:} The question is ambiguous at the document level (\eg the absence of entity, period, exact section or page, \etc), or too broad to exactly answer.

\begin{figure}[!htbp]
    \centering
    \includegraphics[width=0.95\linewidth]{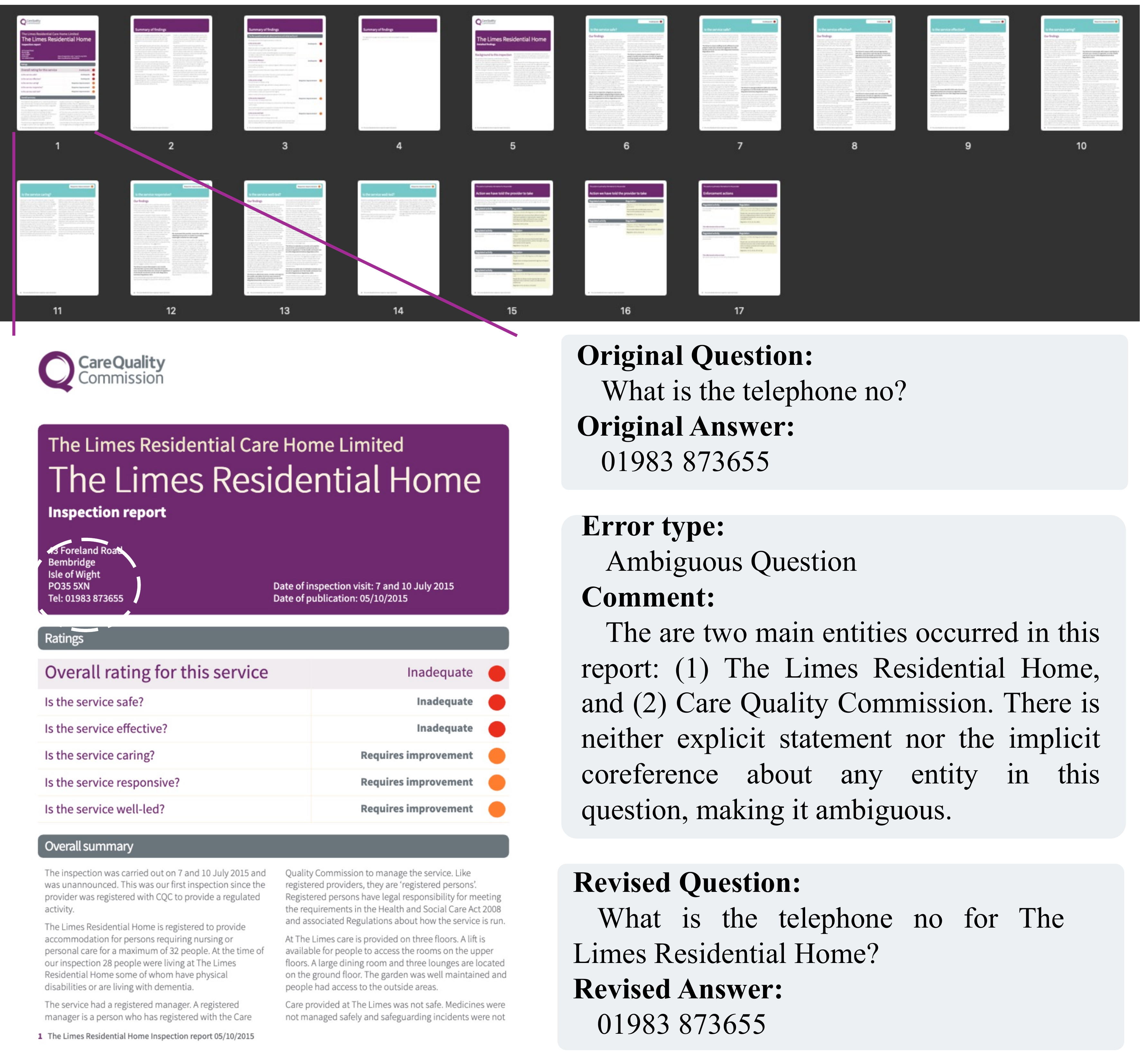}
    \caption{Example of the original annotation with \textit{Ambiguous Question}.}
    \label{fig:enter-label_o}
\end{figure}

\clearpage
\noindent\textbf{4. Potential Shortcut:} The resolution of the question does not rely on two entities (across different pages) but only one of them, \ie there exists a shortcut for this question.

\begin{figure}[!htbp]
    \centering
    \includegraphics[width=0.92\linewidth]{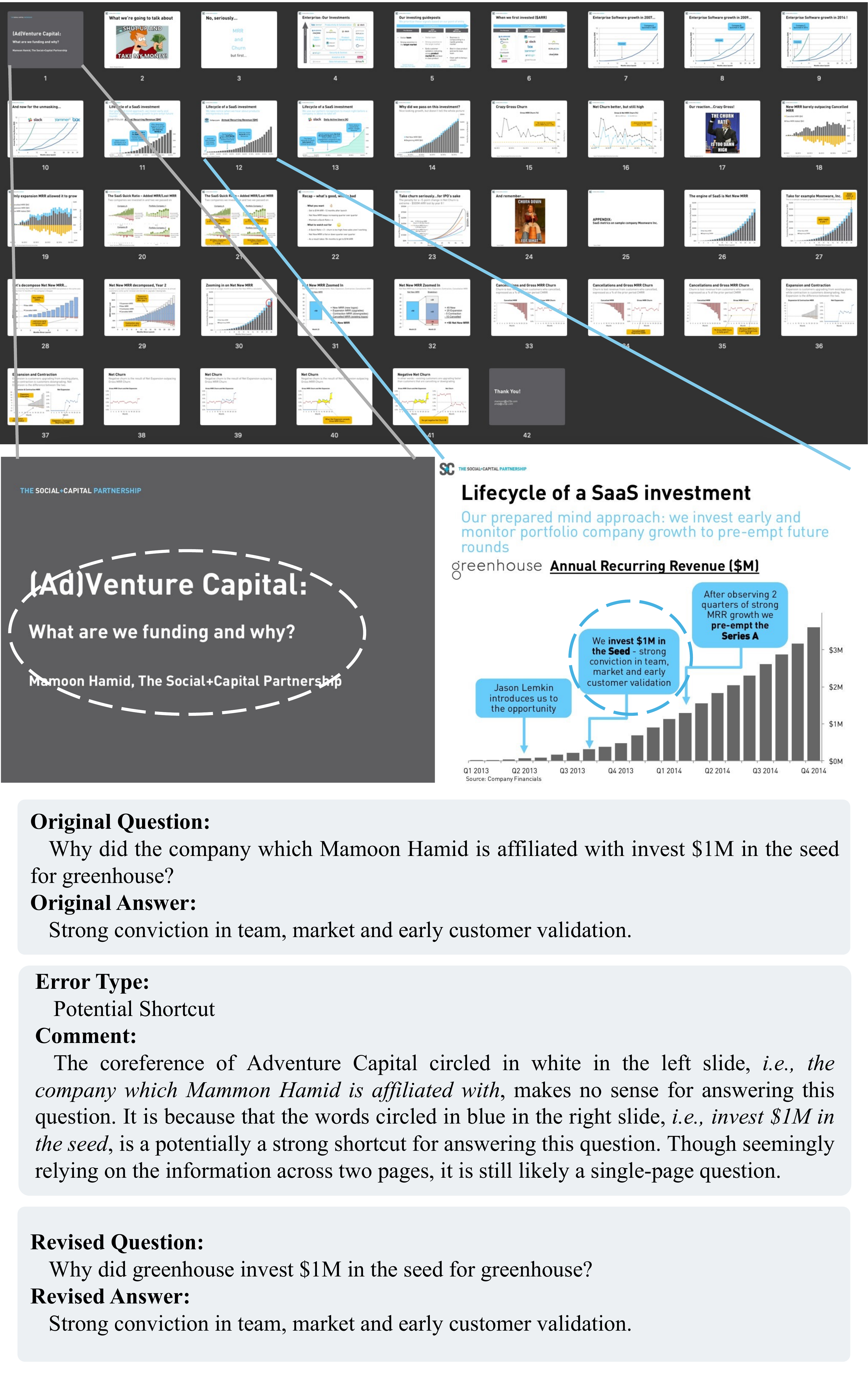}
    \caption{Example of the original annotation with \textit{Potential Shortcut}.}
    \label{fig:enter-label_oa}
\end{figure}

\clearpage
\noindent\textbf{5. Low Document-relevant Question:} The resolution of the question does not rely on the information from the document. It can be solved by the parametric knowledge in the LVLMs.

\begin{figure}[!htbp]
    \centering
    \includegraphics[width=0.95\linewidth]{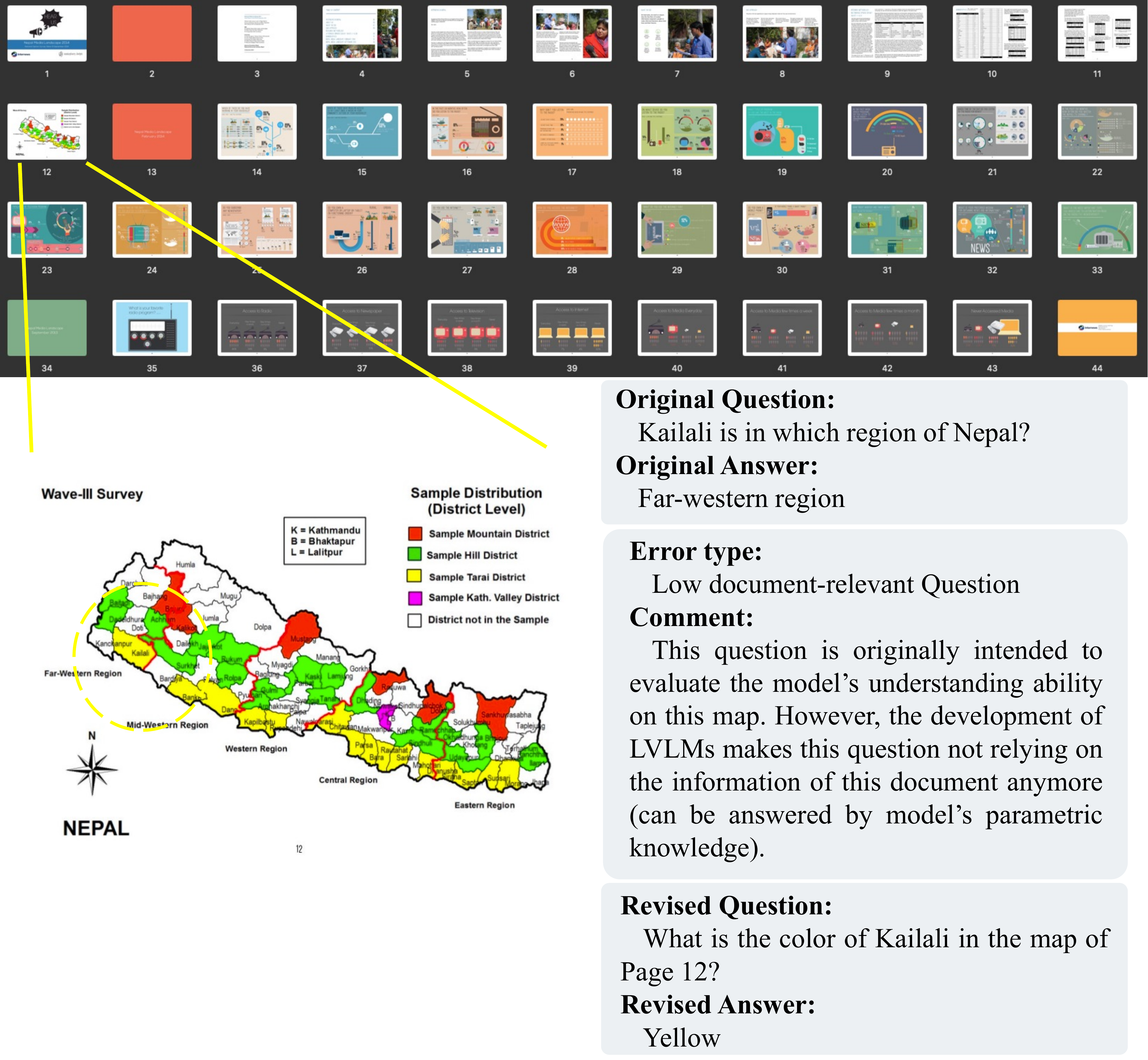}
    \caption{Example of the original annotation with \textit{Low Document-relevant Question}.}
    \label{fig:enter-label_l}
\end{figure}

\clearpage
\noindent\textbf{6. Decontextulization-required Question:} The understanding of the question is conditioned on a single page or even a single component of the document.

\begin{figure}[!htbp]
    \centering
    \includegraphics[width=0.95\linewidth]{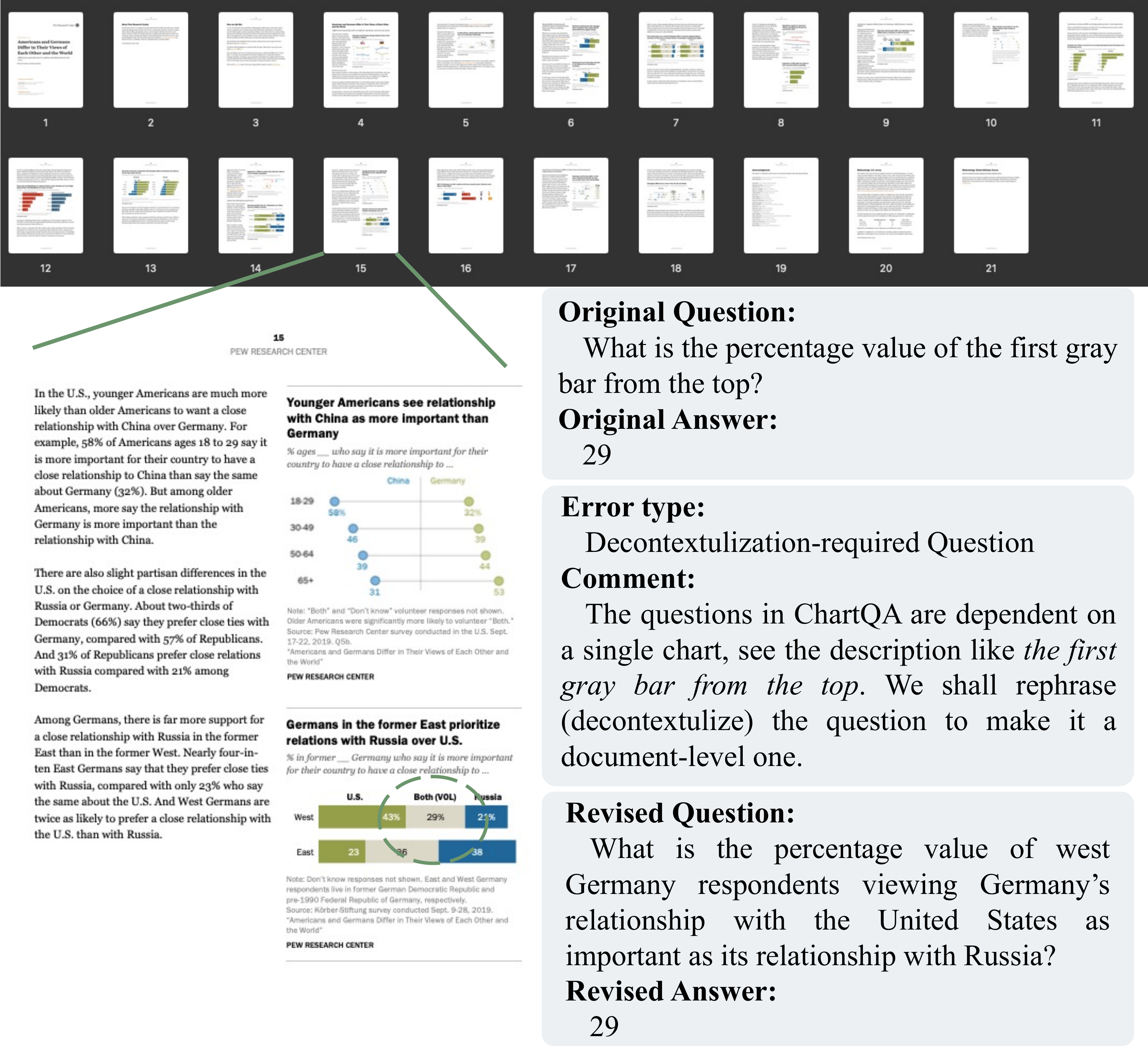}
    \caption{Example of the original annotation with \textit{Decontextulization-required Question}.}
    \label{fig:enter-label_dr}
\end{figure}

When dealing with questions categorized under any of these six problem types, annotators are instructed to either revise or remove them. Typically, repetitive questions and those with potential shortcuts are removed. In contrast, wrongly-annotated or decontextualization-required questions are generally revised. For ambiguous and low document-relevant questions, the course of action depends more on the annotators' discretion.

\clearpage
\subsection{New Question Annotation}
\label{appendix: New Question Annotation}
We annotate new questions on both existing and newly-collected documents. To ensure a diverse range of questions, we impose limitations on the question distributions categorized by their types (\ie single-page, cross-page or unanswerable) and evidence sources (\ie table, chart, image). To balance existing questions which are mostly single-page and text-based, we place greater emphasis on cross-page, unanswerable, table-related, chart-related, and image-related questions. The detailed standards are as follows:

\begin{table}[!htbp]
    \centering
        \begin{tabular}{l|cc|ccc|c}
        \toprule
        \multirow{2}{*}{\textbf{Document Type}}  &  \multicolumn{2}{c}{\textbf{Evidence Page}}  & \multicolumn{3}{c}{\textbf{Evidence Source}} & \multirow{2}{*}{\textbf{All}} \\
        & Cross-page & Unanswerable & Table & Chart & Image \\
        \midrule
        Industrial File & $\geq 2$ & - & \multicolumn{3}{c}{-} \vline & $\geq 3$ \\
        Workshop \& Tutorial & $\geq 2$  & $\geq 1$ & \multicolumn{3}{c}{------ $\geq 3$ ------} \vline & $\geq 6$  \\
        Research Report &  $\geq 3$ & $\geq 1$ & $\geq 2$ & $\geq 2$ & - & $\geq 5$ \\
        Financial Report &  $\geq 5$ & $\geq 2$ & $\geq 7$ & - & - & $\geq 10$ \\
        Academic Paper &  $\geq 3$ & $\geq 1$ & $\geq 2$ & \multicolumn{2}{c}{---- $\geq 3$ ----} \vline & $\geq 6$ \\
        Guidebook & $\geq 3$ & $\geq 1$ & -& - & $\geq 4$ & $\geq 7$ \\ 
        Brochure & $\geq 2$ & $\geq 1$ & -& - & $\geq 3$ & $\geq 7$ \\
        \bottomrule
        \end{tabular}
    \vspace{0.5em}
    \caption{The \textbf{minimum} requirements for the number and distribution of questions, categorized by the evidence page numbers and evidence sources. We have set varying requirements for different document types based on their specific characteristics.}
    \label{tab:my_label}
\end{table}

\subsection{Potential Bias for LVLM-based Quality Checking}
\label{appendix: potential_bias}
As described in Section~\ref{subsec: quality control}, we employ GPT-4o to remove document-agnostic (\ie can be correctly answer without documents) samples and review potential wrongly-labeled samples. A reasonable speculation raises that our final benchmark can be biased toward GPT-4o’s answers, especially when GPT-4o outperforms others by a large margin. We discuss this potential bias as follows. 

We check the effect of GPT-4o's involvement in the quality control step-by-step. Specifically, we compare the performance of samples remained after each step across GPT-4o and two other competitive models (GPT-4V and Gemini-1.5-Pro). We show their results in the table below.

\begin{table}[!htbp]
    \centering
    \resizebox{0.95\linewidth}{!}{
    \begin{tabular}{l|ccc}
    \toprule
        & \textbf{GPT-4o} & \textbf{GPT-4V} & \textbf{Gemini-1.5-Pro} \\
         \midrule
        No quality control & 43.1\% & 35.2\%  & 23.3\%  \\
        \quad + document-relevance detection & 41.2\%  & 31.0\%  & 20.5\%  \\
        \quad + document-relevance detection + self-reflection / cross-checking & 42.7\%  & 31.4\%  & 20.9\%  \\
        \bottomrule
    \end{tabular}
    }
    \vspace{1em}
    \caption{Step-wise performance comparison with and without LVLM-based quality checking}
    \label{tab:perform_compare}
\end{table}

The results illustrate that the potential bias in step 1 (document-relevance detection) actually reduce, rather than increase, the performance gap between GPT4o and other models. It is because that we filter out all samples correctly answered by GPT4o without the access to documents. Under this case, the more significant performance drop of GPT-4V and Gemini-1.5-Pro can only be attributed to their limited document understanding and over-reliance on their internal knowledge. Regarding the step 2 and 3 (self-reflection and cross-checking), we provide inconsistent answers between human annotations and GPT4o's predictions to annotators and ask them to check and revise accordingly. The potential bias of this step does lead to a slight performance bias (1.1\% absolute difference at maximum). We believe that such bias is NOT the main cause of GPT4o's significantly best performance. Without the involvement of GPT-4o in the quality control process, GPT-4o still significantly outperforms GPT-4V by 7.9\% (43.1\% - 35.2\%) and Gemini-1.5-Pro by 19.8\% (43.1\% - 23.3\%). Accordingly, all primary conclusions in our paper still hold.

\clearpage
\subsection{GUI Screenshots}
\label{appendix: GUI Screenshots}
\begin{figure}[!htbp]
    \centering
    \includegraphics[width=0.95\linewidth]{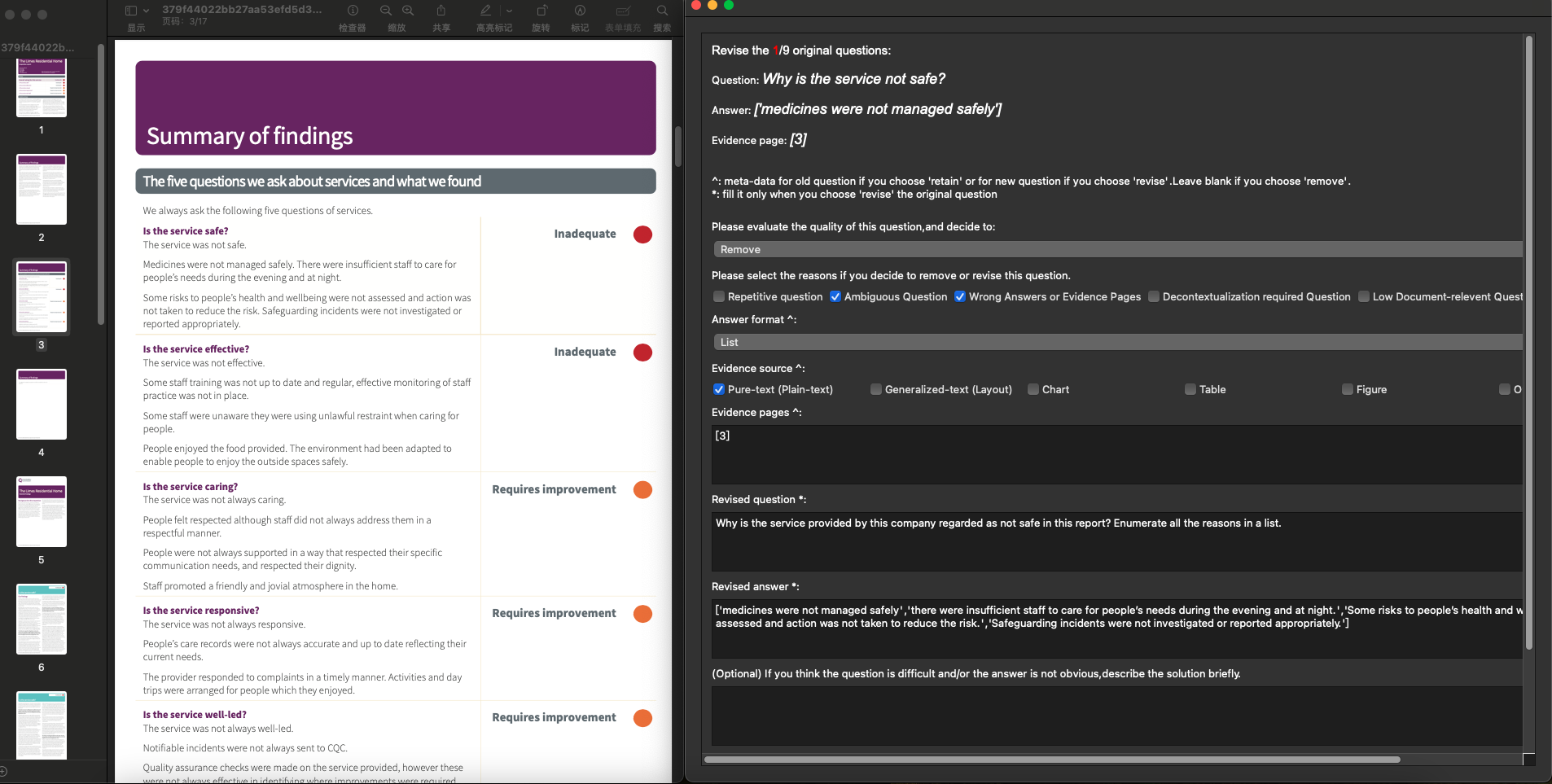}
    \caption{GUI screenshot for editing existing questions (along with reference answers and meta-data)}
    \label{fig: GUI_existing}
\end{figure}

\begin{figure}[!htbp]
    \centering
    \includegraphics[width=0.95\linewidth]{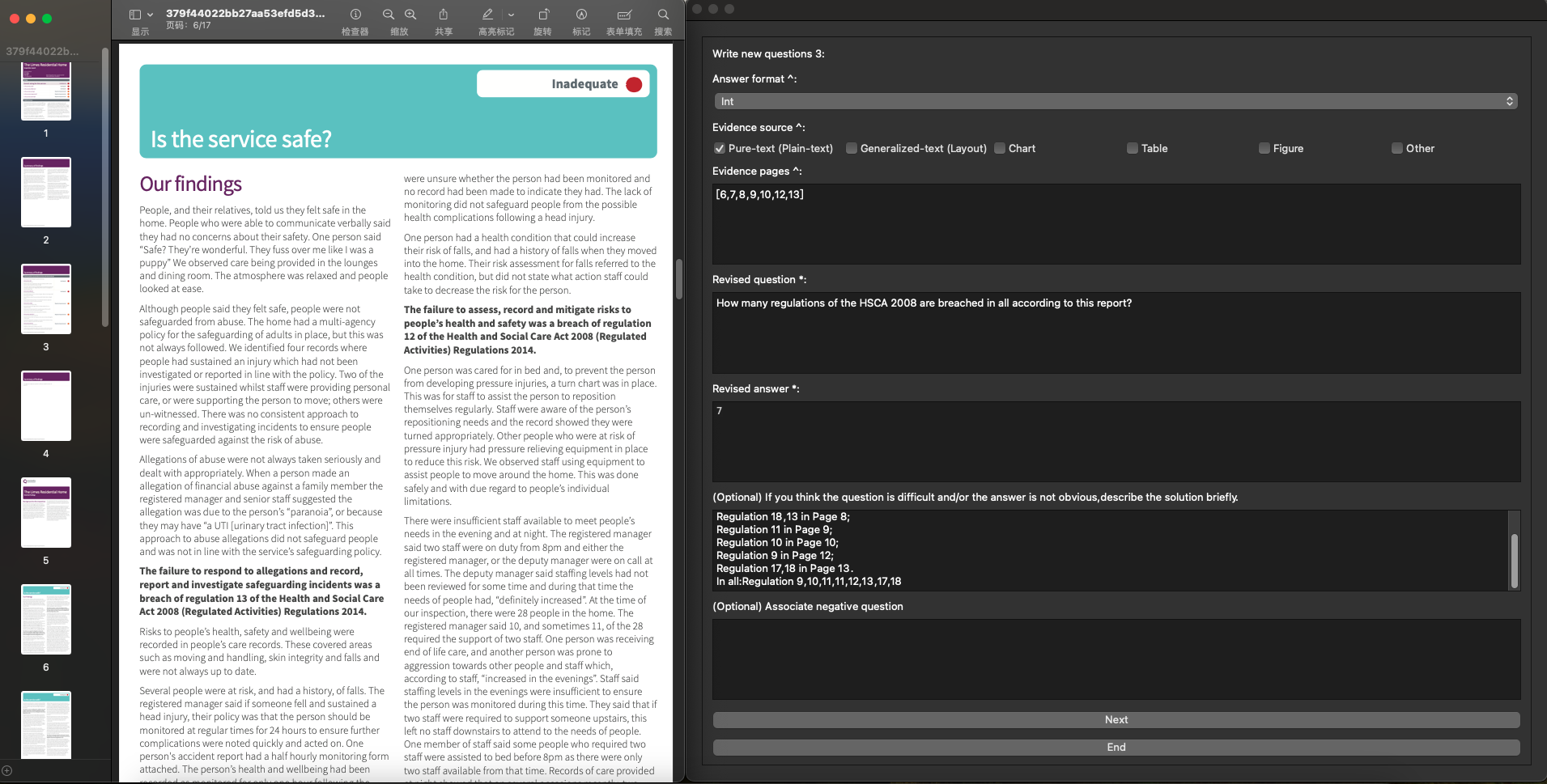}
    \caption{GUI screenshot for annotating new questions (along with reference answers and meta-data)}
    \label{fig: GUI_new}
\end{figure}

We present the screenshots for editing existing questions and annotating new questions (along with their reference answers and meta-data) in Figure~\ref{fig: GUI_existing} and Figure~\ref{fig: GUI_new} respectively.

\subsection{Annotation Cost}
This benchmark is annotated by the authors of this paper. Therefore, the data collection does not need compensation. And we count the time cost of our benchmark as below.

\noindent \textbf{Pre-annotation} (about 45h): the development of annotation interface (10h), the writing of annotation guideline (5h), training session (10h), preliminary annotation and personalized feedback (20h).

\noindent \textbf{Annotation} (about 150h): It takes about 60-90 minutes for the annotation of each document. And all of the 130 documents take about 150 hours.

\noindent \textbf{Post-annotation} (about 45h): quality checking (30h), data processing and release preparation (15h).

In summary, our benchmark annotation approximately takes a total of 45+150+45=240 hours (1.36 man months).
\clearpage
\section{Experimental Details}
\label{appendix: experimental_details}

\subsection{Prompt for Response Generation}
\begin{lstlisting}[caption={Prompt used for response generation. The \texttt{[Document]} is in PNG format (page screenshots) for LVLMs, and TXT format for LLMs.}, style=prompt]
[Document]

Read the above documents and answer this question: 
[question]

Please make your answer as concise as possible.
\end{lstlisting}

\subsection{Prompt for Answer Extraction}
\begin{lstlisting}[caption={Prompt used for answer extraction.}, style=prompt]
Given the question and analysis, you are tasked to extract answers with required formats from the free-form analysis. 
- Your extracted answers should be one of the following formats: (1) Integer, (2) Float, (3) String and (4) List. If you find the analysis the question can not be answered from the given documents, type "Not answerable". Exception: If the analysis only tells you that it can not read/understand the images or documents, type "Fail to answer".
- Please make your response as concise as possible. Also note that your response should be formatted as below:
```
Extracted answer: [answer]
Answer format: [answer format]
```

Please read the following example, then extract the answer from the model response and type it at the end of the prompt. 

---
Question: List the primary questions asked about the services in this report.
Analysis:  The primary questions asked about the services in the report for The Limes Residential Home are:
1. Is the service safe?
2. Is the service effective?
3. Is the service caring?
4. Is the service responsive?
5. Is the service well-led?
Extracted answer: ['Is the servife safe?', 'Is the service effective', 'Is the serve caring?', 'Is the service responsive?', 'Is the service well-led?']
Answer format: List

---
Question: How many regulations of the HSCA 2008 are breached in all according to this report?
Analysis: According to the report, the provider breached 10 Health and Social Care Act 2008 (Regulated Activities) Regulations in total. Here are the specifics:

1. Regulation 13: Safeguarding service users from abuse and improper treatment
2. Regulation 12: Safe care and treatment
3. Regulation 18: Staffing
4. Regulation 11: Need for consent
5. Regulation 10: Dignity and respect
6. Regulation 9: Person-centred care
7. Regulation 17: Good governance
8. Regulation 18 (CQC Registration Regulations 2009): Notification of other incidents
9. Regulation 18: Failure to maintain an accurate and up-to-date care plan
10. Regulation 11: Failure to implement the Mental Capacity Act 2005 code of practice effectively

These breaches involve issues concerning staffing, safeguarding, medicines management, dignity and respect, consent, care planning, governance, and failure to notify the CQC of incidents.
Extracted answer: 10
Answer format: Integer

---
Question: According to the survey that is the percentage of Chinese who are paying more or about the same attention to politics after Trump's election?
Analysis: The survey provided does not specify the percentage of Chinese individuals specifically who are paying more or about the same attention to politics after Trump's election. The report focuses primarily on American demographics and does not include specific details about the Chinese population in relation to this question. If you need information about a different demographic or a summary of the findings from the American demographic, I can certainly help with that!
Extracted answer: Not answerable
Answer format: None

---
Question: How many quotations from male respondent over 50 years old are included in this report?
Analysis: The image you've provided appears to be a screenshot of a document with multiple charts. However, the text is too small and blurry to read accurately. If you can provide a clearer image or more context, I might be able to help you with your question.
Extracted answer: Fail to answer
Answer format: None

---
\end{lstlisting}

\subsection{Rules for Score Calculation}
We evaluate the model's responses by scoring the extracted answers against the reference answers. The scorer is rule-based and employs different strategies according to the format of the reference answer. We detail its rules as below:

\noindent\textbf{String:} We firstly use a series of regular expressions to determine whether the answers require exact matching (\eg telephone numbers, email addresses, website addresses, file names, times, dates, \etc). If an exact match is needed, we perform a straightforward string comparison and score the answer either 0 or 1. Otherwise, we follow previous work~\cite{VanLandeghem2023DocumentUD} and calculate the ANLS (Average Normalized Levenshtein Similarity) with a pre-defined threshold ($\tau = 0.5$).

\noindent\textbf{Integer:} We perform an exact match comparison and score the answer either 0 or 1.

\noindent\textbf{Float:} We view the prediction and reference answers as equal if they fall within a 1\% relative tolerance. 

\noindent\textbf{List:} We adopt a relatively strict rule for scoring answers in list format: predictions that do not have the same number of elements as the reference receive a score of 0. For the remaining predictions, as Eq.~\ref{eq:list_score} indicates, we score each element in order and use the minimum element-wise score as the score for the entire list. The element-wise scoring strategies is determined by the formats of elements (\ie string, integer or float).

\begin{equation}
    \label{eq:list_score}
    \begin{aligned}
    & \texttt{pred\_list}, \texttt{ref\_list}  = \texttt{sorted(pred\_list)}, \texttt{sorted(ref\_list)}   \\
    & \texttt{Score(pred\_list, ref\_list)} = \texttt{min}( \\
    & \quad \quad [\texttt{Score(pred, ref) for pred, ref in zip(pred\_list, ref\_list)}] \\
    & )
    \end{aligned}
\end{equation}

Evaluation detailed in the Appendix~\ref{subsec: Human Evaluation on the Automatic Evaluation Pipeline} shows that while this scorer is not perfect, it aligns well with human judgment. We will continue refining these rules to cover more corner cases and enhance their accuracy.

\subsection{Human Evaluation on the Automatic Evaluation Pipeline}
\label{subsec: Human Evaluation on the Automatic Evaluation Pipeline}
We conduct human evaluations to assess the performance of our automatic evaluation pipeline, which includes the answer extractor and the score calculator. Specifically, we randomly select 100 questions and review their responses from two representative LVLMs: GPT-4o and Gemini-1.5-Pro. We manually evaluate the correctness of each response and compare the results between human evaluation and automatic evaluation. The performance, as shown in Table~\ref{tab:eval_compare}, indicates a high correlation between human judgment and our automatic pipeline.

\begin{table}[!htbp]
    \centering
    \begin{tabular}{l|ccc}
    \toprule
        \multirow{2}{*}{\textbf{Model}} & \multicolumn{3}{c}{\textbf{Inconsistent Evaluation}} \\
         & Ans. Extractor & Scorer & Overall \\
         \midrule
        GPT-4o & 4 & 2 & 6 \\
        Gemini-1.5-Pro & 2 & 2 & 4 \\
        \bottomrule
    \end{tabular}
    \vspace{1em}
    \caption{We manually check 100 responses from GPT-4o and Gemni-1.5-Pro, and compare the evaluation results between humans and our automatic pipeline.}
    \label{tab:eval_compare}
\end{table}

\subsection{Model Hyperparameters}
\label{appendix: Hyperparameters}
The hyperparameters of used LVLMs and LLMs in Section 3.3 are detailed in Table~\ref{tab:hyperparameters}. The temperature is set as $0.0$, and the max\_new\_tokens is set as $1024$ for all the models. The `concatenated\_images' parameter determines the maximum number of images that can be combined into a single input for LVLMs. By concatenating multiple images, we can meet the minimum context window requirements. The `max\_pages' parameter specifies the maximum number of images that can be directly input into the LVLMS without concatenation.

\begin{table}[!htbp]
    \centering
    \resizebox{0.8\linewidth}{!}{
    \begin{tabular}{l|c}
        \toprule
        \textbf{Model} & \textbf{Hyperparameters}  \\
        \midrule
        \multicolumn{2}{l}{\textit{LLM}} \\
        \midrule
        ChatGLM-128k & max\_input\_words=60000 \\
        Mistral-Instruct-v0.2-7B & max\_input\_words=20000  \\ 
        Mixtral-Instruct-v0.1-8x7B & max\_input\_words=20000 \\
        Mixtral-Instruct-v0.1-8x22B & max\_input\_words=40000  \\
        QWen-Plus & max\_input\_words=16000 \\
        DeepSeek-V2 & max\_input\_words=20000  \\
        \midrule
        \multicolumn{2}{l}{\textit{LVLM}} \\
        \midrule
        DeepSeek-VL-Chat & concatenated\_images=5 \\
        Qwen-VL-Chat & concatenated\_images=5 \\
        Idefics2 &concatenated\_images=5 \\
        MiniCPM-Llama3-V2.5 & concatenated\_images=2 \\
        InternLM-XC2-4KHD & concatenated\_images=2 \\
        Monkey-Chat & concatenated\_images=1 \\
        CogVLM2-Llama3-Chat & concatenated\_images=1 \\
        InternVL-Chat-v1.5 & concatenated\_images=5 \\
        EMU2-Chat & concatenated\_images=5 \\
        \midrule
        \multicolumn{2}{l}{\textit{LLM \& LVLM}} \\
        \midrule
        Claude-3 Opus & version=\texttt{claude-3-opus-20240229}, concatenated\_images=20 \\
        Gemini-1.5-Pro & max\_pages=120, version=\texttt{gemini-1.5-pro-latest} \\
        GPT-4-turbo & max\_pages=120, version=\texttt{gpt-4-turbo-2024-04-09} \\
        GPT-4o & max\_pages=120, version=\texttt{gpt-4o-2024-05-13} \\
        \bottomrule
    \end{tabular}
    }
    \vspace{1em}
    \caption{Model Hyperparameters}
    \label{tab:hyperparameters}
\end{table}

\section{Qualitative Study}

\subsection{Error Analysis}
\label{appendix: error_analysis}
\begin{figure}[!htbp]
    \centering
    \includegraphics[width=\linewidth]{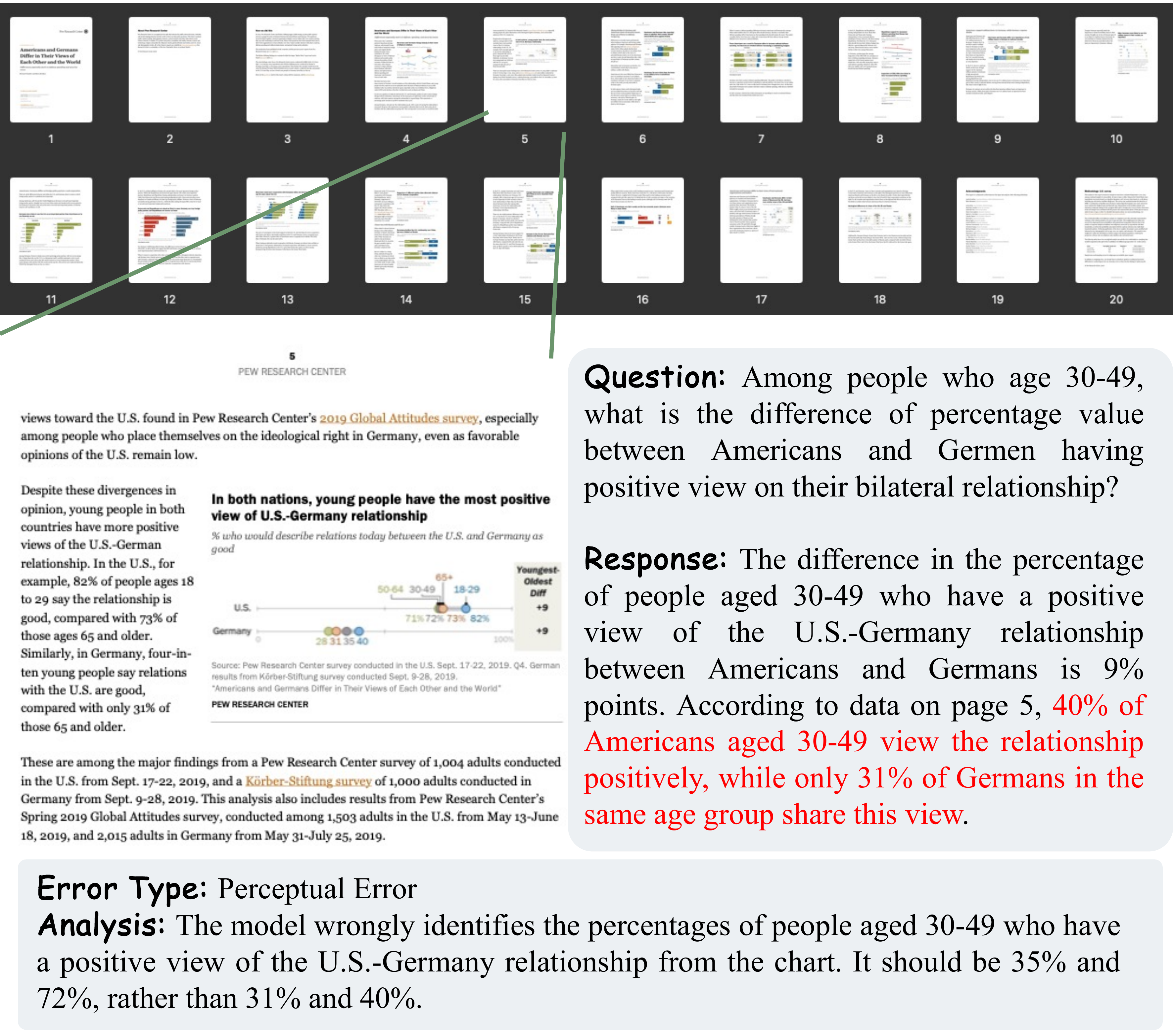}
    \caption{Error example about \textit{Perceptual Error}}
    \label{fig: Perceptual_error_1}
\end{figure}

\begin{figure}[!htbp]
    \centering
    \includegraphics[width=\linewidth]{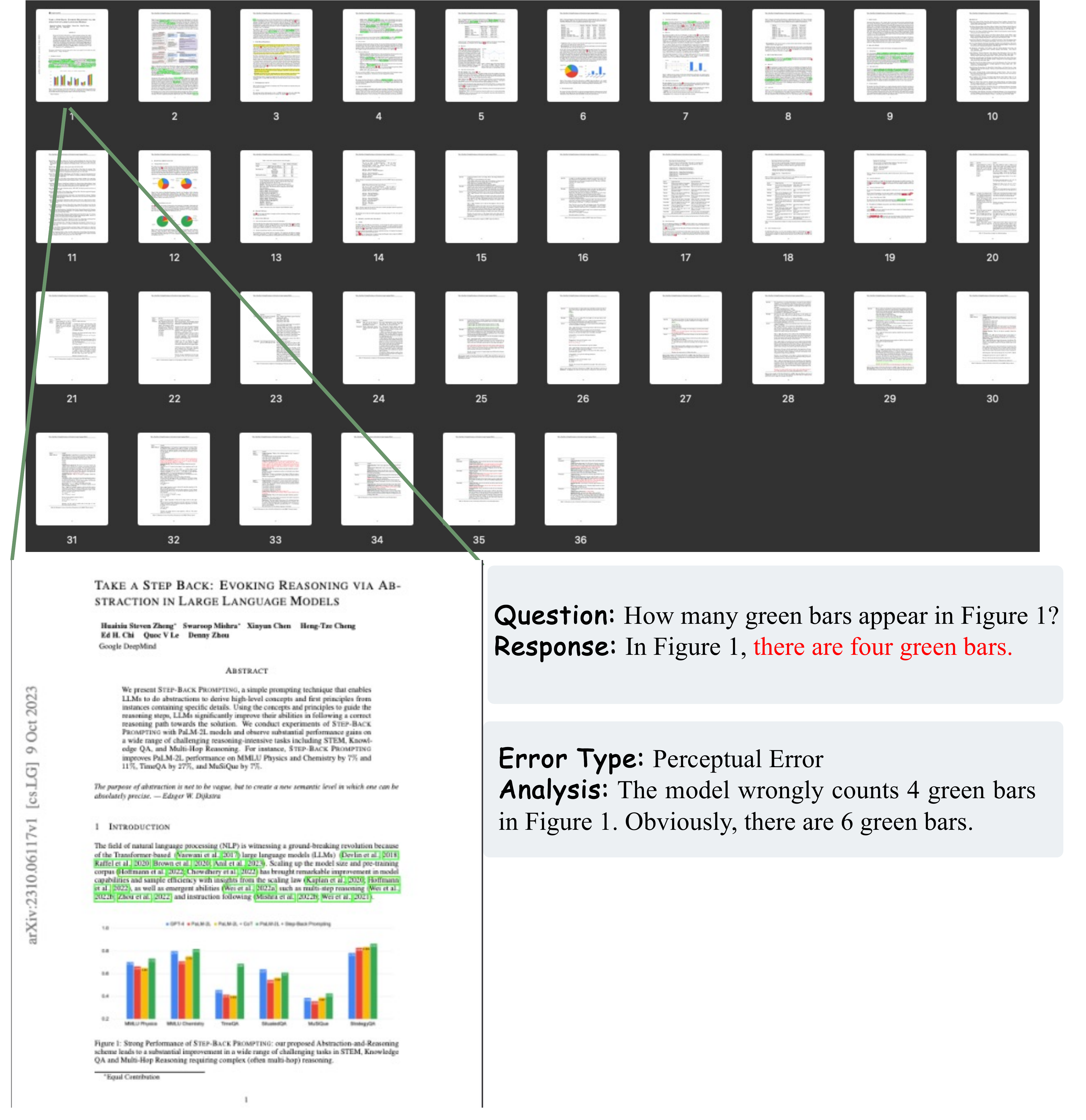}
    \caption{Error example about \textit{Perceptual Error}}
    \label{fig: Perceptual_error_2}
\end{figure}

We delve into the analysis of error by GPT-4o to further understand its bottlenecks and potentials on long-context document understanding. We manually check 72 incorrect responses and categorized their error reasons into 7 types. Except for the \textit{Extraction Error} caused by our automatic evaluation pipeline (see Appendix~\ref{subsec: Human Evaluation on the Automatic Evaluation Pipeline}), we detail and showcase another six reasons as below:

\noindent\textbf{Perceptual Error:} GPT-4o sometimes struggles to extract or understand visual information from document screenshots. For instance, it misinterprets the axes and colored circles in the charts shown in Figure~\ref{fig: Perceptual_error_1}. Additionally, it inaccurately counts the number of green bars in Figure~\ref{fig: Perceptual_error_2}. They demonstrate that even the cutting-edge LVLMs still fall short in fundamental perceptual capabilities.

\begin{figure}[!htbp]
    \centering
    \includegraphics[width=\linewidth]{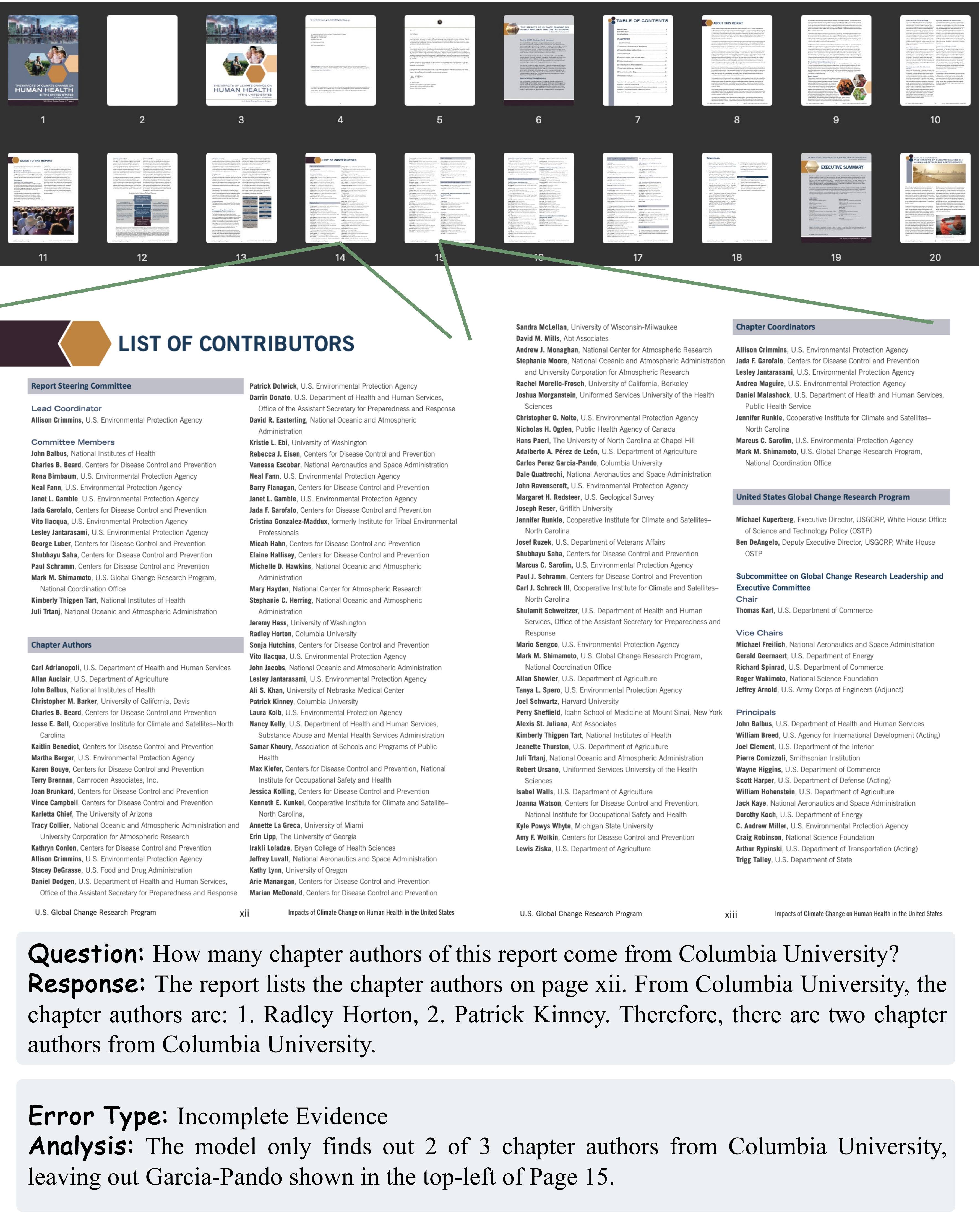}
    \caption{Error example about \textit{Incomplete Evidence}}
    \label{fig: Incomplete_evidence_1}
\end{figure}

\begin{figure}[!htbp]
    \centering
    \includegraphics[width=\linewidth]{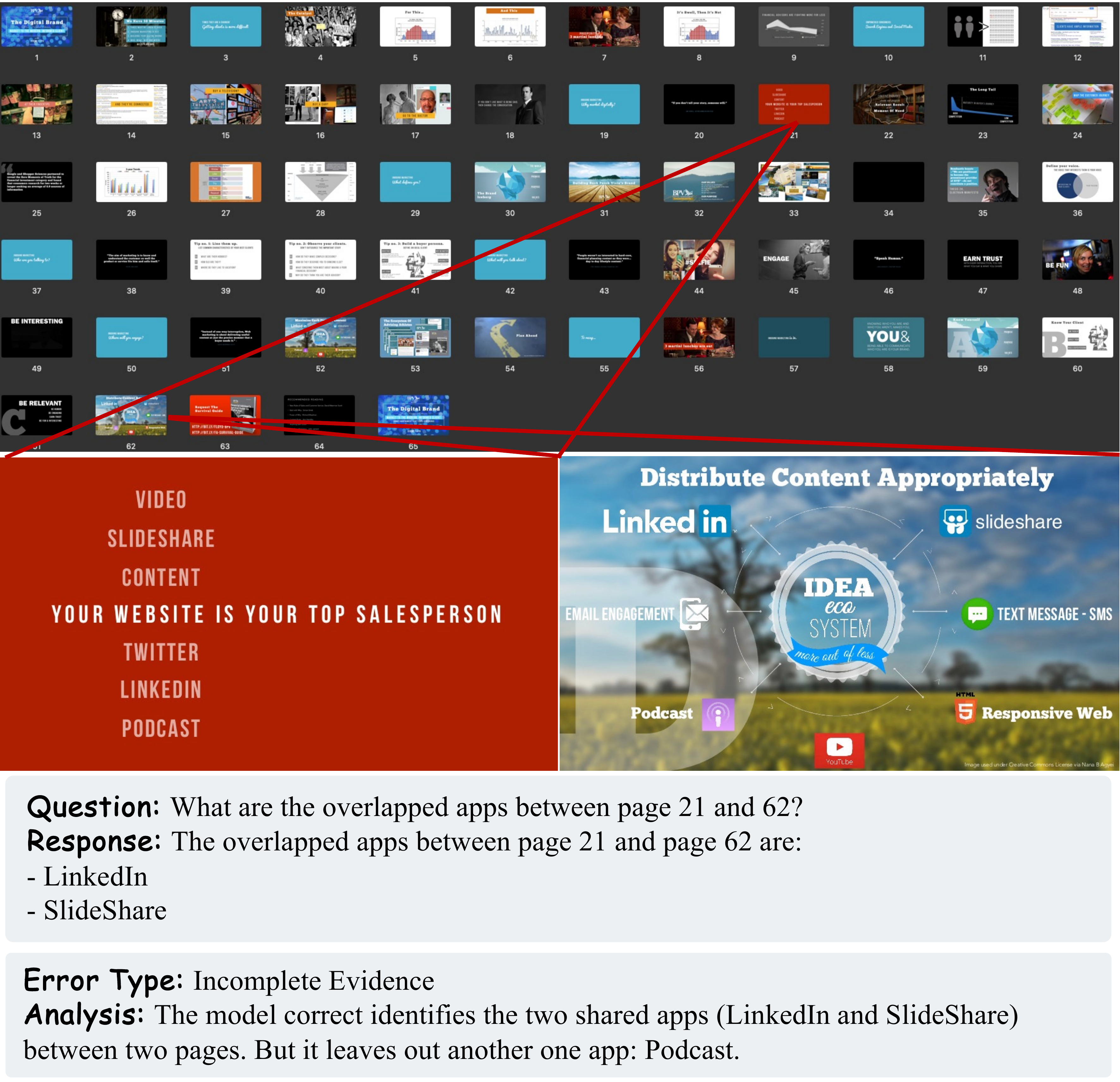}
    \caption{Error example about \textit{Incomplete Evidence}}
    \label{fig: Incomplete_evidence_2}
\end{figure}

\noindent\textbf{Incomplete Evidence:} Though GPT-4o has achieved significantly better \textit{global searching abilities} compared to other models when dealing with lengthy, multi-modal documents, it sometimes still omits certain information. For example, GPT-4o misses one chapter author from Columbia University in the full list (Figure~\ref{fig: Incomplete_evidence_1}). Additionally, it overlooks an app that appears across two pages (Figure~\ref{fig: Incomplete_evidence_2}).

\begin{figure}[!htbp]
    \centering
    \includegraphics[width=0.85\linewidth]{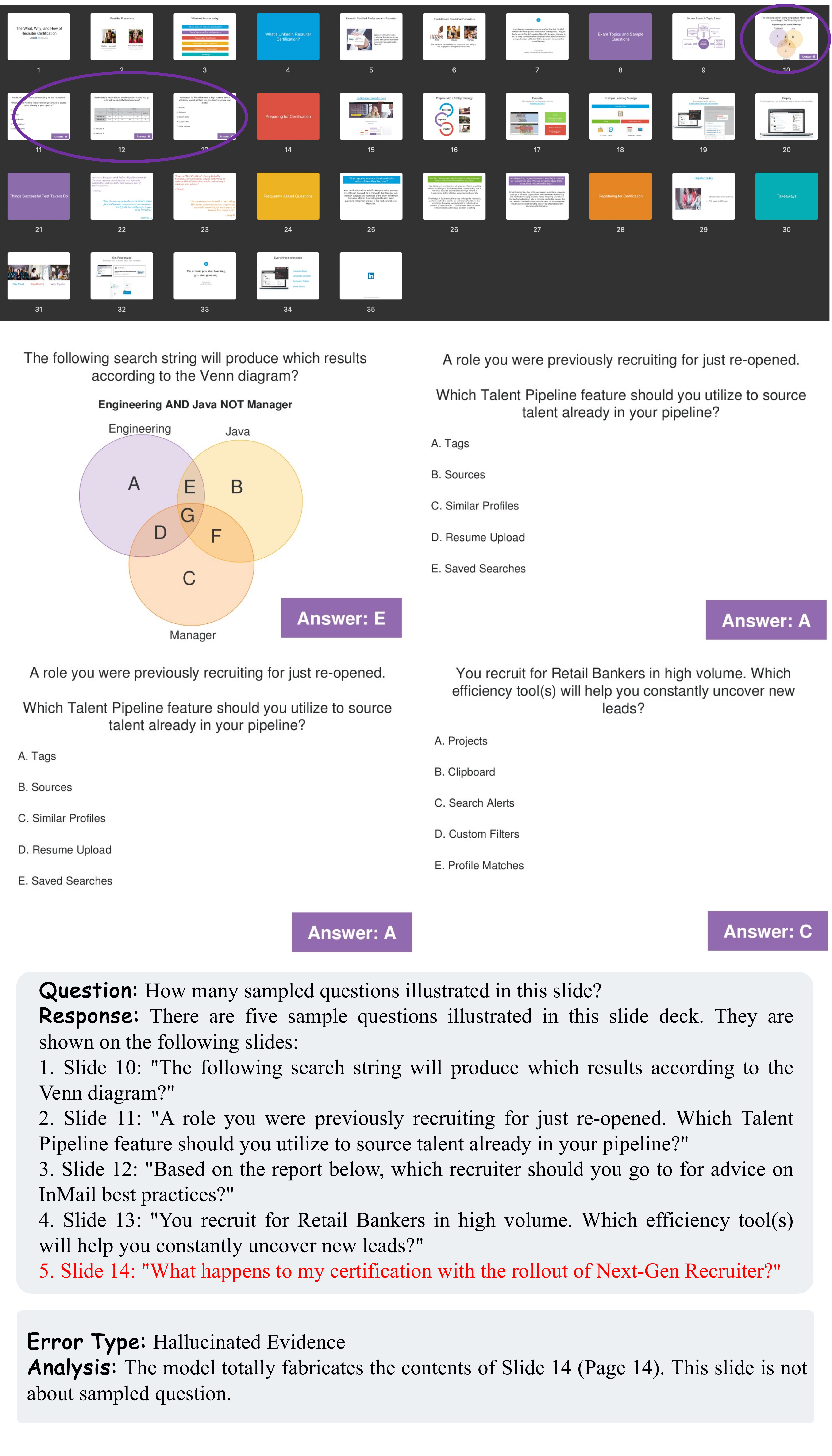}
    \caption{Error example about \textit{Hallucinated Evidence}}
    \label{fig: Hallucinated_evidence_1}
\end{figure}

\begin{figure}[!htbp]
    \centering
    \includegraphics[width=0.85\linewidth]{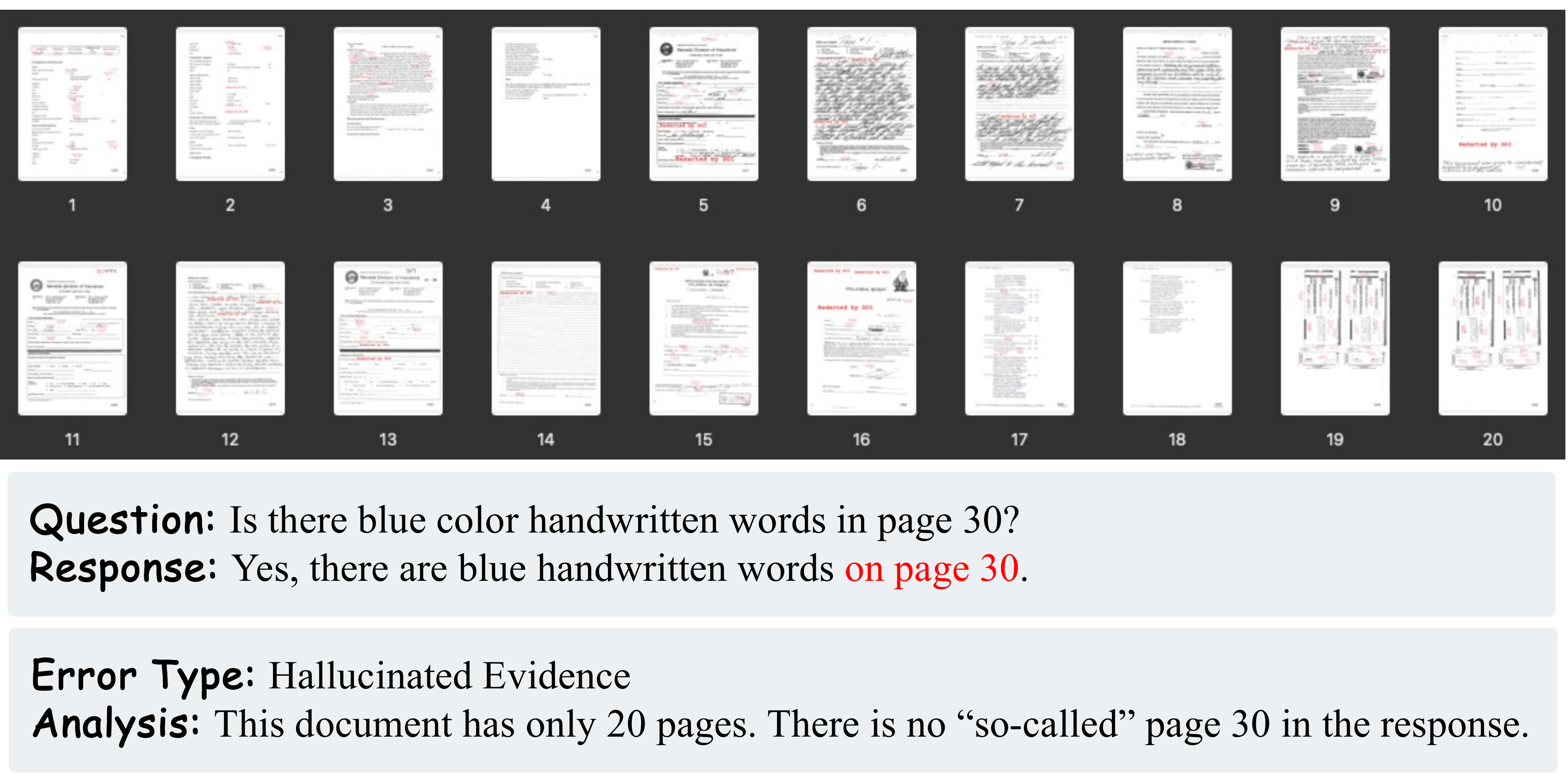}
    \caption{Error example about \textit{Hallucinated Evidence}}
    \label{fig: Hallucinated_evidence_2}
\end{figure}

\noindent\textbf{Hallucinated Evidence:} As stated in Section 3.4, GPT-4o adopts more aggressive strategies and tends to provide more false-positive answers. It sometimes even fabricates non-existent evidence in documents to support its incorrect responses. For example, it references a non-existent page in Figure~\ref{fig: Hallucinated_evidence_1}, and fabricates the content of a page in Figure~\ref{fig: Hallucinated_evidence_2}. The above examples clearly reveal the importance of further research on LVLMs' hallucination and safety.

\begin{figure}[!htbp]
    \centering
    \includegraphics[width=0.95\linewidth]{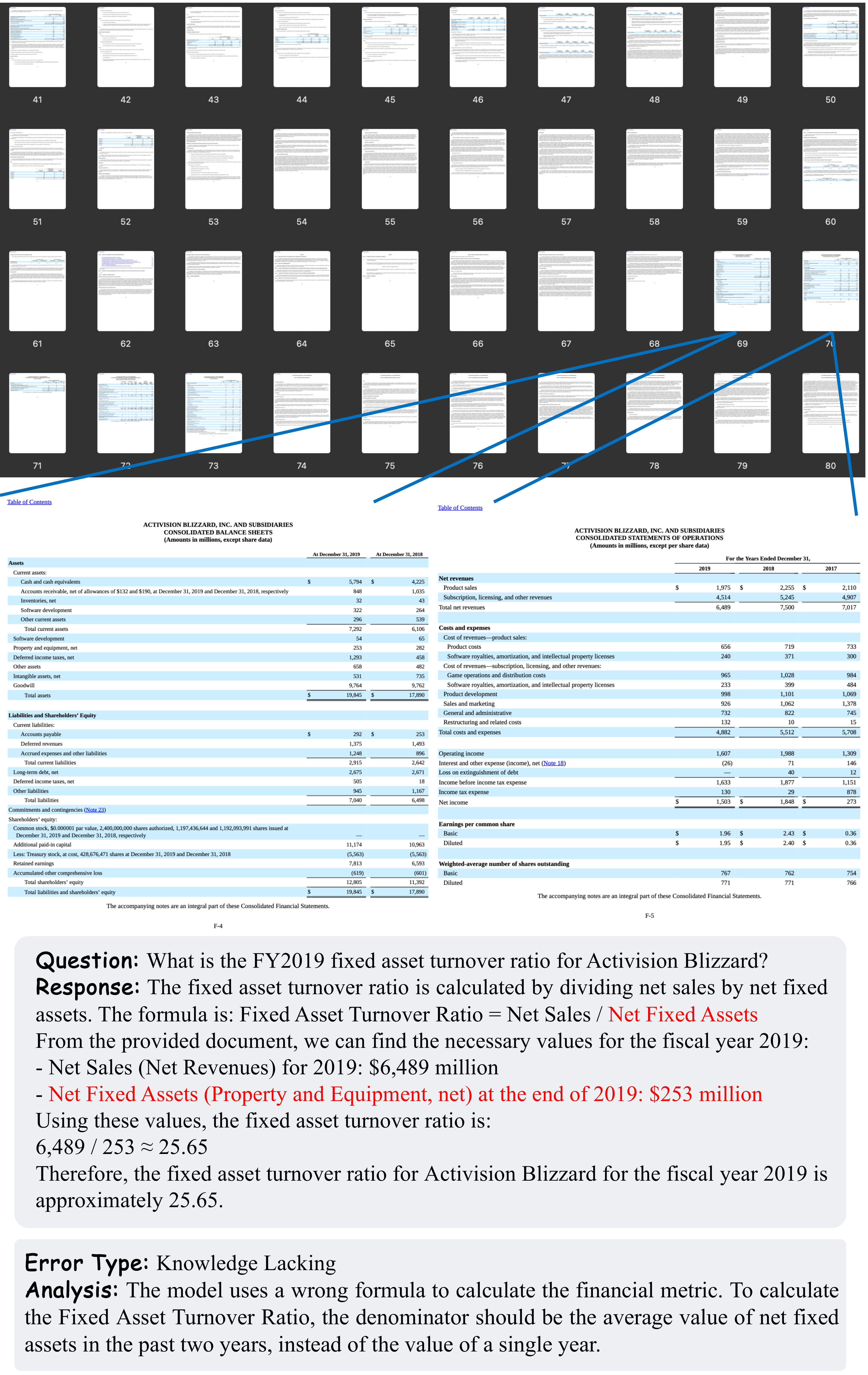}
    \caption{Error example about \textit{Knowledge Lacking}}
    \label{fig: Knowledge_lacking_1}
\end{figure}

\begin{figure}[!htbp]
    \centering
    \includegraphics[width=0.95\linewidth]{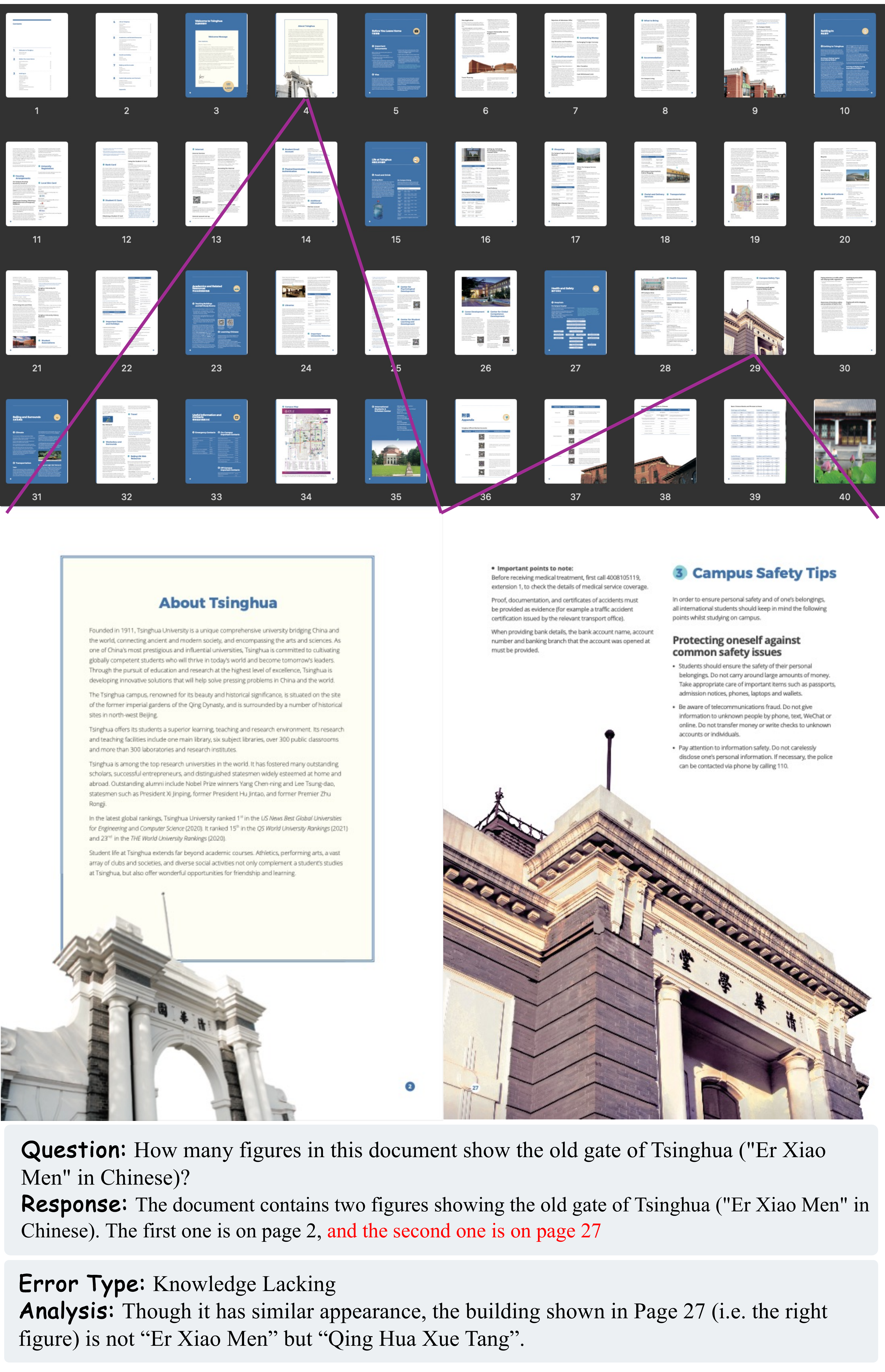}
    \caption{Error example about \textit{Knowledge Lacking}}
    \label{fig: Knowledge_lacking_2}
\end{figure}

\noindent\textbf{Knowledge Lacking:} Resolving certain questions requires both information from the documents and the parametric knowledge within LVLMs. We have observed error cases stemming from the absence of specific knowledge. For example, GPT-4o overlooks details about the \textit{fixed asset turnover ratio} and uses the single-point value instead of the average value to calculate this metric (Figure~\ref{fig: Knowledge_lacking_1}). Additionally, it misidentifies buildings at Tsinghua University in Figure~\ref{fig: Knowledge_lacking_2}.

\begin{figure}[!htbp]
    \centering
    \includegraphics[width=0.85\linewidth]{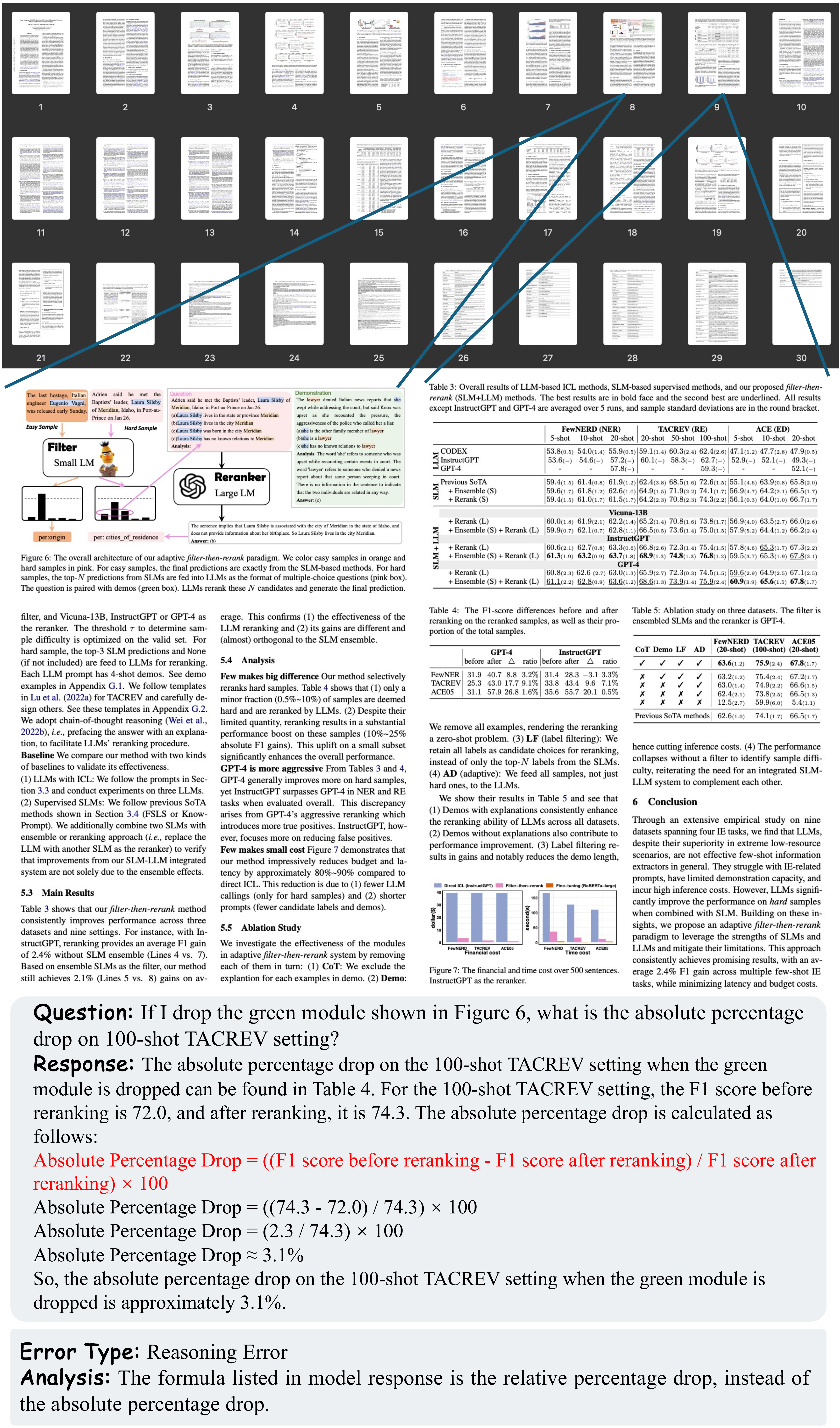}
    \caption{Error example about \textit{Reasoning Error}}
    \label{fig: reasoning_error_1}
\end{figure}

\begin{figure}[!htbp]
    \centering
    \includegraphics[width=0.9\linewidth]{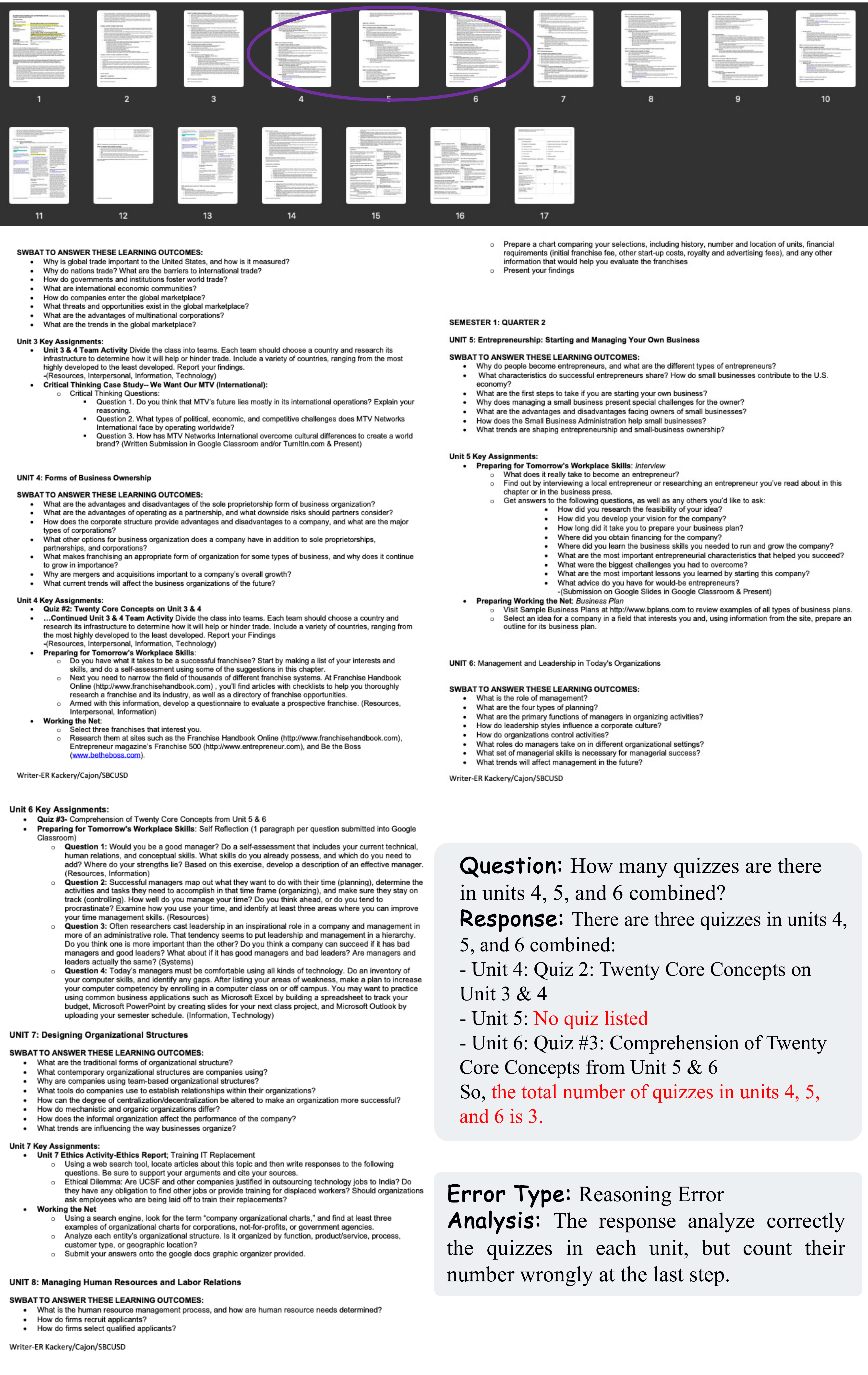}
    \caption{Error example about \textit{Reasoning Error}}
    \label{fig: reasoning_error_2}
\end{figure}

\noindent\textbf{Reasoning Error:} Though not a primary cause, flawed reasoning based on correctly collected evidence and information from documents can sometimes lead to wrong answers. For example, GPT-4o correctly gathers all data but calculates a relative percentage instead of an absolute percentage in Figure~\ref{fig: reasoning_error_1}. Additionally, as shown in Figure~\ref{fig: reasoning_error_2}, it correctly lists all quizzes but inaccurately counts them in the final step.

\begin{figure}[!htbp]
    \centering
    \includegraphics[width=0.85\linewidth]{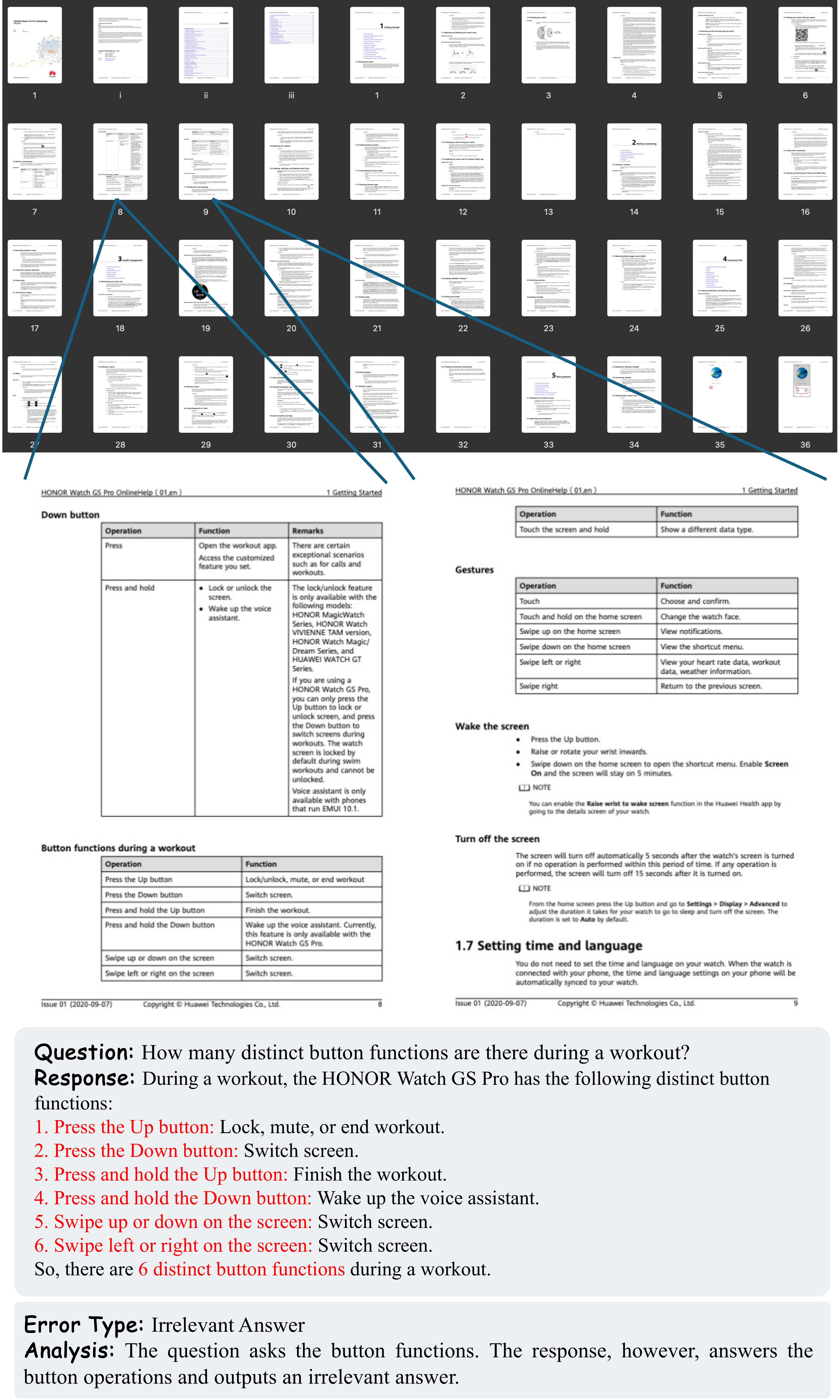}
    \caption{Error example about \textit{Irrelevant Answer}}
    \label{fig: irrelevant_answer_1}
\end{figure}

\begin{figure}[!htbp]
    \centering
    \includegraphics[width=0.75\linewidth]{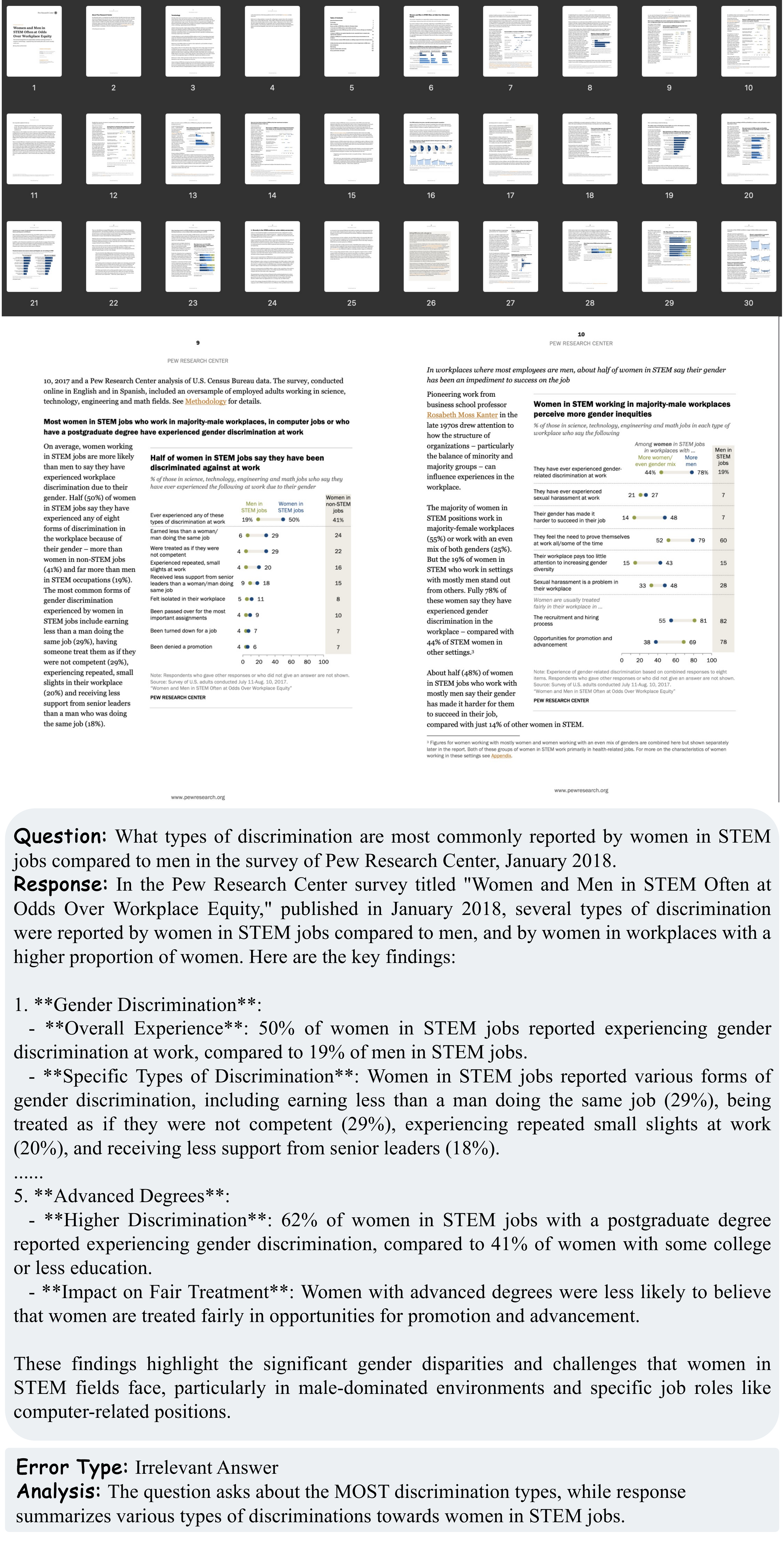}
    \caption{Error example about \textit{Irrelevant Answer}}
    \label{fig: irrelevant_answer_2}
\end{figure}

\noindent\textbf{Irrelevant Answer:} GPT-4o sometimes misunderstands the intent of questions and provides irrelevant responses. For instance, in Figure~\ref{fig: irrelevant_answer_1}, GPT-4o answers about button operations when the question asks about button functions. Similarly, in Figure~\ref{fig: irrelevant_answer_2}, where the question asks for the MOST discrimination type, GPT-4o summarizes all types instead.

\clearpage
\noindent\textbf{Perceptual Error: Case 1}

\clearpage
\noindent\textbf{Perceptual Error: Case 2}

\clearpage
\noindent\textbf{Incomplete Evidence: Case 1}

\clearpage
\noindent\textbf{Incomplete Evidence: Case 2}

\clearpage
\noindent\textbf{Hallucinated Evidence: Case 1}

\clearpage
\noindent\textbf{Hallucinated Evidence: Case 2}

\clearpage
\noindent\textbf{Knowledge Lacking: Case 1}

\clearpage
\noindent\textbf{Knowledge Lacking: Case 2}

\clearpage
\noindent\textbf{Reasoning Error: Case 1}

\clearpage
\noindent\textbf{Reasoning Error: Case 2}

\clearpage
\noindent\textbf{Irrelevant Answer: Case 1}

\clearpage
\noindent\textbf{Irrelevant Answer: Case 2}

\subsection{Case Study}
\label{appendix: case_study}

\begin{figure}[!htbp]
    \centering
    \includegraphics[width=0.95\linewidth]{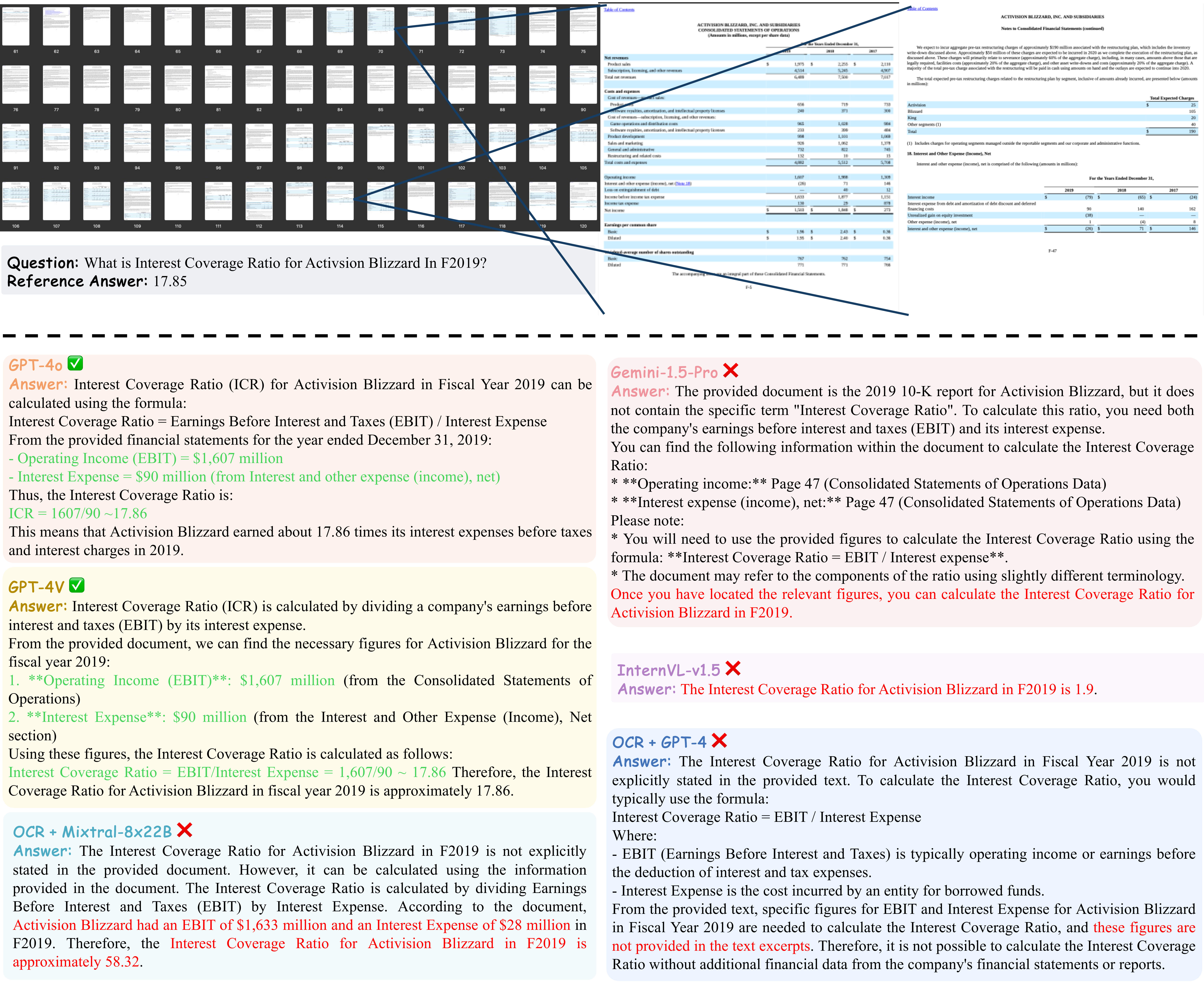}
    \caption{Case Study. Evidence source: table. The evidence pages are zoomed in. The correct extracted information and reasoning are colored in green, and the wrong ones are colored in red.}
    \label{fig: case_3}
\end{figure}

\clearpage
\begin{figure}
    \centering
    \includegraphics[width=0.7\linewidth]{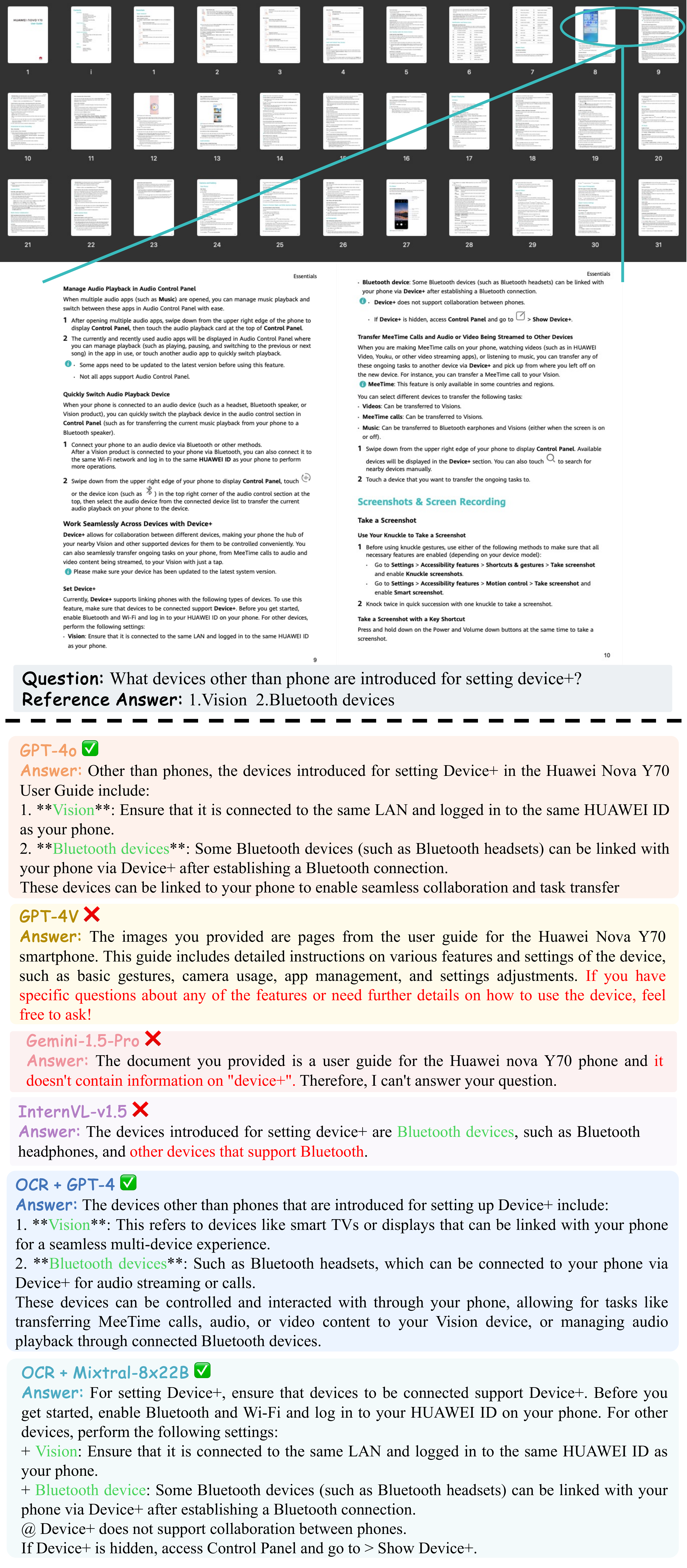}
    \caption{Case Study. Evidence source: text. The evidence pages are zoomed in. The correct extracted information and reasoning are colored in green, and the wrong ones are colored in red.}
    \label{fig: case_4}
\end{figure}

\clearpage
\begin{figure}
    \centering
    \includegraphics[width=0.7\linewidth]{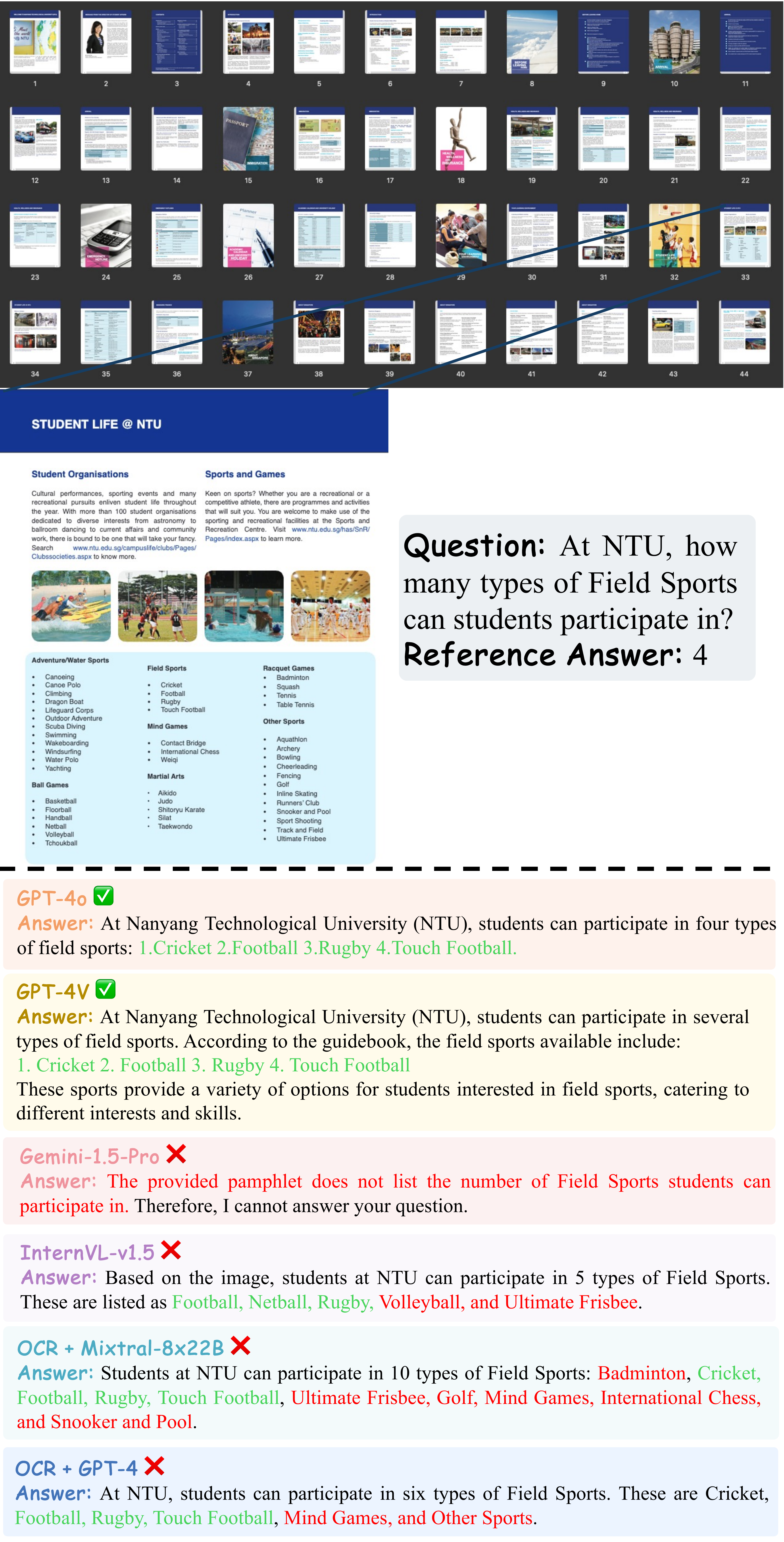}
    \caption{Case Study. Evidence source: layout. The evidence page is zoomed in. The correct extracted information and reasoning are colored in green, and the wrong ones are colored in red.}
    \label{fig: case_5}
\end{figure}

\begin{figure}[!b]
    \centering
    \includegraphics[width=0.7\linewidth]{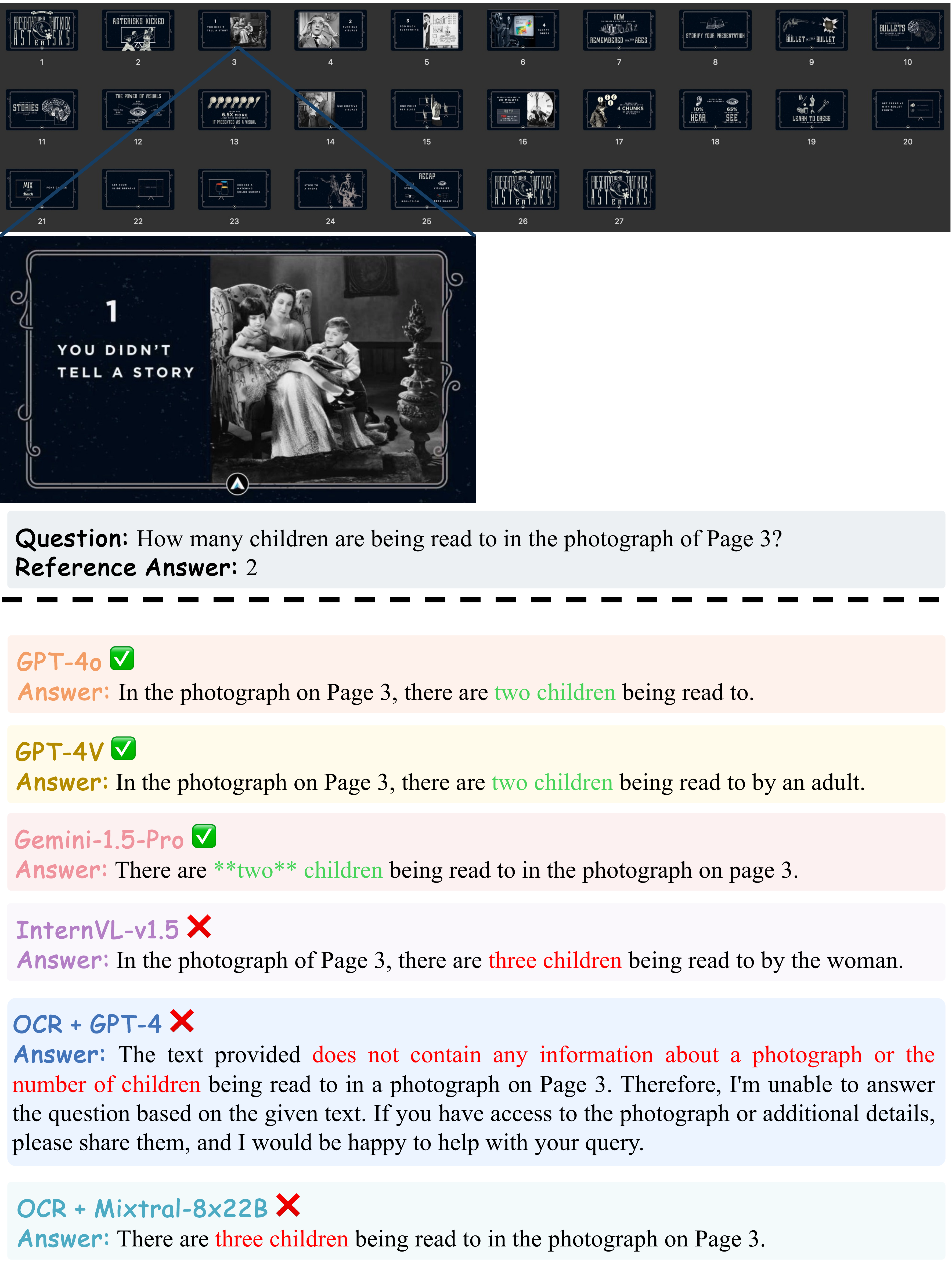}
    \caption{Case Study. Evidence source: image. The evidence page is zoomed in. The correct extracted information and reasoning are colored in green, and the wrong ones are colored in red.}
    \label{fig: case_1}
\end{figure}

\clearpage
\section{Limitations}
\label{appendix: limitations}

\benchmarkname is the first comprehensive benchmark designed to evaluate the long-context document understanding capabilities of LVLMs. While our benchmark addresses significant gaps in the previous datasets, we acknowledge several limitations.

One primary limitation is the scale of the benchmark. Currently, our benchmark includes a test set comprising 135 documents and 1,082 questions. It is much smaller compared to previous datasets. The complexity and difficulty of annotations limit the scale of our benchmark. As a long-context benchmark, our documents average about 50 pages and 20,000 tokens. And most questions require either complicated reasoning or cross-page comprehension. It takes more than one hour for an expert-level annotator to read through a single document, and then edit existing instances and create new instances on this document. Given the purpose of \benchmarkname as an evaluation benchmark, we prioritize annotation quality over quantity. Moreover, the results presented in Sections 3.3 and 3.4 confirm that the scale of our benchmark is sufficient for fine-grained evaluations across different document types, evidence sources, evidence pages, \etc. Additionally, we plan to expand our benchmark by adding more documents and questions in future iterations.

We roughly categorize these questions into three types, \ie single-page, cross-page, and unanswerable questions, based on whether evidence can be found in the documents and the number of evidence pages. However, unlike MMBench~\cite{mmbench2023} or MathVista~\cite{lu2024mathvista}, we provide no further taxonomy to classify some (\eg 7 or 20) fine-grained, evaluated reasoning or perception capabilities out of two main reasons: (1) Prior (\ie pre-annotation) taxonomy limits the diversity of the questions. Therefore we provide no predefined classifications in our guideline and encourage the expert-level annotators to freely write questions without constraints. (2) The intrinsic complexity of document understanding presents significant challenges for establishing a posterior (\ie post-annotation) taxonomy.

While there exist limitations in our benchmark, \benchmarkname surely represents a significant step forward in this field. We would iteratively maintain and refine this benchmark and hope it could push forward the development of long-context document understanding.
\section{Social Impacts}
\label{appendix: social_impacts}
The development and use of \benchmarkname may have potential societal implications. For instance, biased or inaccurate outputs from benchmarked models could perpetuate harmful stereotypes or reinforce existing social inequalities. Additionally, the ability to process and analyze long documents could potentially be used to surveil or monitor individuals' personal information. Developers and users of \benchmarkname benchmark must be aware of these potential consequences and take steps to ensure responsible development and deployment of AI models.

\section{Author Statement}
\label{appendix: author_statement}
The authors state that all of the previous datasets that we collected are licensed under the Creative Commons license (CC-BY) or other open-source licenses. Using this dataset should abide by the \href{https://openai.com/policies/terms-of-use}{policy of OpenAI}. Regarding the newly collected documents, we manually check them to ensure their availability for academic use. Should any authors request the removal of their documents, we will promptly comply.

\end{document}